%% file: KG-subgraph-camera-ready.tex
\documentclass[nohyperref]{article}
\usepackage{iclr2024_conference,times}

\usepackage{hyperref}
\definecolor{citeColor}{RGB}{0,20,115}
\hypersetup{colorlinks,linkcolor={citeColor},citecolor={citeColor},urlcolor={citeColor}} 

\input{math_commands.tex}

\usepackage{hyperref}
\usepackage{url}

\usepackage[utf8]{inputenc} 
\usepackage[T1]{fontenc}    
\usepackage{url}            
\usepackage{booktabs}       
\usepackage{amsfonts}       
\usepackage{nicefrac}       
\usepackage{microtype}      
\usepackage{xcolor}         
\usepackage{ulem}
\usepackage{colortbl} 

\usepackage{array}
\usepackage{algorithm}
\usepackage{algorithmic}
\usepackage{amsmath}
\usepackage{amsthm}
\usepackage{amsfonts}
\usepackage{enumerate}
\usepackage{subfigure}
\usepackage{bm}
\usepackage{tabularx}
\usepackage{enumitem}
\usepackage{multirow}
\usepackage{xcolor}
\usepackage{nicefrac}
\usepackage{booktabs}
\usepackage{bbding}
\usepackage{caption}
\usepackage{graphicx}
\usepackage{bbding}
\usepackage{wrapfig}
\usepackage{lipsum}
\usepackage{pifont}
\usepackage{makecell}
\usepackage{bbm}
\usepackage{makecell}
\usepackage[toc,page,header]{appendix}
\usepackage{minitoc}
\usepackage{centernot}
\usepackage{mathtools}
\usepackage{stmaryrd}

\newtheorem{theorem}{Theorem}

\newtheorem{definition}{Definition}

\newtheorem{advantage}{Advantage}

\newcolumntype{L}[1]{>{\raggedright\let\newline\\\arraybackslash\hspace{0pt}}m{#1}}
\newcolumntype{C}[1]{>{\centering\let\newline  \\\arraybackslash\hspace{0pt}}m{#1}}%
\newcolumntype{R}[1]{>{\raggedleft\let\newline \\\arraybackslash\hspace{0pt}}m{#1}}

\definecolor{shadecolor}{rgb}{0.92,0.92,0.92}
\definecolor{grey}{rgb}{0.5, 0.5, 0.5}

\title{
	Less is More:
    One-shot Subgraph Reasoning \\
    on Large-scale Knowledge Graphs
}

\makeatletter
\renewcommand*{\@fnsymbol}[1]{\ensuremath{\ifcase#1\or \dagger\or \ddagger\or
		\mathsection\or \mathparagraph\or \|\or **\or \dagger\dagger
		\or \ddagger\ddagger \else\@ctrerr\fi}}
\makeatother

\author{
	Zhanke Zhou$^{1}$ \quad Yongqi Zhang$^{2}$ \quad Jiangchao Yao$^{3}$ \quad Quanming Yao$^{4}$ \quad Bo Han$^{1}\!$
	\thanks{Correspondence to Bo Han (bhanml@comp.hkbu.edu.hk).} \\
	$^1$TMLR Group, Hong Kong Baptist University \quad \\
    $^2$The Hong Kong University of Science and Technology (Guangzhou) \quad \\
	$^3$CMIC, Shanghai Jiao Tong University \quad
	$^4$Tsinghua University \quad \\
	\textnormal{\{cszkzhou, bhanml\}@comp.hkbu.edu.hk} 
	\quad 
	\textnormal{yzhangee@connect.ust.hk} \\ 
	\textnormal{sunarker@sjtu.edu.cn}  
	\quad
	\textnormal{qyaoaa@tsinghua.edu.cn}
}

\iclrfinalcopy 
\begin{document}

	\doparttoc 
	\faketableofcontents 
	\parttoc 
	
	\maketitle
	
	\vspace{-8pt}
	\begin{abstract}
		To deduce new facts on a knowledge graph (KG),
		a {link predictor}
		learns from the graph structure
		and collects local evidence to find the answer to a given query.
		However,
		existing methods suffer from a severe scalability problem due to the utilization of the whole KG for {prediction}, which
		hinders their promise on large-scale KGs and cannot be directly addressed by vanilla sampling methods.
		%
		In this work,
		we propose
		the {\textit{one-shot-subgraph} link prediction}
		to achieve efficient and adaptive prediction.
		The design principle is that,
		instead of directly acting on the whole KG,
		the prediction procedure is decoupled into two steps, \textit{i.e.},
		(i) extracting \textit{only one} subgraph according to the query
		and (ii) predicting on this single, query-dependent subgraph.
		We reveal that 
		the non-parametric and computation-efficient heuristics Personalized PageRank (PPR)
		can effectively identify the potential answers and supporting evidence.
		With efficient subgraph-based prediction,
		we further introduce the automated searching
		of the optimal configurations in both data and model spaces.
		Empirically,
		we achieve
		promoted efficiency and leading performances
		on five large-scale benchmarks.
		The code is publicly available at:\\
		\url{https://github.com/tmlr-group/one-shot-subgraph}.
	\end{abstract}

    \vspace{-8pt}
	\section{Introduction}
	\label{sec:intro}
	
	A knowledge graph (KG)
	is graph-structural data with relational facts~\citep{battaglia2018relational, ji2020survey, chen2020review},
	based on which,
	one can conduct {link prediction}
	to deduce new facts from existing ones.
	The typical problem 
	is to find the answer entity for the specific query, \textit{e.g.},
	to find the answer \textit{Los Angeles} to the query \textit{(LeBron, lives\_in, ?)}.
	With continuous advances in recent years, 
	the {link prediction on KG} has been widely applied in
	recommendation systems~\citep{cao2019unifying,wang2019kgat},
	online question answering~\citep{huang2019knowledge}, 
	and drug discovery~\citep{yu2021sumgnn}.

	The {prediction} system learns from the local structure of a KG,
	where existing methods can be generally summarized as two categories:
	(1) \textit{semantic} models
	that implicitly capture the local evidence
	through learning the low-dimensional embeddings of entities and relations
	\citep{bordes2013translating,dettmers2017convolutional,zhang2019quaternion,wang2017knowledge};
	and (2) \textit{structural} models 
	that explicitly explore the KG's structure 
	based on relational paths or graphs
	with
	recurrent neural networks (RNNs)
	or
	graph neural networks (GNNs)
	\citep{das2017go,schlichtkrull2018modeling, sadeghian2019drum,vashishth2019composition,teru2019inductive, qu2021rnnlogic, zhu2021neural, zhang2021knowledge}.
	%
	
	Although achieving leading performances,
	these structural models suffer from a severe scalability problem as
	the entire KG 
	has been potentially or progressively taken for {prediction}.
	This inefficient manner
	hinders their application and optimization on large-scale KGs, 
	\textit{e.g.}, 
	OGB~\citep{hu2020open}.
	Thus, it raises an open question:
	\textit{Is all the information necessary for prediction on knowledge graphs?}
	%
	Intuitively,
	only partial knowledge stored in the human brain 
	is relevant to a {given} question,
	which is extracted by recalling
	and then utilized in the careful thinking procedure.
	Similarly,
	generating candidates and then ranking promising ones
	are common practices
	in large-scale recommendation systems
	with millions even billions of users~\citep{cheng2016wide, covington2016deep}.
	These facts
	motivate us to
	conduct efficient link prediction with an effective sampling mechanism for KGs.

	In this work, we 
	propose the novel
	{\textit{one-shot-subgraph} link prediction} on 
	a knowledge graph.
	This idea {paves} a new way to
	alleviate the scalability problem of existing KG methods
	from a data-centric perspective:
	decoupling the prediction procedure into two steps
	with a corresponding sampler and a predictor.
	Thereby, the {prediction} of a specific query is conducted by
	(i) \textit{fast} sampling of one query-dependent subgraph with the sampler
	and (ii) \textit{slow} {prediction} on the subgraph with predictor.

	Nevertheless,
	it is non-trivial to achieve efficient and effective {link prediction} 
	on large-scale KGs
	due to the two major challenges.
	{(1) \textit{Sampling speed and quality}:}
	The fast sampling of the one-shot sampler 
	should be capable of 
	covering the essential evidence and potential answers
	to support the query.
	{(2) \textit{Joint optimization}:}
	The sampler and predictor 
	should be optimized jointly
	to avoid trivial solutions and 
	to guarantee the expressiveness and adaptivity
	of the overall model
	to a specific KG.

	To solve these challenges technically,
	we first implement the
	one-shot-subgraph link prediction
	by the non-parametric and computation-efficient 
	Personalized PageRank (PPR),
	which is capable of effectively identifying the potential answers
	without requiring learning.
	With the efficient subgraph-based prediction,
	we further propose to search the
	data-adaptive configurations
	in both data and model spaces.
	%
	We show it unnecessary to utilize the whole KG in inference;
	meanwhile,
	only a relatively small proportion of information
	(\textit{e.g.}, 10\% of entities) 
	is sufficient.
	%
	Our main contributions are:

	\vspace{-8pt}
	\begin{itemize}[leftmargin=10pt]
		\item
		We conceptually formalize the new manner of 
		\textit{one-shot-subgraph link prediction} 
		on KGs (Sec.~\ref{sec:one-shot-subgraph})
		and technically instantiate it
		with efficient heuristic samplers and powerful KG predictors
		(Sec.~\ref{ssec: realization}).
		
		\item
		We solve a non-trivial and bi-level optimization problem
		of searching the optimal configurations in both data and model spaces (Sec.~\ref{ssec: optimization})
		and theoretically analyze
		the extrapolation power (Sec.~\ref{ssec: extrapolation}).
		
		\item
		We conduct extensive experiments on five large-scale datasets  
		and achieve an average of $94.4\%$ improvement in efficiency of prediction 
		and $6.9\%$ promotion in effectiveness of prediction (Sec.~\ref{sec:exp}).
	\end{itemize}

	\vspace{-8pt}
	\section{Preliminaries}
	\label{sec:couple}
	\vspace{-8pt}

	\textbf{Notations.}
	A knowledge graph is denoted as $\mathcal{G} \! = \! (\mathcal{V}, \mathcal{R}, \mathcal{E})$,
	where the entity set
	$\mathcal{V}$,
	the relation set 
	$\mathcal{R}$,
	and factual edge set 
	$\mathcal{E}  \! = \! \{(x, r, v): x,v \! \in \! \mathcal{V}, r \! \in \! \mathcal{R}\}$.
	Here,
	a fact is formed as a triplet and denoted as $(x, r, v)$.
	Besides,
	a sampled \textit{subgraph} of $\mathcal{G}$
	is denoted as
	$\mathcal{G}_s \! = \! (\mathcal{V}_s, \mathcal{R}_s, \mathcal{E}_s)$,
	satisfying that
	$\mathcal{V}_s \! \subseteq \! \mathcal{V}, \mathcal{R}_s \! \subseteq \! \mathcal{R}, \mathcal{E}_s \! \subseteq \! \mathcal{E}$.
	The atomic problem of link prediction
	is denoted as a query $(u, q, ?)$, \textit{i.e.},
	given the query entity $u$ and query relation $q$,
	to find the answer entity $v$, making $(u, q, v)$ valid.

	\textbf{Semantic models}
	encode entities and relations to low-dimensional 
	entity embeddings $\bm{H}_{\mathcal{V}} \! \in \! \mathbb{R}^{|\mathcal{V}| \! \times \! D_{v}}$
	and relation embeddings $\bm{H}_{\mathcal{R}} \! \in \! \mathbb{R}^{|\mathcal{R}|  \! \times \! D_{r}}$,
	where $D_{v}, \! D_{r}$ are dimensions.
	A scoring function $f_{\bm{\theta}}$, \textit{e.g.},
	TransE~\citep{bordes2013translating}
	or QuatE~\citep{zhang2019quaternion},
	{is necessary here to quantify the plausibility of a query triplet $(u,q,v)$
	with the learned embeddings $(\bm{h}_u, \bm{h}_q, \bm{h}_v)$ as}
	$f_{\bm{\theta}}: (\mathbb{R}^{D_{v}}, \mathbb{R}^{D_{r}}, \mathbb{R}^{D_{v}})  \! \mapsto \! \mathbb{R}$.
	
	
	\textbf{Efficient semantic models}
	aim to reduce the size of entity embeddings.
	NodePiece~\citep{galkin2022nodepiece}
	proposes an anchor-based approach that obtains fixed-size embeddings 
	as $\mathcal{G} \! \xmapsto{f_{\bm{\theta}}} \! \bm{\hat{H}}_{\mathcal{V}} \! \in \! \mathbb{R}^{N \! \times \! D_{v}}$
	and inference as $(\bm{\hat{H}}, \mathcal{G}) \xmapsto{f_{\bm{\theta}}, (u, q)} \hat{\bm{Y}}$,
	where 
 $\hat{\bm{Y}}$ are the scores of candidate answers,
 and $N  \! \ll \! |\mathcal{V}|$.
	Designed to reduce the embedding size,
	NodePiece cannot reduce the graph size for structural models.

	
	\textbf{Structural models}
	are based on 
	relational \textit{paths} or \textit{graphs} for {link prediction}.
	Wherein
	the \textit{path}-based models, \textit{e.g.},
	MINERVA~\citep{das2017go}, 
	DRUM~\citep{sadeghian2019drum}, 
	and RNNLogic~\citep{qu2021rnnlogic},
	aim to learn probabilistic and logical rules
	and well capture the sequential patterns in KGs.
	The \textit{graph}-based models
	such as 
	R-GCN~\citep{schlichtkrull2018modeling} 
	and CompGCN~\citep{vashishth2019composition}
	propagate the low-level entity embeddings
	among the neighboring entities to obtain high-level embeddings.
	Recent methods
	NBFNet~\citep{zhu2021neural}
	and RED-GNN~\citep{zhang2021knowledge}
	progressively propagate from $u$ to its 
	neighborhood in a breadth-first-searching (BFS) manner.
	%


	
	\textbf{Sampling-based structural models}
	adopt graph sampling approaches 
	to decrease the computation complexity,
	which can be categorized into two-fold as follows.
	First,
	\textit{subgraph-wise} methods such as 
	GraIL~\citep{teru2019inductive} and CoMPILE~\citep{mai2021communicative}
	extract enclosing subgraphs 
	between query entity $u$ and each candidate answer $v$.
	Second,
	\textit{layer-wise} sampling methods 
	extract a subgraph for message propagation
	in each layer of a model.
	Wherein
	designed for node-level tasks on homogeneous graphs,
	GraphSAGE~\citep{hamilton2017inductive} and FastGCN~\citep{chen2018fastgcn}
	randomly sample neighbors around the query entity.
	While the {KG} sampling methods, \textit{e.g.},
	DPMPN~\citep{xu2019dynamically}, AdaProp~\citep{zhang2022learning}, and AStarNet~\citep{zhu2022learning},
	extract a learnable subgraph in $\ell$-th layer by the GNN model in $\ell$-th layer,
	coupling the procedures of sampling and prediction.

	\section{\textit{One-shot-subgraph}  link prediction on knowledge graphs}
	\label{sec:one-shot-subgraph}
	
	
	To achieve efficient link prediction,
	we conceptually design the 
	\textit{one-shot-subgraph} manner
	that avoids directly predicting with the entire KG.
	We formalize this new manner in the following Def.~\ref{def: KG decouple}.
	
	
	\begin{definition}[One-shot-subgraph Link Prediction on Knowledge Graphs]
		\label{def: KG decouple}
		Instead of directly predicting on the original graph $\mathcal{G}$,
		the {prediction} procedure is decoupled to two-fold:
		(1) one-shot sampling of a query-dependent subgraph and 
		(2) {prediction} on this subgraph.
		The prediction pipeline becomes
		\begin{equation}	
			\mathcal{G} \xmapsto{g_{\bm{\phi}}, (u, q)} \mathcal{G}_s \xmapsto{f_{\bm{\theta}}} \hat{\bm{Y}},
			\label{eqn: decouple}
		\end{equation}
		where the sampler $g_{\bm{\phi}}$ generates only one subgraph $\mathcal{G}_s$ 
		(satisfies $|\mathcal{V}_s| \! \ll \! 	|\mathcal{V}|, |\mathcal{E}_s| \! \ll \! 	|\mathcal{E}|$)
		conditioned on the given query $(u, q, ?)$.
		Based on subgraph $\mathcal{G}_s$, the predictor $f_{\bm{\theta}}$ 
		outputs the final predictions $\hat{\bm{Y}}$.
	\end{definition}

	\textbf{Comparison with existing manners of prediction.}
	%
	In brief,
	semantic models follow the manner of
	encoding the entire $\mathcal{G}$ to the embeddings 
	$\bm{H} \! = \! (\bm{H}_{\mathcal{V}}, \bm{H}_{\mathcal{R}})$ 
	and {prediction} (inference) without $\mathcal{G}$, \textit{i.e.},
	\begin{equation}	
		\bm{H} \xmapsto{f_{\bm{\theta}}, (u, q)} \hat{\bm{Y}},
		\; \text{s.t.} \;
		\mathcal{G} \! \xmapsto{f_{\bm{\theta}}} \! \bm{H},
		\nonumber
	\end{equation}
	which is parameter-expensive, 
	especially when encountering a large-scale graph 
	with a large entity set.
	%
	On the other hand,
	structural models adopt the {way} of
	learning and {prediction} with $\mathcal{G}$,
	\textit{i.e.},
	\begin{equation}	
		\mathcal{G} \xmapsto{f_{\bm{\theta}}, (u, q)}  \hat{\bm{Y}},
		\nonumber
	\end{equation}
	that directly or progressively conduct {prediction} with the entire graph $\mathcal{G}$.
	Namely,
	all the entities and edges can be potentially taken in the {prediction} of one query,
	which is computation-expensive.
	%
	By contrast,
	our proposed one-shot prediction manner (Def.~\ref{def: KG decouple})
	enjoys the advantages \ref{advantage: complexity} \& \ref{advantage: scope} as follows.
	
	\begin{advantage}[Low complexity of computation and parameter]
		The {one-shot-subgraph} model is
		(1) computation-\textit{efficient}:
		the extracted subgraph 
		is much smaller than the original graph, \textit{i.e.}, 
		$|\mathcal{V}_s| \! \ll \! 	|\mathcal{V}|$ and
		$|\mathcal{E}_s| \! \ll \! 	|\mathcal{E}|$;
		and (2) parameter-\textit{efficient}:
		it avoids learning the expensive entities' embeddings.
	\label{advantage: complexity}
	\end{advantage}
	
	\begin{advantage}[{Flexible propagation scope}]
		The scope here refers to
		the range of message propagation starting from the query entity $u$.
		Normally,
		an $L$-layer structural method
		will propagate to the full $L$-hop neighbors of $u$.
		By contrast,
		the adopted one-shot sampling 
		enables the bound of propagation scope within the extracted subgraph,
		where the scope is decoupled from the model's depth $L$.
		\label{advantage: scope}
	\end{advantage}


	\textbf{Comparison with existing sampling methods.}
	Although promising to the scalability issue,
	existing sampling methods for structural models
	are not efficient or effective enough
	for learning and {prediction} on large-scale KGs.
	To be specific,
	the \textit{random} layer-wise sampling methods
	cannot guarantee the coverage of answer entities,
	\textit{i.e.}, $\mathbbm{1}(v \! \in \! \mathcal{V}_{u})$.
	By contrast,
	the \textit{learnable} layer-wise sampling methods 
	extract the query-dependent subgraph $\mathcal{G}^{(\ell)}_s$ in $\ell$-th layer 
	via the GNN model $f_{\bm{\theta}}^{(\ell)}$ in $\ell$-th layer as
	\begin{equation}	
		\mathcal{G}
		\xmapsto{f_{\bm{\theta}}^{(1)}, (u, q)}  \mathcal{G}^{(1)}_s 
		\xmapsto{f_{\bm{\theta}}^{(2)}, (u, q)}  \mathcal{G}^{(2)}_s 
		\mapsto \cdots \mapsto
		\mathcal{G}^{(L-1)}_s  \xmapsto{f_{\bm{\theta}}^{(L)}, (u, q)} \hat{\bm{Y}},
		\nonumber
	\end{equation}
	coupling the sampling and prediction procedures that
	(1) are bundled with specific architectures
	and (2) with extra computation cost in the layer-wise sampling operation.
	%
	Besides, the subgraph-wise sampling methods
	extract the enclosing subgraphs 
	between query entity $u$ and each candidate answer $v \! \in \! \mathcal{V}$,
	and then independently reason on each of these subgraphs 
	to obtain the final prediction $\hat{\bm{Y}}$ as
	\begin{equation}	
		\big\{ \hat{\bm{Y}}_v: \mathcal{G} \xmapsto{(u, v)} \mathcal{G}_{s}^{(u,v)}
		\xmapsto{f_{\bm{\theta}}, (u, q, v)} \hat{\bm{Y}}_v 
		\big\}_{v \in \mathcal{V}}  
		\mapsto \hat{\bm{Y}}.
		\nonumber
	\end{equation}
	Note these approaches are extremely expensive on large-scale graphs,
	as each candidate $(u, v)$ corresponds to a subgraph to be scored.
	By contrast, one-shot sampling manner enjoys the advantage~\ref{advantage: efficient sampling}.
	
	\begin{advantage}[High efficiency in subgraph sampling]
		Our proposed {prediction} {manner}
		requires \textbf{only one subgraph} for answering \textbf{one query},
		which is expected to cover all the potential answers and supporting facts.
		Notably, this query-specific subgraph is extracted
		in a \textbf{one-shot} and \textbf{decoupled} manner
		that does not involve the predictor, reducing the computation cost in subgraph sampling.
		%
	\label{advantage: efficient sampling}
	\end{advantage}
	
	\begin{figure}[t!]
		\centering
		\includegraphics[width=13cm]{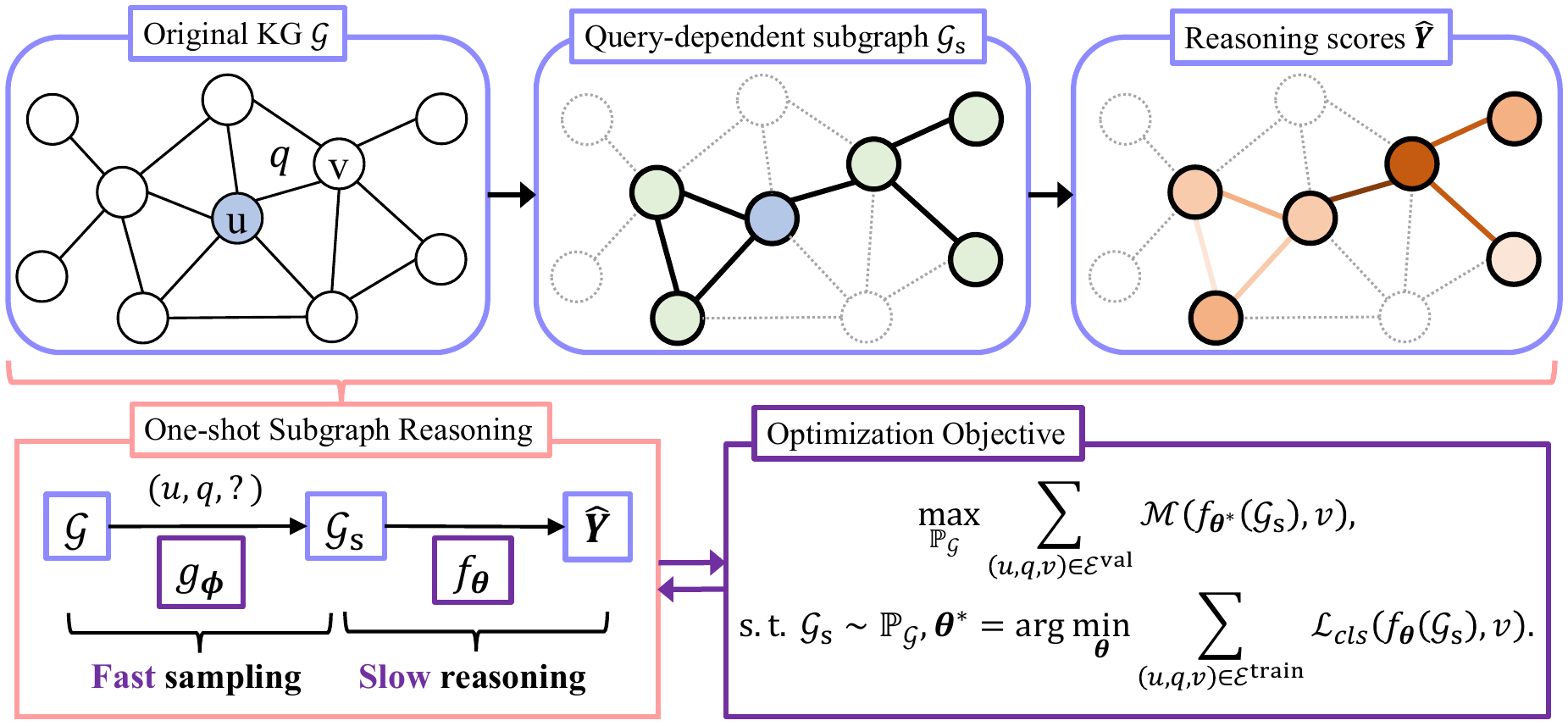}
		\vspace{-4px}
		\caption{
			The proposed framework of {one-shot-subgraph link prediction}.
			Specifically,
			(1) the sampler $g_{\bm{\phi}}$ extracts a subgraph $\mathcal{G}_s$ 
			from the whole graph $\mathcal{G}$
			with regard to the given query,
			and (2) the predictor $f_{\bm{\theta}}$ conducts deliberative {prediction}  
			on the extracted subgraph $\mathcal{G}_s$ 
			and obtains the final predictions $\hat{\bm{Y}}$.
		}
		\label{fig: decoupled framework}
		\vspace{-10px}
	\end{figure}

	\vspace{-6px}
	\section{Instantiating the one-shot-subgraph link prediction}
	\label{sec:decouple instantiation}
	
	Note that
	it is non-trivial to achieve Def.~\ref{def: KG decouple}, wherein
	(i) the implementation of sampler, 
	(ii) the architecture of predictor,
	and (iii) the method to optimize these two modules
	need to be figured out.
	Here,
	the major challenge lies in the sampler,
	which is required to be
	efficient, query-dependent, and local-structure-preserving.
	In this section, 
	we elaborate on the detailed implementation (Sec.~\ref{ssec: realization}),
	set up a bi-level problem for optimization (Sec.~\ref{ssec: optimization}), 
	and 
	investigate the extrapolation power (Sec.~\ref{ssec: extrapolation}).

	\subsection{Realization: {Three-step prediction with personalized pagerank}}
	\label{ssec: realization}
	
	
	\textbf{Overview.}
	As illustrated in Fig.~\ref{fig: decoupled framework},
	the three key steps of our method
	are 
	\textbf{(1)} generating the sampling distribution
	$\mathbb{P}_{\mathcal{G}}$
	by sampler $g_{\bm{\phi}}$,
	\textbf{(2)} sampling a subgraph from the distribution as
	$\mathcal{G}_s \! \sim \! \mathbb{P}_{\mathcal{G}}$
	with top-$K$ entities and edges,
	and \textbf{(3)} predicting on the subgraph $\mathcal{G}_s$ 
	and acquiring the final prediction $\hat{\bm{Y}}$
	by predictor $f_{\bm{\theta}}$.
	The three-step procedure is 
	summarized in Algorithm~\ref{alg:KGD-forward}
	and elaborated on as follows.

	%
	%

	\textbf{Step-1. Generate sampling distribution.}
	Previous studies show that
	$v$ is generally near to $u$
	\citep{zhu2021neural, zhang2021knowledge},
	and the relational paths connecting $u$ and $v$
	that support the query
	also lie close to $u$ 
	\citep{das2017go, sadeghian2019drum}.
	To efficiently capture the local evidence of $u$, 
	we choose the heuristic
	Personalized PageRank (PPR) 
	\citep{page1999pagerank, jeh2003scaling}
	as the sampling indicator.
	Note that PPR is not only
	\textit{efficient} for its non-parametric nature
	but also \textit{query-dependent} and \textit{local-structure-preserving}
	for its single-source scoring that starts from $u$. 

	Specifically,
	PPR starts propagation from $u$
	to evaluate the importance of each neighbor of $u$
	and generates the PageRank scores as the sampling probability
	that encodes the local neighborhood of the query entity $u$.
	Besides,
	it can also preserve the locality and connectivity of subgraphs
	by leveraging the information from a large neighborhood.
	Given a query entity $u$,
	we obtain the probability $\bm{p} \! \in \! \mathbb{R}^{|\mathcal{V}|}$ 
	\vspace{-6px}
	\begin{equation}	
		\texttt{Non-parametric indicator}: \;
		\bm{p}^{(k+1)} 
		\leftarrow \alpha \cdot \bm{s} + (1-\alpha)  \cdot \bm{D}^{-1} \bm{A} \cdot \bm{p}^{(k)},
		\label{eqn: PPR sampler}
	\end{equation}
	by iteratively updating the scores
	up to $K \! = \! 100$ steps
	to approximate the converged scores efficiently.
	Here, the initial score
	$\bm{p}^{(0)} \! = \! \bm{s} \! = \! \mathbbm{1}(u) \! \in \! \{0,1\}^{|\mathcal{V}|}$
	indicates the query entity $u$ to be explored.
	{
	The two-dimensional degree matrix $\bm{D} \! \in \! \mathbb{R}^{|\mathcal{V}| \times |\mathcal{V}|}$ 
	and adjacency matrix $\bm{A} \! \in \! \{0,1\}^{|\mathcal{V}| \times |\mathcal{V}|}$ 
	together work as the transition matrix,
	wherein $\bm{A}_{ij} \! = \! 1$ means an edge $(i, r, j) \! \in \! \mathcal{E}$
	and $\bm{D}_{ij} \! = \! \text{degree}(v_i)$ if $i \! = \! j$ else $\bm{D}_{ij} \! = \! 0$.
	The damping coefficient $\alpha$ 
	($=0.85$ by default)
	controls the differentiation degree.
	}
	
	\textbf{Step-2. Extract a subgraph.} 
	Based on the PPR scores $\bm{p}$ (Eqn.~\ref{eqn: PPR sampler}), 
	the subgraph 
	$\mathcal{G}_s \! = \! (\mathcal{V}_s, \mathcal{E}_s, \mathcal{R}_s)$
	(where $\mathcal{R}_s \! = \! \mathcal{R}$) is extracted
	with the most important entities and edges.
	Denoting
	the sampling ratios of entities and edges as 
	$r^{q}_{\mathcal{V}}, r^{q}_{\mathcal{E}} \! \in \! (0,1]$
	that depend on the query relation $q$,
	we sample
	$|\mathcal{V}_s| \! = \! r^{q}_{\mathcal{V}} \! \times \! |\mathcal{V}|$ entities
	and 
	$|\mathcal{E}_s| \! = \! r^{q}_{\mathcal{E}} \! \times \! |\mathcal{E}|$ edges 
	from the full graph $\mathcal{G}$.
	With the $\texttt{TopK}(D, P, K)$ operation
	that picks up top-$K$ elements from candidate $D$ \textit{w.r.t.} probability $P$,
	the entities $\mathcal{V}_s$ and edges $\mathcal{E}_s$  are given as
	\vspace{-2px}
	\begin{equation}
		\begin{split}
			\texttt{Entity Sampling:}& \;
			\mathcal{V}_s \leftarrow \texttt{TopK} \Big( \mathcal{V}, \; \bm{p}, \; K \! = \! r^{q}_{\mathcal{V}} \! \times \! |\mathcal{V}| \Big),
			\\
			\texttt{Edge Sampling:}& \;
			\mathcal{E}_s \leftarrow \texttt{TopK} \Big( \mathcal{E}, \; 
			\{\bm{p}_x \! \cdot \! \bm{p}_o: x,o \! \in \! \mathcal{V}_s, (x,r,o) \! \in \! \mathcal{E} \},
			\; K \! = \! r^{q}_{\mathcal{E}} \! \times \! |\mathcal{E}| \Big).
		\end{split}	
		\vspace{-14px}
	\end{equation}
	%
	\textbf{Step-3. Reason on the subgraph.}
	From the model's perspective,
	we build the configuration space of the predictor
	and further utilize the advantages
	of existing structural models introduced in Sec.~\ref{sec:couple}.
	Three query-dependent message functions \texttt{MESS($\cdot$)}
	are considered, including
	DRUM, NBFNet, and RED-GNN,
	which are elaborated in Appendix.~\ref{app:implementation}.
	Note the effective message is propagated from $u$
	to the sampled entities $o \in \mathcal{V}_s$.
	Generally,
	the layer-wise updating of representations is formulated as 
	\vspace{-2px}
	\begin{equation}	
		\begin{split}
			\texttt{Indicating:}
			\bm{h}^{(0)}_{o} \leftarrow& \mathbbm{1}(o = u),
			\\
			\texttt{Propagation:}
			\bm{h}^{(\ell+1)}_{o} \leftarrow&
			\texttt{DROPOUT} 
			\biggl( \!
			\texttt{ACT} 		
			\Bigl( \!
			\texttt{AGG} 
			\big\{
			\texttt{MESS}
			(\bm{h}^{(\ell)}_{x}, \bm{h}^{(\ell)}_{r}, \bm{h}^{(\ell)}_{o}) \! : \!
			(x,r,o) \! \in \! \mathcal{E}_s 
			\! \big\} 
			\! \Bigr)
			\! \biggr),
		\end{split}
		\label{eqn: general predictor}
		\vspace{-20px}
	\end{equation}
	where 
	$\mathbbm{1}(o = u)$
	is the indicator function that only labels the 
	query entity $u$ in a query-dependent manner.
	%
	After a $L$-layer propagation,
	the predictor outputs the final score $\hat{\bm{y}}_{o}$ 
	of each entity $o \! \in \! \mathcal{V}_s$
	based on their representations $\bm{h}^{(L)}_{o}$ as
	$\hat{\bm{y}}_{o} \! = \! \texttt{Readout}(\bm{h}^{(L)}_{o}, \bm{h}^{(L)}_{u}) \! \in \! \mathbb{R}$.
	The loss function $\mathcal{L}_{cls}$ adopted in the training phase
	is the commonly-used binary cross-entropy loss on all the sampled entities. 
	Namely,
	$\mathcal{L}_{cls} \! = \! -\sum_{o \in \mathcal{V}_s} y_{o} \log(\hat{\bm{y}}_o) \! + \! (1-y_{o})\log(1 \! - \! \hat{\bm{y}}_o)$,
	where $y_{o} \! = \! 1$ if $o \! = \! v$
	else $y_{o} \! = \! 0$.
	
	
	
	\vspace{-8px}
	\begin{algorithm}[H]
		\caption{One-shot-subgraph Link Prediction on Knowledge Graphs}
		\label{alg:KGD-forward}
		\begin{algorithmic}[1]
			\REQUIRE 
			{
			KG $\mathcal{G} \! = \! (\mathcal{V}, \mathcal{R}, \mathcal{E})$, 
			degree matrix $\bm{D} \! \in \! \mathbb{R}^{|\mathcal{V}| \times |\mathcal{V}|}$,
			adjacency matrix $\bm{A} \! \in \! \{0,1\}^{|\mathcal{V}| \times |\mathcal{V}|}$,
			damping coefficient $\alpha$,
			maximun PPR iterations $K$ ,}
			query $(u, q, ?)$, sampler $g_{\bm{\phi}}$, predictor $f_{\bm{\theta}}$.
			\STATE {\color{gray} \texttt{\# Step-1. Generate sampling distribution}}
			\STATE
			initialize $\bm{s} \leftarrow \mathbbm{1}(u), \; \bm{p}^{(0)} \leftarrow \mathbbm{1}(u)$.
			\FOR{$k=1\dots K$}
			\STATE
			$\bm{p}^{(k+1)} 
			\leftarrow \alpha \cdot \bm{s} + (1-\alpha)  \cdot \bm{D}^{-1} \bm{A} \cdot \bm{p}^{(k)}$.
			\ENDFOR
			
			\STATE {\color{gray} \texttt{\# Step-2. Extract a subgraph $\mathcal{G}_s$}}
			\STATE
			$
			\mathcal{V}_s \leftarrow \texttt{TopK} 
			( \mathcal{V}, \; \bm{p}, \; K \! = \! r^q_{\mathcal{V}} \! \times \! |\mathcal{V}| )
			$.
			\STATE
			$
			\mathcal{E}_s \leftarrow \texttt{TopK} 
			( \mathcal{E}, \; 
			\{\bm{p}_u \! \cdot \! \bm{p}_v: u,v \! \in \! \mathcal{V}_s, (u,r,v) \! \in \! \mathcal{E} \},
			\; K \! = \! r^q_{\mathcal{E}} \! \times \! |\mathcal{E}| )
			$.
			\STATE {\color{gray} \texttt{\# Step-3. Reason on the subgraph}}
			\STATE
			initialize representations
			$\bm{h}^{(0)}_{o} \leftarrow \mathbbm{1}(o = u)$.
			
			\FOR{$\ell=1\dots L$}
			\STATE
			$
			\bm{h}^{(\ell)}_{o} 
			\! \leftarrow \!
			\texttt{DROPOUT} 
			(
			\texttt{ACT} 		
			(
			\texttt{AGG} 
			\{ 
			\texttt{MESS}
			(\bm{h}^{(\ell-1)}_{x}, \! \bm{h}^{(\ell-1)}_{r}, \! \bm{h}^{(\ell-1)}_{o}) \! : \!
			(x,r,o) \! \in \! \mathcal{E}_s 
			\} 
			)
			)
			$.
			\ENDFOR
			\RETURN Prediction $\hat{\bm{y}}_{o} \! = \! \texttt{Readout}(\bm{h}^{(L)}_{o}, \bm{h}^{(L)}_{u})$ 
			for each entity $o \in \mathcal{V}_{s}$.
			\label{step:return}
		\end{algorithmic}
	\end{algorithm}
	\vspace{-20px}
	
	\subsection{Optimization: Efficient searching for data-adaptive configurations}
	\label{ssec: optimization}
	
\begin{figure}[t!]
		\centering
		\includegraphics[width=14cm]{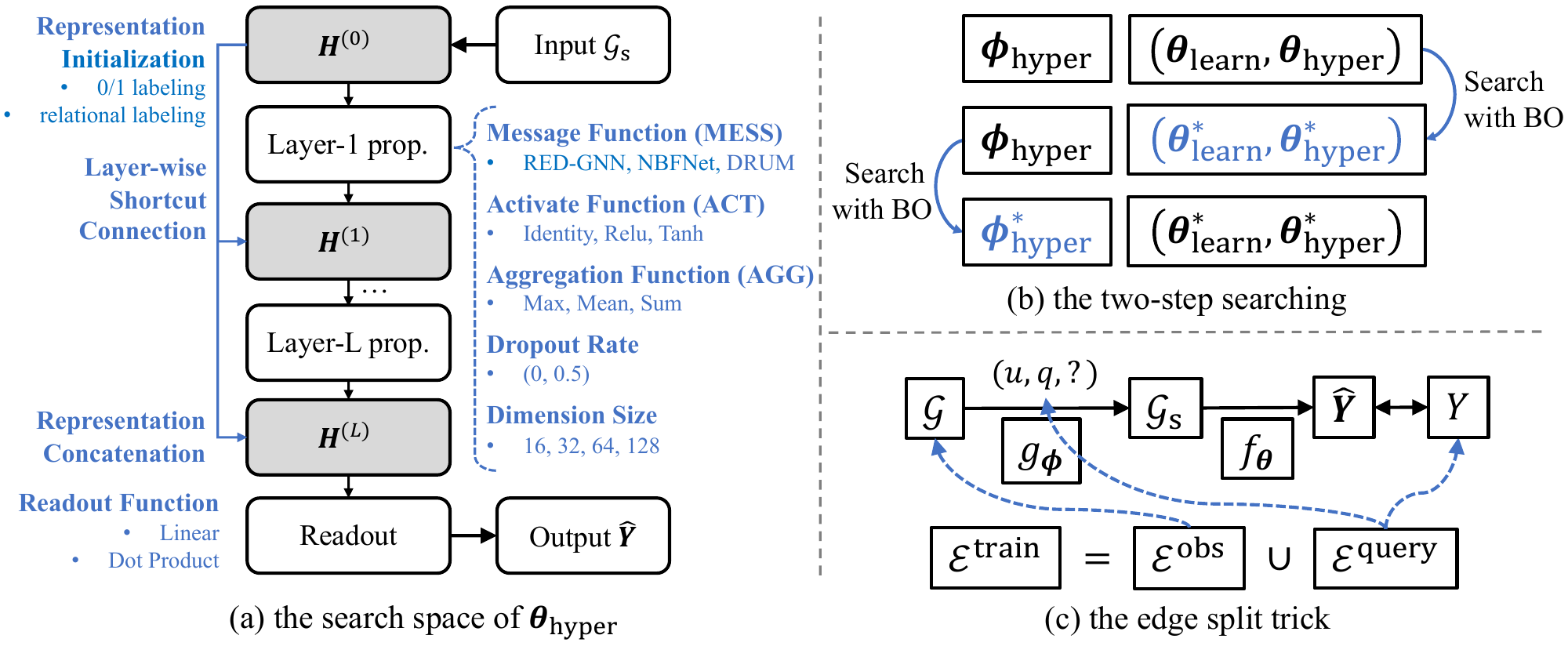}
		\vspace{-18px}
		\caption{
			Illustrations of the optimization procedure.
			Note the predictor in (a) is with hyper-parameters and learnable parameters
			in each layer's propagation,
			and the $\bm{H}^{(L)}$ indicates the representations in $L$-th layer.			
			By contrast, the sampler only contains hyper-parameters
			as it does not require learning.
		}
		\label{fig: optimization}
		\vspace{-14px}
	\end{figure}

	\textbf{Search space.}
	Note that hyperparameters $(r^{q}_{\mathcal{V}}, r^{q}_{\mathcal{E}})$ and $L$ 
	play important roles in Algorithm~\ref{alg:KGD-forward}.
	Analytically,
	a larger subgraph  
	with larger $r^{q}_{\mathcal{V}}, r^{q}_{\mathcal{E}}$
	does not indicate a better performance,
	as more irrelevant information is also covered.
	Besides,
	a deeper model with a larger $L$ can capture more complex patterns
	but is more likely to suffer from the
	over-smoothing problem~\citep{oono2019graph}.
	Overall,
	the $(r^{q}_{\mathcal{V}}, r^{q}_{\mathcal{E}})$
	are for sampler's hyper-parameters $\bm{\phi}_{\text{hyper}}$.
	In addition to $L$,
	predictor's hyper-parameters $\bm{\theta}_{\text{hyper}}$ contain
	several intra-layer or inter-layer designs, as illustrated in Fig.~\ref{fig: optimization}(a).

	
	{	
	\textbf{Search problem.}
	Next, we propose the bi-level optimization problem
	to adaptively search for the optimal configuration 
	$(\bm{\phi}^*_{\text{hyper}}, \bm{\theta}^*_{\text{hyper}})$
	of design choices on a specific KG, namely,
	}
	\vspace{-4px}
	\begin{equation}
		\begin{split}
				\bm{\phi}^*_{\text{hyper}} =&
				\arg \max_{\bm{\phi}_{\text{hyper}}} 
				\mathcal{M}(
				f_{(\bm{\theta}_{\text{hyper}}^*, \bm{\theta}_{\text{learn}}^*)},
				g_{\bm{\phi}_{\text{hyper}}}, \mathcal{E}^{\text{val}}),  \\
				\; \text{s.t.} \;
				\bm{\theta}_{\text{hyper}}^*  =&
				\arg \max_{\bm{\theta}_{\text{hyper}}}
				\mathcal{M}(f_{(\bm{\theta}_{\text{hyper}}, \bm{\theta}_{\text{learn}}^*)}, g_{\bar{\bm{\phi}}_{\text{hyper}}}, \mathcal{E}^{\text{val}}),
				%
		\end{split}
	\label{eqn: search objective}
	\vspace{-20px}
	\end{equation}
	where the performance measurement $\mathcal{M}$
	can be Mean Reciprocal Ranking (MRR) or Hits@k.
	Note the non-parametric sampler $g_{\bm{\phi}}$
	only contains hyper-parameters $\bm{\phi}_{\text{hyper}}$.
	As for predictor $f_{\bm{\theta}}$,
	its $\bm{\theta} \! =  \! (\bm{\theta}_{\text{hyper}}, \bm{\theta}_{\text{learn}})$ 
	also includes learnable $\bm{\theta}_{\text{learn}}$ that
	$\bm{\theta}_{\text{learn}}^* \! = \! 
	\arg \min_{\bm{\theta}_{\text{learn}}}
	\mathcal{L}_{cls}(f_{(\bm{\theta}_{\text{hyper}}, \bm{\theta}_{\text{learn}})}, g_{\bar{\bm{\phi}}_{\text{hyper}}}, \mathcal{E}^{\text{train}})$.

	{
	\textbf{Search algorithm.}	
	Directly searching on both data and model spaces
	is expensive
	due to the large space size and data scale.
	Hence,
	we split the search into two sub-processes as Fig.~\ref{fig: optimization}(b), \textit{i.e.},}
	\vspace{-4pt}
	\begin{itemize}[leftmargin=10pt]
		\item
		First, we freeze the sampler $g_{\bar{\bm{\phi}}}$
		(with constant $\bm{\phi}_{\text{hyper}}$)
		to search for the optimal predictor $f_{\bm{\theta}^*}$
		with
		(1) the hyper-parameters optimization 
		for $\bm{\theta}^*_{\text{hyper}}$
		and 
		(2) the stochastic gradient descent 
		for $\bm{\theta}^*_{\text{learn}}$.
	
		\item
		Then, we freeze the predictor $f_{\bm{\theta}^*}$
		and search for the optimal sampler $g_{\bm{\phi}^*}$,
		simplifying to pure hyper-parameters optimization
		for $\bm{\phi}^*_{\text{hyper}}$
		in a zero-gradient manner 
		with low computation complexity.
	\end{itemize}
	
	Specifically,
	we follow the sequential model-based
	Bayesian Optimization (BO)~\citep{bergstra2013hyperopt, hutter2011sequential}
	to obtain $\bm{\phi}^*_{\text{hyper}}$ and $\bm{\theta}^*_{\text{hyper}}$.
	Random forest (RF)~\citep{breiman2001random}
	is chosen
	as the surrogate model
	because it has a stronger power 
	for approximating the complex and discrete curvature~\citep{grinsztajn2022tree},
	compared with other common surrogates, \textit{e.g.},
	Gaussian Process (GP)~\citep{williams1996gaussian} or Multi-layer Perceptron (MLP)~\citep{gardner1998artificial}.
	
	\textbf{Acceleration in searching.}
	We adopt a data split trick
	that balances observations and predictions.
	It saves time as the training does not necessarily
	traverse all the training facts.
	Specifically,
	partial training facts are sampled as training queries
	while the others are treated as observations, \textit{i.e.},
	we randomly separate the training facts into two parts as
	$\mathcal{E}^{\text{train}} \! = \! \mathcal{E}^{\text{obs}} \cup \mathcal{E}^{\text{query}}$,
	where the overall prediction system $f_{\bm{\theta}} \circ g_{\bm{\phi}}$
	takes $\mathcal{E}^{\text{obs}}$ as input 
	and then predicts $\mathcal{E}^{\text{query}}$ (see Fig.~\ref{fig: optimization}(c)).
	Here, the split ratio $r^{\text{split}}$ 
	is to balance the sizes of these two parts as 
	$r^{\text{split}} \! = \! 
	\nicefrac{|\mathcal{E}^{\text{obs}}|}{|\mathcal{E}^{\text{query}}|}$.
	Thus, the training becomes
	$
	\bm{\theta}^* \! = \!
	\arg\min_{\bm{\theta}} 
	\Sigma_{(u, q, v) \in \mathcal{E}^{\text{query}}} 
	\mathcal{L}_{cls}\big( f_{\bm{\theta}}(\mathcal{G}_s), v \big)
	$ 
	with the split query edges $\mathcal{E}^{\text{query}}$,
	where
	$\mathcal{G}_s \! = \! g_{\bm{\phi}}(\mathcal{E}^{\text{obs}} , u, q)$
	with the split observation edges $\mathcal{E}^{\text{obs}}$.
	More technical details can be found in the Appendix.~\ref{app:implementation}.
	
	\vspace{-4pt}
	\subsection{Theory: The extrapolation power of one-shot-subgraph link prediction}
	\label{ssec: extrapolation}
	\vspace{-4pt}

	Further, we investigate the extrapolation power of 
	link prediction across graph scales,
	\textit{i.e.}, training and inference on different scales of graphs.
	For example,
	training on small subgraphs $\mathcal{G}^{\text{train}}_s$  
	and testing on large subgraphs $\mathcal{G}^{\text{test}}_s$,
	where the ratio of entities
	$\nicefrac{|\mathcal{V}^{\text{test}}_s|}{|\mathcal{V}^{\text{train}}_s|} \! \gg \! 1$.
	This scenario is practical for recommendation systems on social networks
	that can encounter much larger graphs in the testing phase.
	Intuitively,
	training on smaller $\mathcal{G}^{\text{train}}_s$ 
	can save time for its faster convergence on subgraphs~\citep{shi2023lmc},
	while predicting on larger $\mathcal{G}^{\text{test}}_s$
	might gain promotion for more support of facts in $\mathcal{G}^{\text{test}}_s$.

	
	Nonetheless,
	the Theorem~\ref{theorem:extrapolation} below
	proves that {link prediction} can become unreliable
	as the test graph grows.
	That is,
	if we use a \textit{small} subgraph for training
	and a \textit{large} subgraph for testing,
	the predictor will struggle to 
	give different predictions within and across 
	sampling distributions by $g$, 
	even when these probabilities are arbitrarily different in the underlying graph model.
	Our empirical results in Fig.~\ref{fig:heatmap} 
	support this theoretical finding.
	Hence,
	it is necessary to strike a balance of subgraphs' scale.
	
	\begin{theorem}
		\label{theorem:extrapolation}
		Let $\mathcal{G}^{\text{train}}_s \! \sim \! \mathbb{P}_{\mathcal{G}}$
		and $\mathcal{G}^{\text{test}}_s \! \sim \! \mathbb{P}_{\mathcal{G}}$
		be the training and testing graphs that are sampled from distribution $\mathbb{P}_{\mathcal{G}}$.
		Consider any two test entities $u,v \! \in \! \mathcal{V}_s^{\text{test}}$,
		for which we can make a {prediction} decision of fact $(u, q, v)$
		with the predictor $f_{\bm{\theta}}$, \textit{i.e.}, 
		$\hat{\bm{y}}_{v} \! = \! f_{\bm{\theta}}(\mathcal{G}^{\text{test}}_s)_{v} \! \neq \! \tau$.
		Let $\mathcal{G}^{\text{test}}$ be large enough to satisfy
		$\nicefrac{\sqrt{|\mathcal{V}_s^{\text{test}}|}}{\sqrt{\log(2|\mathcal{V}_s^{\text{test}}|/p)}} \geq \nicefrac{4\sqrt{2}}{d_{\min}}$,
		where $d_{\min}$ is the constant of graphon degree~\citep{diaconis2007graph}.
		Then, for an arbitrary threshold $\tau \! \in \! [0,1]$, 
		the testing subgraph $\mathcal{G}_{s}^{\text{test}}$ satisfies that
		\begin{equation}
			\frac{\sqrt{|\mathcal{V}_s^{\text{test}}|}}{\sqrt{\log(2|\mathcal{V}_s^{\text{test}}|/p)}} 
			\; \geq \;
			\frac{2(C_1 + C_2\lVert g \lVert_{\infty})}{|f_{\bm{\theta}}(\mathcal{G}^{\text{test}}_s)_{v} - \tau| / L(M^{\text{train}})}.
		\end{equation}
		where the underlying generative function 
		of graph signal $g \! \in \! L^{\infty}$
		is with the essential supreme norm
		as in \citep{maskey2022generalization, zhou2022ood}.
		The $p, C_1, C_2$ are constants
		and depend on $M^{\text{train}}$ where
		$\min(supp(|\mathcal{V}_s^{\text{test}}|)) \! \gg \! M^{\text{train}} \! = \! 
		\max(supp(|\mathcal{V}_s^{\text{train}}|))$.
		{It means any test graph can be much larger 
		than the largest possible training graph,
		and $supp$ indicates the support of a distribution.}
		Then, if $u$ and $v$ are isomorphic in topology
		and with the same representations,
		we have a probability at least 
		$1 \! - \! \sum_{\ell=1}^{L}2(|\bm{h}^{(\ell)}| \! + \! 1)p$
		with hidden size $|\bm{h}^{(\ell)}|$
		that the same predictions can be obtained
		whether $u,v$ are generated by the same or distinct $g$.
		The detailed proof can be found in Appendix.~\ref{app:theory}.
	\end{theorem}

	\clearpage
	\section{Experiments}
	\label{sec:exp}

	In this section, we empirically verify the effectiveness 
	of the proposed framework.
	The major experiments are conducted with PyTorch~\citep{paszke2017automatic}
	and one NVIDIA RTX 3090 GPU.
	The OGB datasets are run with one NVIDIA A100 GPU.
	We use five benchmarks 
	with more than ten thousand entities (see Tab.~\ref{tab:dataset}),
	including 
	{WN18RR}~\citep{dettmers2017convolutional}, 
	{NELL-995}~\citep{xiong2017deeppath},
	{YAGO3-10}~\citep{suchanek2007yago},
	OGBL-BIOKG, 
	and OGBL-WIKIKG2~\citep{hu2020open}.


	\begin{table*}[t!]
		\centering
		\caption{Empirical results of {WN18RR, NELL-995, YAGO3-10} datasets.
			Best performance is indicated by the \textbf{bold face} numbers,
			and the \underline{underline} means the second best. 
			``--'' means unavailable results.
			``H@1'' and ``H@10'' are short for Hit@1 and Hit@10 (in percentage), respectively.}
		\label{tab:main-result}
		\vspace{-8px}
		\fontsize{8}{8}\selectfont
		\setlength\tabcolsep{2.8pt}
		\begin{tabular}{cc|ccc|ccc|ccc}
			\toprule
			\multirow{2}{*}{type}  &   \multirow{2}{*}{models}   &  \multicolumn{3}{c|}{{WN18RR}}   &  \multicolumn{3}{c|}{{NELL-995}}    &  \multicolumn{3}{c}{{YAGO3-10}}  \\
			&  &  MRR$\uparrow$     & H@1$\uparrow$     & H@10$\uparrow$   &	MRR$\uparrow$     & H@1$\uparrow$     & H@10$\uparrow$   &MRR$\uparrow$     & H@1$\uparrow$     & H@10$\uparrow$   \\   \midrule
			\multirow{3}{*}{\makecell{Semantic Models}} 
			& ConvE   &  0.427 &  39.2 &  49.8 & 0.511  & 44.6 & 61.9  & 0.520 & 45.0 &  66.0	\\ 
			&  QuatE    	& 0.480&44.0&55.1 &	0.533 & 46.6 & 64.3    & 0.379 & 30.1 & 53.4  \\
			& RotatE   &  0.477 & 42.8 & 57.1  & 0.508 & 44.8 & 60.8 & 0.495 & 40.2 & 67.0  \\
			\midrule
			\multirow{8}{*}{\makecell{Structural Models}} 
			&  MINERVA   &  0.448&41.3&51.3	 & 0.513&41.3&63.7   &  -- & -- & --   \\
			&  DRUM   & 0.486 & 42.5 & 58.6	 & 0.532 &  46.0 & \textbf{66.2}    & 0.531 &  45.3 & 67.6    \\
			&  RNNLogic   &	0.483& 44.6&55.8	&	0.416  &  36.3  & 47.8     & 0.554 &  \underline{50.9} & 62.2   \\
			&  CompGCN  &  0.479 & 44.3 & 54.6	&	0.463  &  38.3	 & 	59.6   &  0.489 & 39.5 & 58.2   \\
			& DPMPN & 0.482 & 44.4 & 55.8 & 0.513 & 45.2 & 61.5 & 0.553 & 48.4 & 67.9 \\
			& NBFNet   & \underline{0.551}   &  \underline{49.7}  &  \textbf{66.6}  &  0.525   &    45.1   &   63.9     &  0.550  &	47.9  &	68.3 \\
			&  RED-GNN   &  0.533 & 48.5 & \underline{62.4} &  \underline{0.543} & \underline{47.6} &  \underline{65.1}  &   \underline{0.559}  &  48.3	&   \underline{68.9}    \\
			\cmidrule{2-11}
			& \textbf{{one-shot-subgraph}}  & \textbf{0.567} & \textbf{51.4} & \textbf{66.6} &  \textbf{0.547} & \textbf{48.5} & \underline{65.1} & \textbf{0.606} & \textbf{54.0} & \textbf{72.1} \\
			\bottomrule
		\end{tabular}
		\vspace{-8px}
	\end{table*}

	\begin{table*}[t!]
		\centering
		\caption{Empirical results of two OGB datasets~\citep{hu2020open} with regard to official leaderboards. 
		}
		\fontsize{8}{8}\selectfont
		\setlength\tabcolsep{1.7pt}
		\label{tab:result-ogb}
		\vspace{-8px}
		\begin{tabular}{cc|ccc|ccc}
			\toprule
			\multirow{2}{*}{type}  &  \multirow{2}{*}{models}   &  \multicolumn{3}{c|}{{OGBL-BIOKG}}   &  \multicolumn{3}{c}{{OGBL-WIKIKG2}} \\ 
			& & Test MRR$\uparrow$ &  Valid MRR$\uparrow$ &  \#Params$\downarrow$ & Test MRR$\uparrow$ &  Valid MRR$\uparrow$ &  \#Params$\downarrow$ \\ \midrule
			\multirow{7}{*}{Semantic Models} 
			& TripleRE & 0.8348 & 0.8360 & 469,630,002 & 0.5794 & 0.6045 & 500,763,337 \\
			& AutoSF & 0.8309 & 0.8317 & 93,824,000 & 0.5458 & 0.5510 & 500,227,800 \\
			& PairRE & 0.8164 & 0.8172 & 187,750,000 &  0.5208 & 0.5423 & 500,334,800 \\
			& ComplEx & 0.8095 & 0.8105  &  187,648,000 & 0.4027 & 0.3759 & 1,250,569,500 \\
			& DistMult & 0.8043 & 0.8055 & 187,648,000 & 0.3729 & 0.3506 & 1,250,569,500 \\
			& RotatE & 0.7989 &	0.7997 &  187,597,000 & 0.4332 & 0.4353 & 1,250,435,750	 \\
			& TransE & 0.7452 &	0.7456 &  187,648,000 & 0.4256  & 0.4272 & 1,250,569,500 \\
			\midrule
			Structural Models & \textbf{{one-shot-subgraph}}  & \textbf{0.8430} & \textbf{0.8435} &  \textbf{976,801} & \textbf{0.6755} &  \textbf{0.7080}  &   \textbf{6,831,201}  \\
			\bottomrule
		\end{tabular}
		\vspace{-8px}
	\end{table*}

	\textbf{Metrics.}
	We adopt the filtered ranking-based metrics for evaluation,
	\textit{i.e.}, mean reciprocal ranking (MRR) and Hit@$k$
	(\textit{i.e.}, both Hit@$1$ and Hit@$10$),
	following~\citep{bordes2013translating,teru2019inductive,wang2017knowledge,zhu2021neural}.
	For both metrics,
	a higher value indicates a better performance.

	\textbf{Main Results.}
	As results shown in 
	Tab.~\ref{tab:main-result} and Tab.~\ref{tab:result-ogb},
	our {one-shot-subgraph link prediction} method
	achieves leading performances
	on all five large-scale benchmarks over all the baselines.
	Especially on the largest OGBL-WIKIKG2 dataset,
	a $16.6\%$ improvement in Test MRR can be achieved.
	Note the results attribute to
	\textit{a deep GNN (high expressiveness)
	and small subgraphs (essential information)}
	extracted by sampling $10\%$ of entities on average
	for answering specific queries.
	Which means,
	it is \textit{unnecessary} to utilize the whole KG in link prediction;
	meanwhile,
	only a small proportion of entities and facts are essential
	for answering specific queries
	that can be quickly identified by the PPR heuristics.
	%
	In what follows,
	we conduct an in-depth analysis 
	of the properties of the proposed method.

\begin{figure}[t!]
		\centering
		\hfill
		{\includegraphics[width=4.2cm]{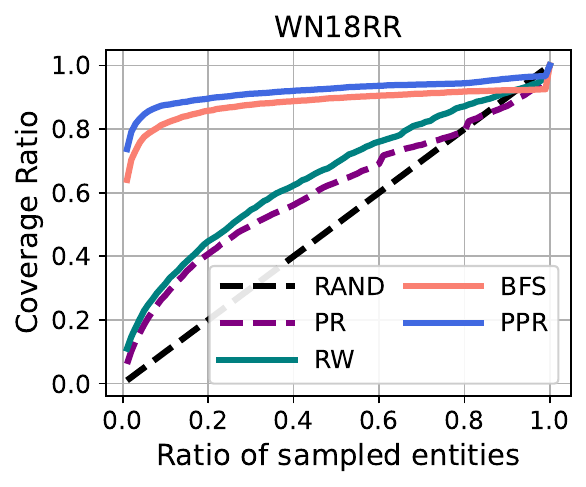}}
		\hfill
		{\includegraphics[width=4.2cm]{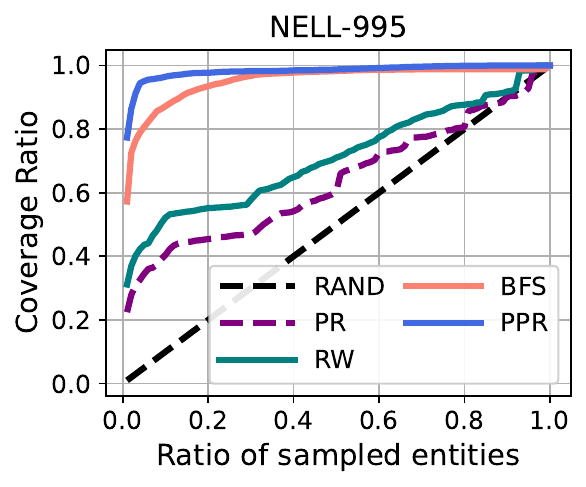}}
		\hfill
		{\includegraphics[width=4.2cm]{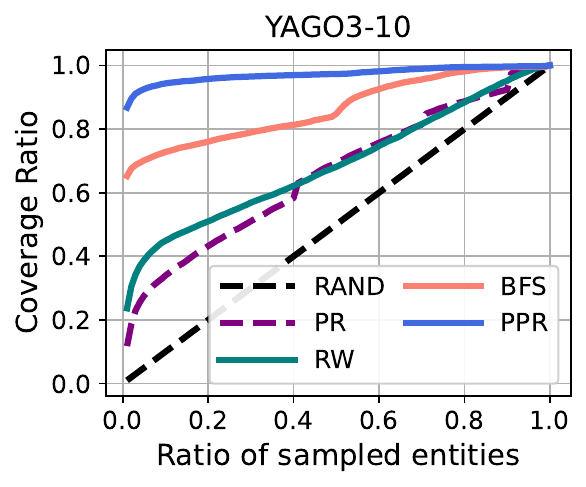}}
		\hfill
		\hfill
		\hfill
		\hfill
		\hfill
		\vspace{-10px}
		\caption{
			Coverage Ratio (CR) of different heuristics
			($\texttt{CR} \! = \!
			\nicefrac{1}{|\mathcal E^{\text{test}}|}
			\sum\nolimits_{(u, q, v)\in \mathcal E^{\text{test}}}
			\mathbb I \{v \! \in \! \mathcal{V}_{s}\}$).
		}
		\label{fig:cover-ratio}
		\vspace{-12px}
	\end{figure}

	\textbf{The Sampling Distribution.}
	We empirically evaluate to what extent
	the entities relevant to a specific query
	can be identified by heuristics, 
	\textit{e.g.}, BFS, RW, and PPR.
	%
	We quantify their power of identifying potential answers 
	via the metric of Coverage Ratio
	$\texttt{CR} \! = \!
	\nicefrac{1}{|\mathcal E^{\text{test}}|}
	\sum\nolimits_{(u, q, v)\in \mathcal E^{\text{test}}}
	\mathbb I \{v \! \in \! \mathcal{V}_{s}\}$,
	\textit{i.e.},
	the ratio of covered answer entities
	that remain 
	in the set of sampled entities $\mathcal{V}_{s}$.
	As shown in 
	Tab.~\ref{tab:cover-ratio} and Fig.~\ref{fig:cover-ratio}, 
	PPR gets a much higher CR and
	notably outperforms other heuristics
	in identifying potential answers.

	\begin{table*}[t!]
		\centering
		\caption{Coverage Ratio of different heuristics.
			\textbf{Bold face} numbers indicate the best results in column.
		}
		\fontsize{8}{8}\selectfont
		\setlength\tabcolsep{1.8pt}
		\label{tab:cover-ratio}
		\vspace{-8px}
		\begin{tabular}{c|ccc|ccc|ccc}
			\toprule
			\multirow{2}{*}{heuristics} &  
			\multicolumn{3}{c|}{{WN18RR}}   &  \multicolumn{3}{c|}{{NELL-995}}    &  \multicolumn{3}{c}{{YAGO3-10}}\\
			&   $r^q_{\mathcal{V}} \! = \! 0.1$     & $r^q_{\mathcal{V}} \! = \! 0.2$     & $r^q_{\mathcal{V}} \! = \! 0.5$ & $r^q_{\mathcal{V}} \! = \! 0.1$     & $r^q_{\mathcal{V}} \! = \! 0.2$     & $r^q_{\mathcal{V}} \! = \! 0.5$ & $r^q_{\mathcal{V}} \! = \! 0.1$     & $r^q_{\mathcal{V}} \! = \! 0.2$     & $r^q_{\mathcal{V}} \! = \! 0.5$  
			\\ \midrule
			Random Sampling (RAND) &  0.100 & 0.200 & 0.500 & 0.100 & 0.200 & 0.500 & 0.100 & 0.200 & 0.500 \\
			PageRank (PR) &  0.278 & 0.407 & 0.633 & 0.405 & 0.454 & 0.603 & 0.340 & 0.432 & 0.694 \\
			Random Walk (RW)   &  0.315 & 0.447 & 0.694 & 0.522 & 0.552 & 0.710 & 0.449 & 0.510 & 0.681 \\
			Breadth-first-searching (BFS)   &   0.818 & 0.858 & 0.898 & 0.872 & 0.935 & 0.982 & 0.728 & 0.760 & 0.848 \\
			Personalized PageRank (PPR)   &   \textbf{0.876} & \textbf{0.896} & \textbf{0.929} & \textbf{0.965} & \textbf{0.977} & \textbf{0.987} & \textbf{0.943} & \textbf{0.957} & \textbf{0.973} \\
			\bottomrule
		\end{tabular}
		\vspace{-8px}
	\end{table*}

	\begin{figure}[t!]
		\centering
		\hfill
		{\includegraphics[width=4.2cm]{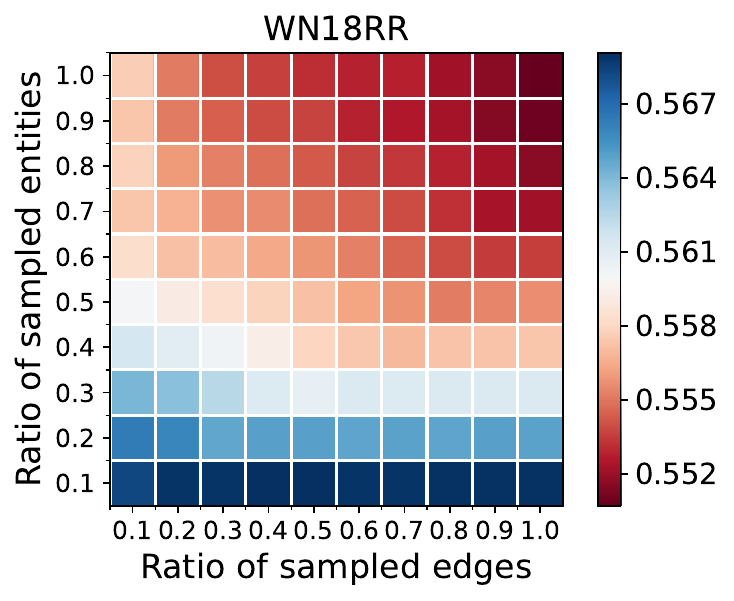}}
		\hfill
		{\includegraphics[width=4.2cm]{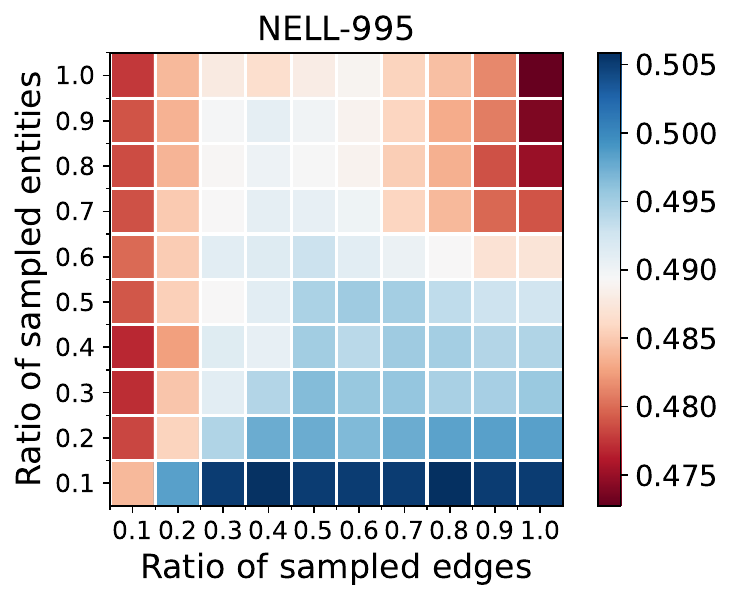}}
		\hfill
		{\includegraphics[width=4.2cm]{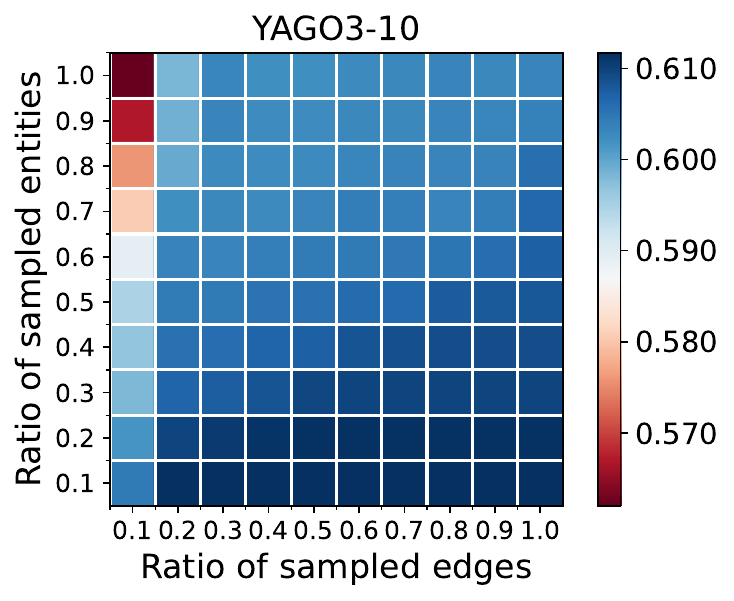}}
		\hfill
		\vspace{-10px}
		\caption{
			Heatmaps of validate MRR (the higher, the better) \textit{w.r.t.}  $r^q_{\mathcal{V}}$ and $r^q_{\mathcal{E}}$ on three benchmarks.
		}
		\label{fig:heatmap}
		\vspace{-8px}
	\end{figure}
	
	\begin{table*}[t!]
		\centering
		\caption{Comparison of effectiveness with regard to subgraph sampling.}
		\label{tab:ablation-layer}
		\vspace{-8px}
		\fontsize{8}{8}\selectfont
		\setlength\tabcolsep{6.8pt}
		\begin{tabular}{ccc|ccc|ccc|ccc}
			\toprule
			\multirow{2}{*}{\#layers ($L$)} & \multirow{2}{*}{$r^q_{\mathcal{V}}$}   & \multirow{2}{*}{$r^q_{\mathcal{E}}$} &  
			\multicolumn{3}{c|}{{WN18RR}}   &  \multicolumn{3}{c|}{{NELL-995}}    &  \multicolumn{3}{c}{{YAGO3-10}}\\
			&   &   & MRR     & H@1     & H@10 & MRR     & H@1     & H@10 & MRR     & H@1     & H@10  \\ \midrule
			8  & 1.0 & 1.0 &  \multicolumn{3}{c|}{Out of memory} &  \multicolumn{3}{c|}{Out of memory}  &  \multicolumn{3}{c}{Out of memory}  \\
			8  & 0.1 & 1.0 & \textbf{0.567} & \textbf{51.4} & \textbf{66.6} &  \textbf{0.547} & \textbf{48.5} & \textbf{65.1} & \textbf{0.606} & \textbf{54.0} & \textbf{72.1} \\
			\midrule
			6  & 1.0 & 1.0 & 0.543 & 49.2 & 64.3 &  0.519 & 45.3 & 62.7 & 0.538 & 46.9 & 66.0 \\
			6  & 0.1 & 1.0 & 0.566 & 51.2 & 66.5 & 0.540 & 48.0 & 63.8 & 0.599 & 53.1  & 71.8 \\
			\midrule
			4  & 1.0 & 1.0 & 0.513 & 46.6  & 59.8 & 0.518 & 45.4 & 61.5 & 0.542 &  47.6 &  66.1 \\
			4  & 0.1 & 1.0 & 0.523 & 47.7 & 60.5 & 0.538 & 47.4 & 63.4 & 0.589 & 52.2 &  70.4  \\
			\bottomrule
		\end{tabular}
		\vspace{-12px}
	\end{table*}

	\textbf{Ablation Study.}
	{\textbf{(1) Training with varying scales and layers:}}
	We train the predictor from scratch with various scales of subgraphs 
	and the number of layers.
	As can be seen from Tab.~\ref{tab:ablation-layer},
	involving all the entities with $r^q_{\mathcal{V}} \! = \! 1.0$
	degenerates the {prediction}, as too many irrelevant entities are covered.
	Besides, a deeper predictor with a larger $L$ 
	consistently brings better results.
	These observations enlighten us to learn a deeper predictor with small subgraphs.
	{\textbf{(2)  Training with different heuristics:}}
	We replace the PPR sampling module with four other common heuristics.
	However, as shown in Tab.~\ref{tab:ablation-sampling-methods},
	their final prediction performances are outperformed by PPR.
	{\textbf{(3) Test-time extrapolation power across scales:}}
	As in Theorem.~\ref{theorem:extrapolation},
	we evaluate the extrapolation power
	by generalizing to various scales of subgraphs
	that are different from the scale of training graphs,
	\textit{e.g.}, the whole graph $r^q_{\mathcal{V}} \! = \! r^q_{\mathcal{E}} \! = \! 1.0$.
	As shown in Fig.~\ref{fig:heatmap}, 
	the predictor also suffers when {prediction} with more irrelevant entities,
	especially with larger $r^q_{\mathcal{V}}$,
	while the generally good cases are 
	a lower $r^q_{\mathcal{V}}$ to focus on the relevant entities
	and a high $r^q_{\mathcal{E}}$ to preserve the local structure (of head $u$) within the sampled entities.

	\textbf{Training and Inference Efficiency.}
	Next, we conduct an efficiency study
	to investigate the improvement of efficiency 
	brought by the proposed {one-shot-subgraph link prediction} framework.
	The running time and GPU memory 
	of an $8$-layer GNN
	are summarized in Tab.~\ref{tab:efficiency}.
	As can be seen,
	a notable advantage of decoupling is that 
	it has less computing cost
	in both terms of less running time and also less memory cost.
	Particularly,
	on the YAGO3-10 dataset,
	the existing GNN-based methods will run out of memory 
	with a deep architecture.
	However,
	with the subgraph sampling 
	of lower ratios of $r^q_{\mathcal{V}}$ and $r^q_{\mathcal{E}}$,
	the learning and {prediction} of GNNs become feasible
	that is with less memory cost and achieving state-of-the-art performance.
	Hence,
	we show that our method
	is effective and also efficient that
	it supports the learning and prediction of deep GNNs on large-scale knowledge graphs.

	Besides, we provide a detailed efficiency comparison 
	between our method (with 10\% entities) 
	and the original implementation (with 100\% entities) 
	on two SOTA methods, NBFNet and RED-GNN.
	Tab.~\ref{tab:efficiency-two-GNNs} shows 
	that the training time is significantly reduced 
	when learning with our method.
	Notably, on the YAGO3-10 dataset, 
	$94.3\%$ and $94.5\%$ of training time (for one epoch) 
	can be saved for NBFNet and RED-GNN, respectively. 
	Besides, 
	our method boosts the performance
	{as advantage~\ref{advantage: scope}},
	where the performance improvement 
	can come from a deeper GNN and a smaller observation graph
	(detailed analysis in Appendix.~\ref{app:explanation}).
	Full evaluations and more discussions
	are elaborated in the Appendix.~\ref{app:evaluation}.

	\begin{figure*}[t!]
		\centering
		\hfill
		\includegraphics[width=6.8cm]{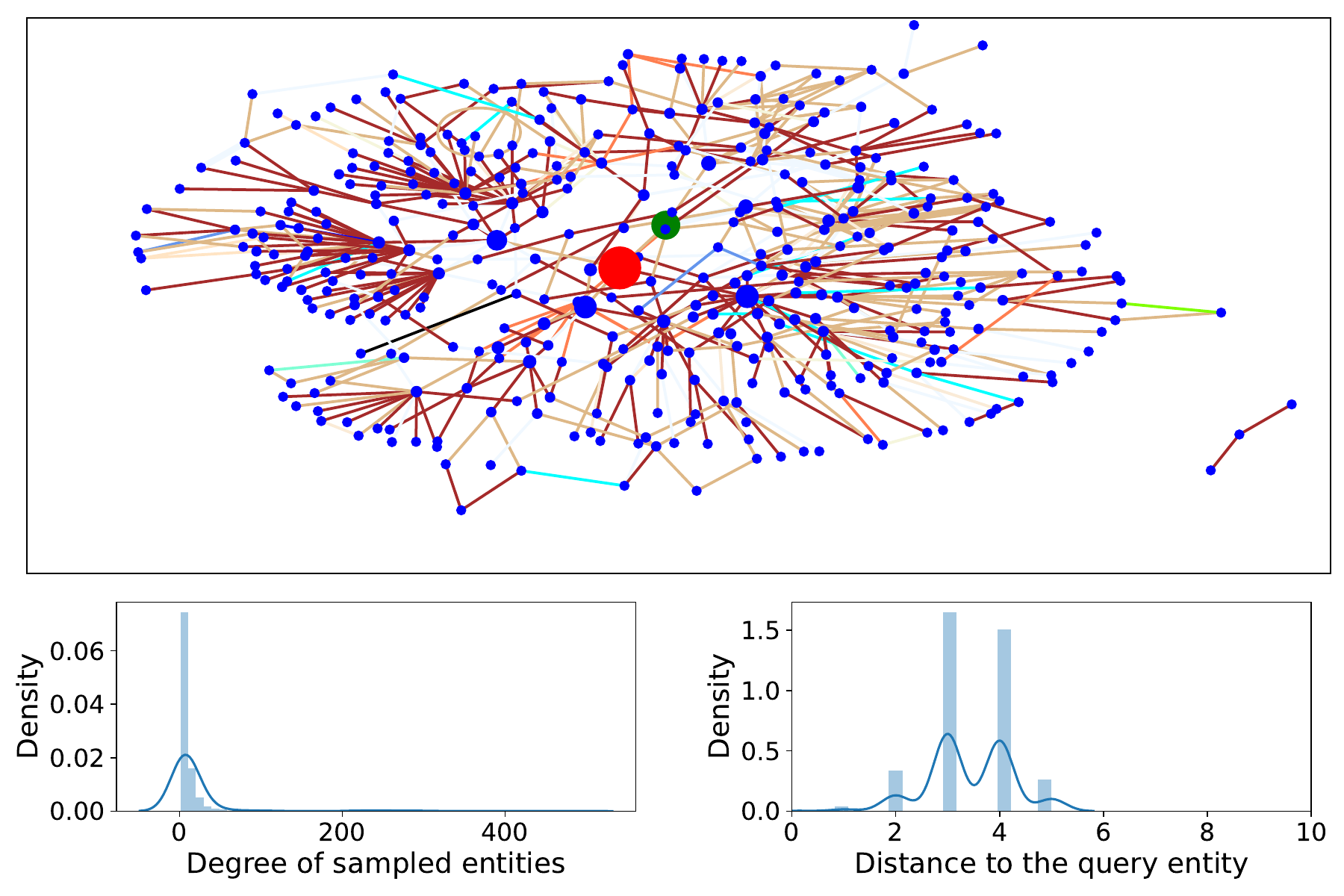}
		\hfill
		\includegraphics[width=6.8cm]{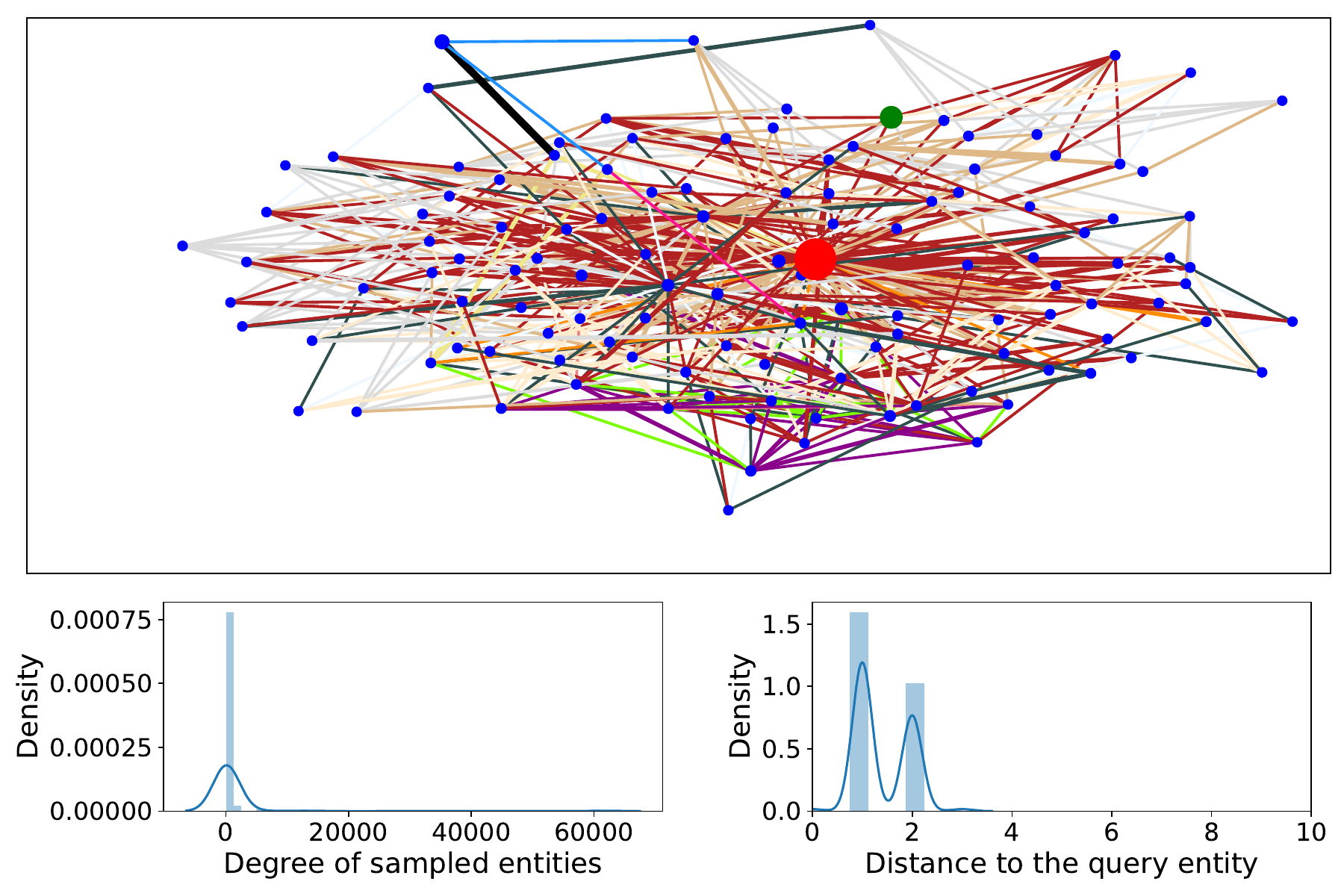}
		\hfill
		\vspace{-8px}
		\caption{
			Exemplar subgraphs sampled from WN18RR (left) and YAGO3-10 (right).
			The red and green nodes indicate 
			the query entity and answer entity.
			The colors of the edges indicate relation types.
			The bottom distributions of degree and distance
			show the statistical properties of each subgraph.
		}
		\label{fig: case study}
		\vspace{-4px}
	\end{figure*}
	
	\begin{table}[t!]
		\centering
		\caption{Comparison of {prediction} performance with different sampling heuristics.}
		\label{tab:ablation-sampling-methods}
		\vspace{-10px}
		\fontsize{8}{8}\selectfont
		\setlength\tabcolsep{12.1pt}
		\begin{tabular}{c|ccc|ccc}
			\toprule
			\multirow{2}{*}{heuristics} &   \multicolumn{3}{c|}{{WN18RR}}   &  \multicolumn{3}{c}{{YAGO3-10}} \\
			& MRR     & H@1 &  H@10 & MRR    & H@1 & H@10 \\ \midrule
			Random Sampling (RAND) & 0.03 & 43.4 & 3.5 & 0.057 & 5.1 & 6.5 \\
			PageRank (PR) & 0.124 & 11.5 & 14.2 & 0.315 & 28.9 & 35.9 \\
			Random Walk (RW) & 0.507 & 45.8 & 59.8 & 0.538 & 46.3 & 67.2 \\
			Breadth-first-searching (BFS) & 0.543 & 49.6 & 63.0 & 0.562 &  49.4 & 69.0 \\
			\textbf{Personalized PageRank (PPR)} & \textbf{0.567} & \textbf{51.4} & \textbf{66.6} &  \textbf{0.606} & \textbf{54.0} & \textbf{72.1} \\
			\bottomrule
		\end{tabular}
		\vspace{-6px}
	\end{table}

	\begin{table*}[t!]
		\centering
		\caption{Comparison of {prediction} performance with two recent GNN methods.}
		\label{tab:efficiency-two-GNNs}
		\vspace{-10px}
		\fontsize{8}{8}\selectfont
		\setlength\tabcolsep{2.7pt}
		\begin{tabular}{c|cccc|cccc}
			\toprule
			\multirow{2}{*}{methods} &   \multicolumn{4}{c|}{{WN18RR}}   &  \multicolumn{4}{c}{{YAGO3-10}} \\
			& MRR     & H@1     & H@10 &  Time & MRR     & H@1     & H@10 & Time \\ \midrule
			NBFNet (100\% entities) & 0.551 & 49.7 & 66.6 & 32.3 min & 0.550 & 47.9 & 68.3 & 493.8 min \\
			NBFNet \textbf{+ {one-shot-subgraph} (10\% entities)} & \textbf{0.554} & \textbf{50.5} & \textbf{66.3} & \textbf{2.6 min} & \textbf{0.565} & \textbf{49.6} & \textbf{69.2} & \textbf{28.2 min} \\
			RED-GNN (100\% entities) & 0.533 & 48.5 & 62.4 & 68.7 min & 0.559 & 48.3 & 68.9 & 1382.9 min \\
			RED-GNN \textbf{+ {one-shot-subgraph} (10\% entities)} & \textbf{0.567} & \textbf{51.4} & \textbf{66.6} & \textbf{4.5 min} & \textbf{0.606} & \textbf{54.0} & \textbf{72.1} & \textbf{76.3 min} \\
			\bottomrule
		\end{tabular}
		\vspace{-6px}
	\end{table*}

	\begin{table*}[t!]
		\centering
		\caption{ 
			Comparison of efficiency 
			with an $8$-layer predictor
			and different $r^q_{\mathcal{V}}, r^q_{\mathcal{E}}$.
		}
		\fontsize{8}{8}\selectfont
		\setlength\tabcolsep{11.0pt}
		\label{tab:efficiency}
		\vspace{-8px}
		\begin{tabular}{ccc|cc|cc|cc}
			\toprule
			\multirow{2}{*}{phase}  &   \multirow{2}{*}{$r^q_{\mathcal{V}}$}   & \multirow{2}{*}{$r^q_{\mathcal{E}}$} &  
			\multicolumn{2}{c|}{{WN18RR}}   &  \multicolumn{2}{c|}{{NELL-995}}    &  \multicolumn{2}{c}{{YAGO3-10}}   \\
			&   &   & Time & Memory & Time & Memory  & Time & Memory \\ \midrule
			\multirow{6}{*}{Training} 
			& 1.0 & 1.0 & \multicolumn{2}{c|}{Out of memory} & \multicolumn{2}{c|}{Out of memory} & \multicolumn{2}{c}{Out of memory}  \\
			& 0.5 & 0.5 & 26.3m & 20.3GB & 1.6h &  20.1GB & \multicolumn{2}{c}{Out of memory}  \\
			\cmidrule{2-9}
			& 0.2 & 1.0 & 12.8m & 20.2GB & 1.2h &  18.5GB & \multicolumn{2}{c}{Out of memory}  \\
			& 0.2 & 0.2 & 6.7m & 6.4GB & 0.6h &  8.9GB & 2.1h & 23.1GB \\
			\cmidrule{2-9}
			& 0.1 & 1.0 & 7.2m & 9.8GB  &  0.8h & 12.1GB  & 1.3h &  13.9GB \\
			& 0.1 & 0.1 &6.6m & 5.1GB & 0.3h & 5.3GB & 0.9h & 10.2GB \\
			\midrule
			\multirow{6}{*}{Inference} 
			& 1.0 & 1.0 & 7.3m & 6.7GB & 17.5m & 12.8GB & 1.6h & 15.0GB  \\
			& 0.5 & 0.5 & 6.0m & 4.3GB & 8.3m & 4.5GB  & 1.1h & 10.1GB  \\
			\cmidrule{2-9}
			& 0.2 & 1.0 & 3.2m & 5.8GB & 4.2m &  12.1GB & 0.7h & 14.7GB  \\
			& 0.2 & 0.2 & 2.8m & 1.9GB & 3.6m & 2.5GB  & 0.6h & 3.7GB \\
			\cmidrule{2-9}
			& 0.1 & 1.0 & 2.7m & 2.7GB & 3.1m & 9.4GB  & 0.4h &  9.7GB \\
			& 0.1 & 0.1 & 2.3m & 1.7GB & 2.9m & 1.9GB  & 0.4h &  3.1GB \\
			\bottomrule
		\end{tabular}
		\vspace{-14px}
	\end{table*}

	
	

	\textbf{Case Study.}
	We visualize the sampled subgraph in Fig.~\ref{fig: case study}
	with the histograms of degree and distance distributions.
	As can be seen,
	the local structure of query entity $u$ is well preserved,
	while the true answers $v$ are also covered in the subgraphs.
	More cases and analyses can be found in Appendix.~\ref{app:case}.

	%
	
	\vspace{-8px}
	\section{Conclusion}
	\label{sec:conclude}
	\vspace{-6px}

	In this paper, 
	we propose the
	one-shot-subgraph link prediction
	to alleviate the scalability problem of structural methods
	and achieve efficient as well as adaptive learning on large-scale KGs.
	We 
	discover that 
	the non-parametric and computation-efficient heuristics PPR
	can effectively identify the potential answers and support to the {prediction}.
	We further introduce 
	the automated searching for adaptive configurations
	in both data space and model space.
	%
	Extensive experiments on five large-scale benchmarks
	verify the effectiveness and efficiency of our method.
	Importantly,
	we show it unnecessary to utilize the whole KG 
	for answering specific queries;
	meanwhile,
	only a small proportion of information is essential
	and can be identified by the PPR heuristics without learning.
	
	\clearpage
    
	\section*{Acknowledgments}
	ZKZ and BH were supported by the NSFC General Program No. 62376235, Guangdong Basic and Applied Basic Research Foundation Nos. 2022A1515011652 and 2024A1515012399, HKBU Faculty Niche Research Areas No. RC-FNRA-IG/22-23/SCI/04, and HKBU CSD Departmental Incentive Scheme.
	JCY was supported by 111 plan (No. BP0719010) and National Natural Science Foundation of China (No. 62306178).
	QMY was in part supported by NSFC (No. 92270106) and National Key Research and Development Program of China under Grant 2023YFB2903904.
	The authors thank Haobo Xu for his assistance in experiments.
	
	\section*{Ethics Statement}
	We would claim that
	this work does not raise any ethical concerns. 
	Besides, this work does not involve any human subjects, 
	practices to data set releases, potentially harmful insights, 
	methodologies and applications, potential conflicts of interest 
	and sponsorship, discrimination/bias/fairness concerns, privacy 
	and security issues, legal compliance, and research integrity issues.
	
	\section*{Reproducibility Statement}
	The experimental setups for training and evaluation 
	are described in detail in Sec.~\ref{sec:exp} and Appendix.~\ref{app:evaluation}, 
	and the experiments are all conducted using public datasets. 
	The code is publicly available at:
	\url{https://github.com/tmlr-group/one-shot-subgraph}.
	
	\bibliography{acmart}
	\bibliographystyle{iclr2024_conference}
	
	\clearpage
	\appendix
	
	\addcontentsline{toc}{section}{Appendix} 
	\part{Appendix} 
	\parttoc 
	
	\clearpage
	\section{Theoretical Analysis}
	\label{app:theory}
	
	\noindent
	\subsection{Notations}
	
	We summarize the frequently used notations in Tab.~\ref{tab:notations}.
	%

	\begin{table*}[ht]
		\centering
		\caption{The most frequently used notations in this paper.}
		\label{tab:notations}
		\vspace{-8px}
		\begin{tabular}{c|c}
			\toprule
			notations  & meanings \\
			\midrule
			$\mathcal{V}, \mathcal R, \mathcal E$ & the set of entities, relations, facts (edges) of the original KG \\  
			\midrule
			$\mathcal{G} = (\mathcal{V}, \mathcal{R}, \mathcal{E})$ & the original KG \\
			\midrule
			$\mathcal{V}_s, \mathcal R_s, \mathcal E_s$ & the set of entities, relations, facts (edges) of the sampled KG \\  
			\midrule
			$\mathcal{G}_s = (\mathcal{V}_s, \mathcal{R}_s, \mathcal{E}_s)$ & the sampled KG generated by the sampler $g_{\bm{\phi}}$ \\
			\midrule
			$(x, r, o)$  & a fact triplet in $\mathcal E$ with subject entity $x$, relation $r$ and object entity $o$ \\
			\midrule
			$(u, q, v)$ & a query triple with query entity $u$, query relation $q$ and answer entity $v$ \\
			\midrule
			$L$ & the total number of propagation steps \\
			\midrule
			$\ell$  & the $\ell$-th propagation step and $\ell \in \{0\dots L\}$ \\
			\midrule
			$K$ &   the maximum steps of updating the PPR scores \\
			\midrule
			$k$  & the $k$-th propagation step and $k \in \{0\dots K\}$ \\
			\midrule
			$\bm h_{o}^{(\ell)}$ & the representation of entity $o$ at step $\ell$ \\
			\midrule
			$g_{\bm{\phi}}$ & the sampler of the {one-shot-subgraph link prediction} framework \\
			\midrule
			$f_{\bm{\theta}}$ & the predictor of the {one-shot-subgraph link prediction} framework \\
			\midrule
			$r^q_{\mathcal{V}}, r^q_{\mathcal{E}}$ & the sampling ratios of entities and edges \\ 
			\bottomrule
		\end{tabular}
	\end{table*}

	\subsection{Complexity Analysis}
	
	Next, we compare three different {manners}
	from perspectives of computation and parameter.

	In a nutshell,
	semantic models 
	are computation-efficient but parameter-expensive,
	while
	structural models 
	are parameter-efficient but computation-expensive.
	Here, we aim to make the best of the both worlds
	by designing a framework that is
	parameter-efficient and computation-efficient.
	The proposed {one-shot-subgraph link prediction} models,
	as will be elaborated in Sec.~\ref{sec:one-shot-subgraph},
	is with the new {prediction} {manner} that
	$\mathcal{G} \xmapsto{g_{\bm{\phi}}, (u_q, r_q)} \mathcal{G}_s \xmapsto{f_{\bm{\theta}}} \hat{\bm{Y}}$.
	The key difference here is that,
	instead of the original $\mathcal{G}$,
	the predictor $f_{\bm{\theta}}$ is acting on the query-dependent subgraph $\mathcal{G}_s$
	extracted by the sampler $g_{\bm{\phi}}$.
	
	By contrast, 
	we show that
	the proposed subgraph models
	make the best of both worlds:
	(1) subgraph models are computation-efficient,
	as the extracted subgraph 
	is much smaller than the original graph, \textit{i.e.}, 
	$|\mathcal{V}_s| \! \ll \!     |\mathcal{V}|$ and
	$|\mathcal{E}_s| \! \ll \!     |\mathcal{E}|$;
	(2) inherits from~\citet{sadeghian2019drum, zhu2021neural, zhang2021knowledge},
	subgraph models are also parameter-efficient:
	only requires the relations' embeddings
	but not the expensive entities' embeddings.
	By contrast,
	the semantic models need to learn entities's embeddings,
	which are parameter-expensive and only applicable in transductive settings.
	
	A detailed comparison of parameter and computation complexity
	is summarized in Tab.~\ref{tab:related work comparison}.
	
	\begin{figure}[ht]
		\centering
		\subfigure[semantic model]
		{\includegraphics[height=2.0cm]{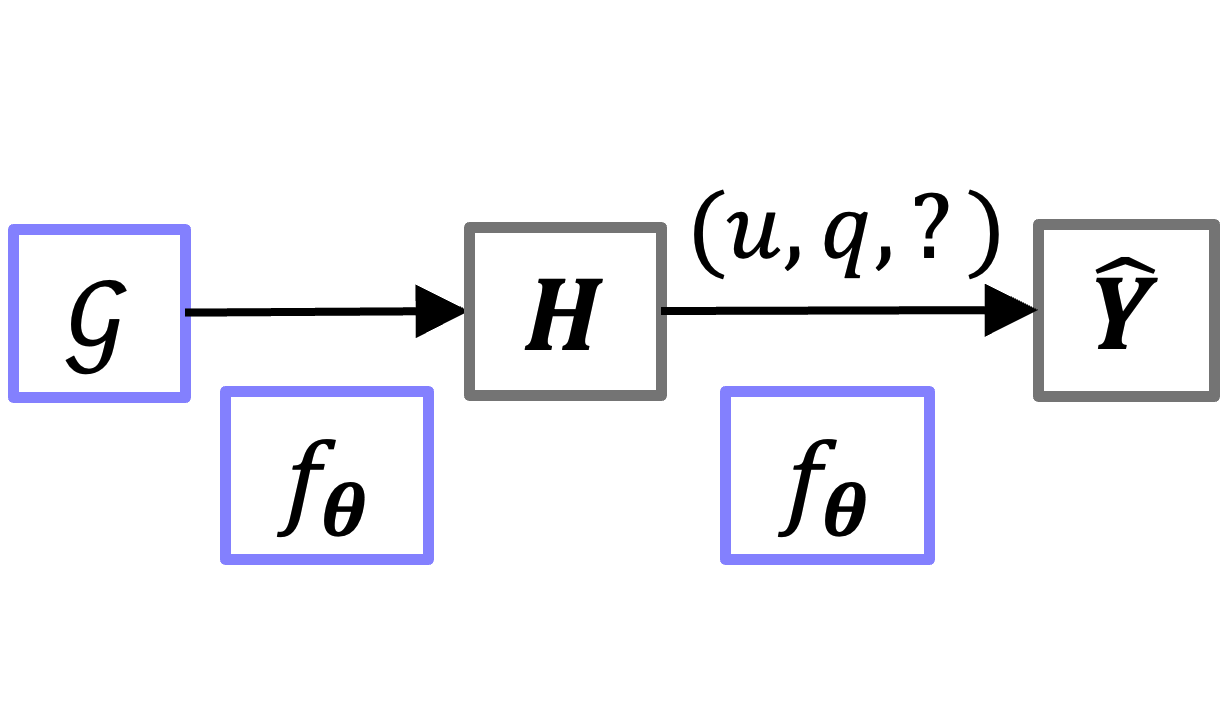}}
		\hfill
		\subfigure[structural model]
		{\includegraphics[height=2.0cm]{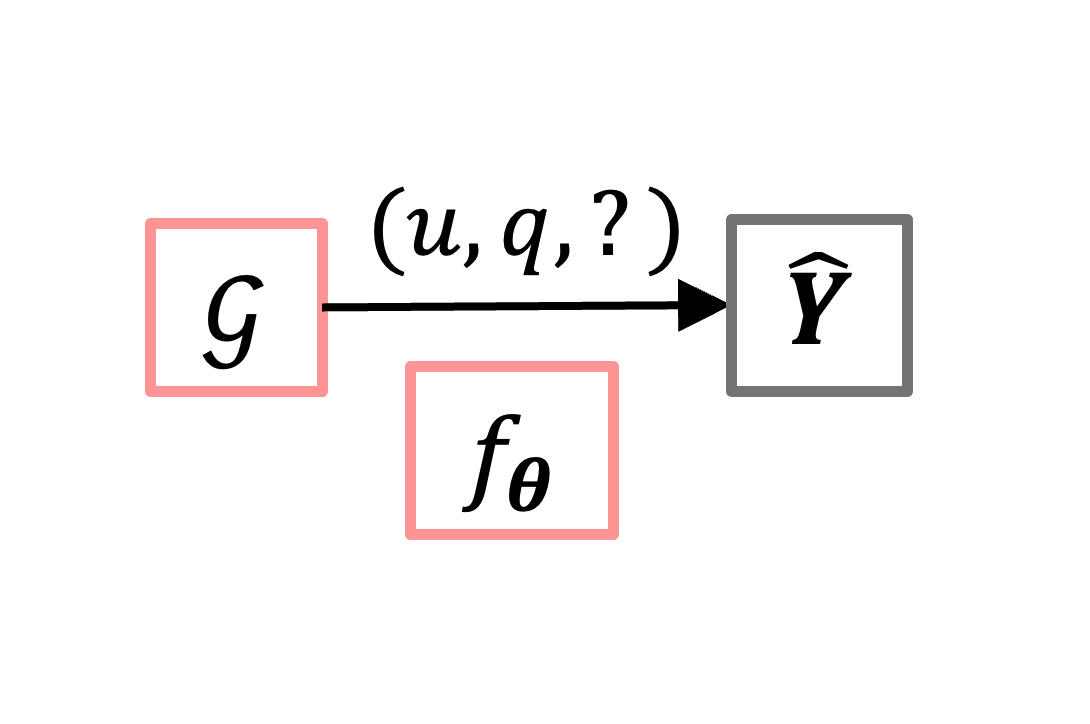}}
		\hfill
		\subfigure[one-shot-subgraph model]
		{\includegraphics[height=2.1cm]{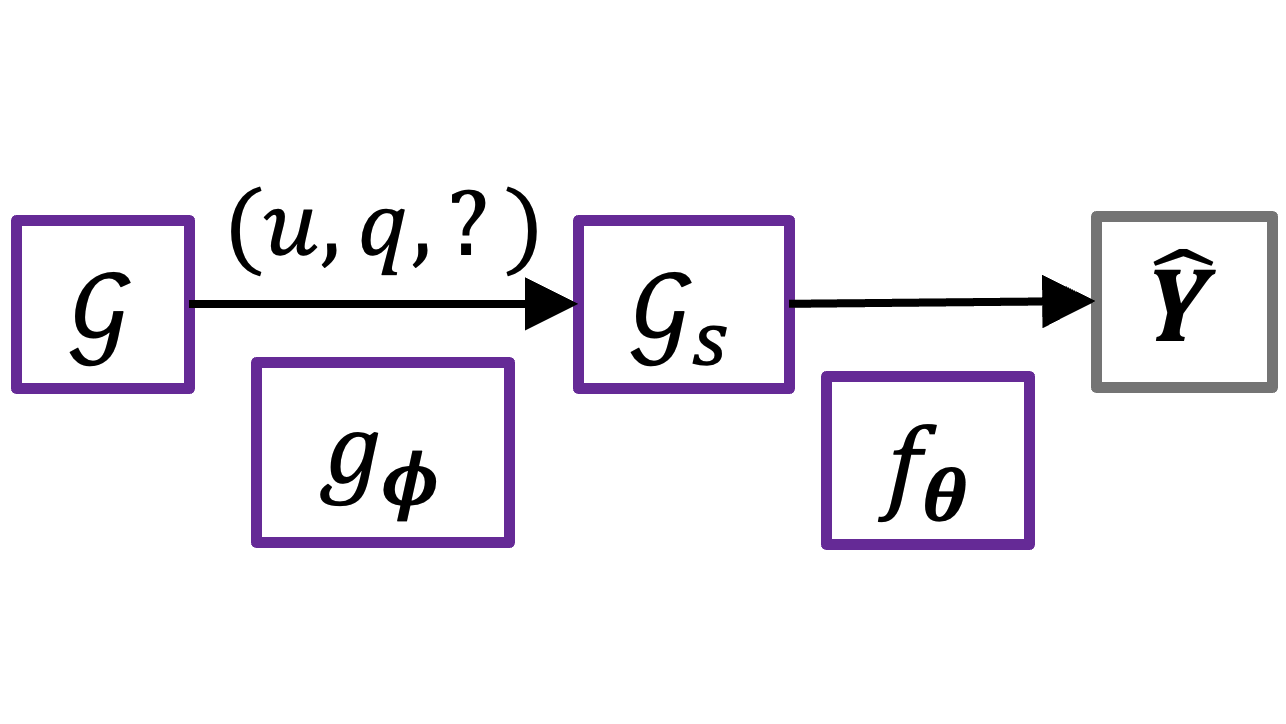}}
		\hfill
		\subfigure[complexity comparison]
		{\includegraphics[height=2.0cm]{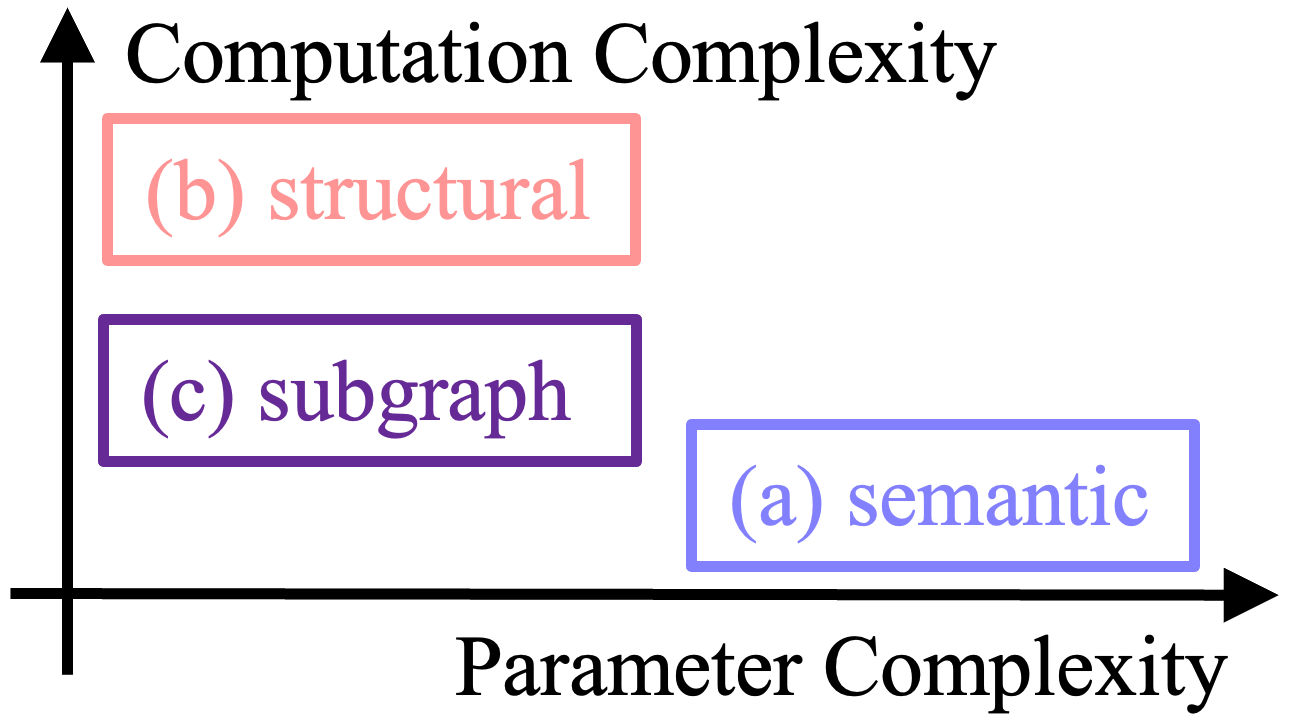}}
		\vspace{-8px}
		\caption{
			Illustrations of {prediction} {manners}.
			Semantic and structure models 
			implicitly or explicitly 
			take the whole graph $\mathcal{G}$ for {prediction}.
			Our \textit{one-shot-subgraph} model
			\textit{only requires one subgraph $\mathcal{G}_s$ for the prediction of one query}
			that decreases computation complexity
			via decoupling $f_{\bm{\theta}}$ and $\mathcal{G}$.
		}
	\end{figure}

	\begin{figure}[ht]
		\centering
		\subfigure[layer-wise sampling]
		{\includegraphics[height=2.8cm]{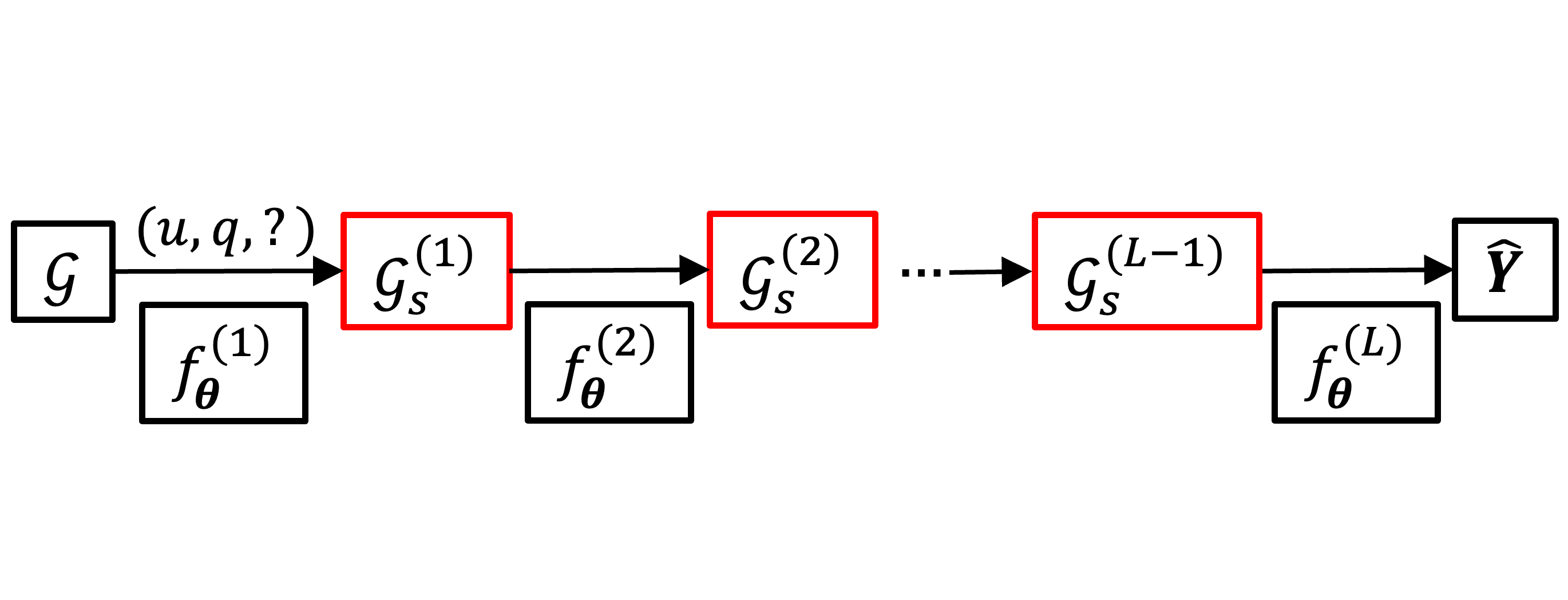}}
		\hfill
		\subfigure[subgraph-wise sampling]
		{\includegraphics[height=2.8cm]{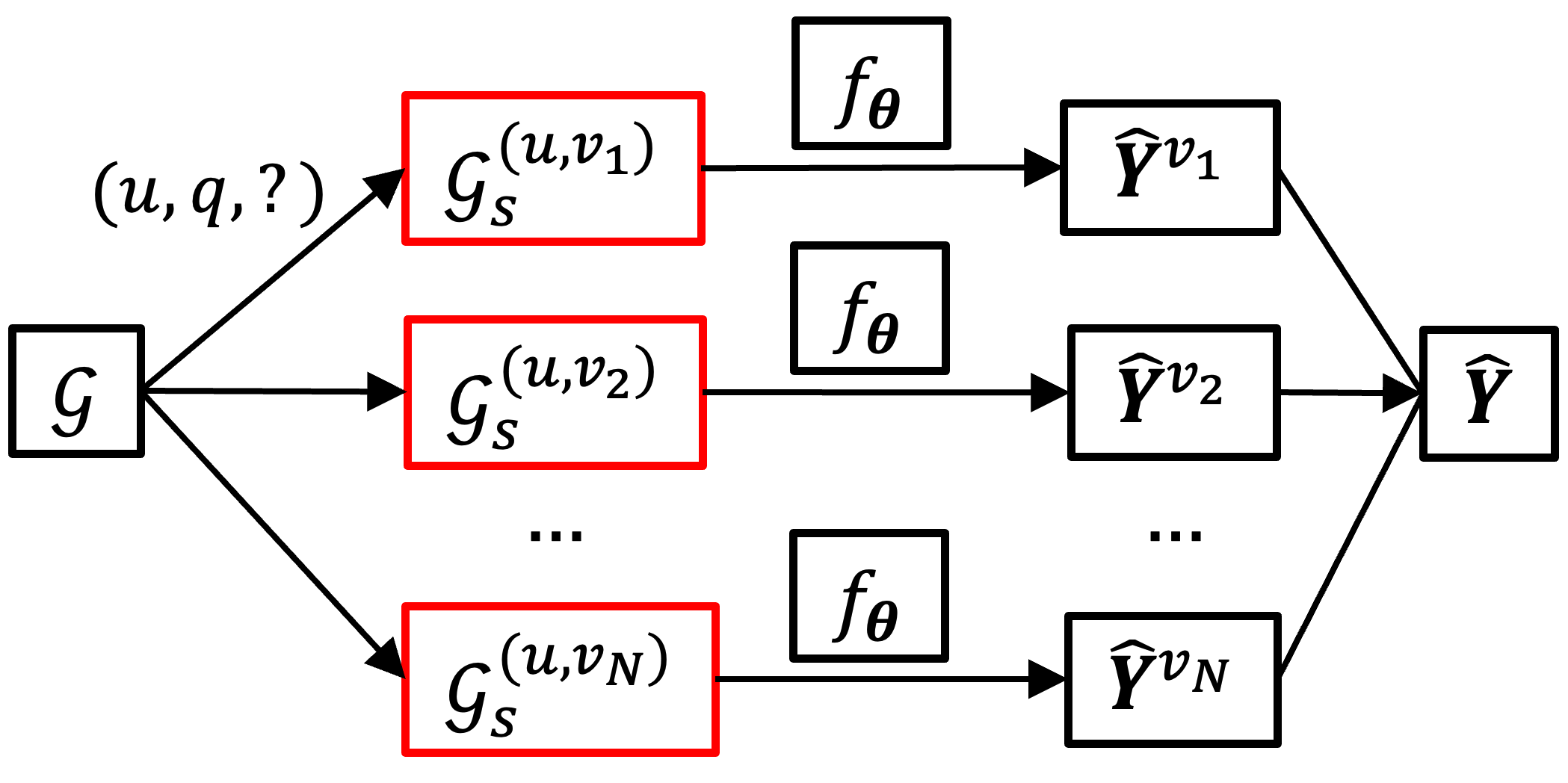}}
		\vspace{-8px}
		\caption{
			Illustrations of sampling methods.
			Detailed analysis is corresponding to Sec.~\ref{sec:one-shot-subgraph}.
		}
	\end{figure}

	\begin{table}[ht]
		\centering
		\caption{
			The comparison of related works with
			parameter complexity and computation complexity.
		}
		\vspace{-8px}
		\label{tab:related work comparison}
		\begin{tabular}{c|cc}
			\toprule
			Category &  Parameter complexity & Computation complexity \\ \midrule
			semantic models & ${O}(|\mathcal{V}| \! \cdot \! D_{\mathcal{V}}) \! + \! {O}(|\mathcal{R}| \! \cdot \! D_{\mathcal{R}})$ & ${O}(|\mathcal{R}| \! \cdot \! D_{\mathcal{V}})$ \\
			structural models & ${O}(|\mathcal{R}| \! \cdot \! D_{\mathcal{R}})$ &  ${O}(|\mathcal{E}| \! \cdot \! D_{\mathcal{R}}  \! \cdot \! L)$ \\
			subgraph models & ${O}(|\mathcal{R}| \! \cdot \! D_{\mathcal{R}})$ & ${O}(|\mathcal{E}| \! \cdot \! K) \! + \! {O}(|\mathcal{E}_s| \! \cdot \! D_{\mathcal{R}} \! \cdot \! L)$ \\
			\bottomrule
		\end{tabular}
	\end{table}

	\subsection{Proof for Theorem~\ref{theorem:extrapolation}}
	\begin{proof}
		
		%
		%
		
		Here,
		let $\mathcal{G}^{\text{train}}_s \! \sim \! \mathbb{P}_{\mathcal{G}}$
		and $\mathcal{G}^{\text{test}}_s \! \sim \! \mathbb{P}_{\mathcal{G}}$
		be the training and testing graphs that are sampled distribution $\mathbb{P}_{\mathcal{G}}$.
		Let $\mathcal{G}^{\text{test}}$ be large enough to satisfy
		$\nicefrac{\sqrt{|\mathcal{V}_s^{\text{test}}|}}{\sqrt{\log(2|\mathcal{V}_s^{\text{test}}|/p)}} \geq \nicefrac{4\sqrt{2}}{d_{\min}}$,
		where $d_{\min}$ is the constant of graphon degree~\citep{diaconis2007graph}.
		
		In this part,
		we follow~\citet{maskey2022generalization, zhou2022ood}
		with the definitions 
		of 
		graph message-passing neural network $\Phi$
		and 
		continuous message passing neural network $\Psi$.
		Note that $\Phi$ is equivalent to the message passing function
		of the predictor $f_{\bm w}$
		that generates the entity representation $\bm{h}_v^{(\ell)}$
		in each layer $\ell=1,2,\cdots,L$.
		Besides,
		$\Psi$
		is the continuous counterpart of
		$\Phi$.
		It uses the continuous aggregation
		instead of the discrete one 
		w.r.t. the discrete graph topology
		in $\Phi$.
		
		With the two above functions,
		$\lVert \Phi \lVert_{\infty}, \lVert \Psi \lVert_{\infty}$ 
		can be trained and determined by 
		the original graph $\mathcal{G}^{\text{train}}$ 
		and hyperparameters $r^q_{\mathcal{V}}, r^q_{\mathcal{E}}$ and $L$.
		Following~\citet{maskey2022generalization, zhou2022ood},
		let $p \in (0, \nicefrac{1}{\sum_{\ell=1}^{L}2|\bm h^{(\ell)}|  + 1 })$
		we with probability at least
		$1 \! - \! \sum_{\ell=1}^{L}2(|\bm{h}^{(\ell)}| \! + \! 1)p$,
		the difference $\delta$
		between $\lVert \Phi \lVert_{\infty}$
		and
		$\lVert \Psi \lVert_{\infty}$ 
		is bounded by
		\begin{align}
			\begin{split}
				\delta :=&
				\max_{i = 1,2,\cdots, |\mathcal{V}_s^{\text{test}}|}
				\Vert \Phi _i - \Psi_i \Vert_{\infty}
				\\
				\leq&
				\sum_{\ell=1}^{L} L_{\Psi}^{(\ell)}(M^{\text{train}})
				\Big( 2\sqrt{2} 
				\frac{
					(L_{\Phi}^{(\ell)}(M^{\text{train}}) \lVert g \lVert_{\infty} + \lVert \Phi^{(\ell)} \lVert_{\infty})
					\sqrt{\log(\nicefrac{2|\mathcal{V}_s^{\text{test}}|}{p})}
				}
				{\sqrt{|\mathcal{V}_s^{\text{test}}|}} \Big)
				\\&
				\times
				\prod_{\ell'=\ell+1}^{L}
				\Big(
				(L_{\Psi}^{(\ell')}(M^{\text{train}}))^2 + 2(L_{\Phi}^{(\ell')}(M^{\text{train}}))^2 (L_{\Psi}^{(\ell')}(M^{\text{train}}))^2
				\Big).
			\end{split}
			\nonumber
		\end{align}
		
		Then, based on the proof for Lemma B.9 in~\citep{maskey2022generalization},
		we can derive
		$\lVert g \lVert_{\infty} \leq B_1^{(\ell)}+ B_2^{(\ell)}\lVert g \lVert_{\infty}$,
		where $B_1^{(\ell)}, B_2^{(\ell)}$ are independent of $g$.
		Specifically,
		\begin{align}
			\begin{split}
				B_1^{(\ell)}=&
				\sum_{k=1}^{\ell}
				\Big(
				L_{\Psi}^{(k)}(M^{\text{train}})\lVert \Phi^{(k)} \lVert_{\infty} 
				+ \lVert \Psi^{(k)} \lVert_{\infty}
				\Big)
				\times \prod_{k'=k+1}^{\ell}
				\Big(
				L_{\Psi}^{(k')}(M^{\text{train}})
				(1 + L_{\Psi}^{(k')}(M^{\text{train}}))
				\Big),
				\\
				B_2^{(\ell)}=&
				\prod_{k=1}^{\ell}
				L_{\Psi}^{(k)}(M^{\text{train}})
				\Big( 1 + L_{\Phi}^{(k)}(M^{\text{train}}) \Big).
			\end{split}
			\nonumber
		\end{align}
		
		Then, we define constants $C_1, C_2$ as follows.
		\begin{align}
			\begin{split}
				C_1 =
				\sum_{\ell=1}^{L}
				L_{\Psi}^{(\ell)}(M^{\text{train}})
				\Big( 
				2\sqrt{2} (L_{\Phi}^{(\ell)}(M^{\text{train}}) B_1^{(\ell)}
				+ \Vert \Phi^{(\ell+1)} \lVert_{\infty})
				\Big)
				\\ \times
				\prod_{\ell'=\ell+1}^{L}
				\Big( 
				(L_{\Psi}^{(\ell')}(M^{\text{train}}))^2 + 2(L_{\Phi}^{(\ell')}(M^{\text{train}}))^2 (L_{\Psi}^{(\ell')}(M^{\text{train}}))^2
				\Big),
			\end{split}
			\nonumber
		\end{align}
		\begin{align}
			\begin{split}
				C_2 =
				\sum_{\ell=1}^{L}
				L_{\Psi}^{(\ell)}(M^{\text{train}})
				\Big( 
				2\sqrt{2} 
				L_{\Phi}^{(\ell)}(M^{\text{train}}) B_2^{(\ell)}
				\Big)
				\\ \times
				\prod_{\ell'=\ell+1}^{L}
				\Big(
				(L_{\Psi}^{(\ell')}(M^{\text{train}}))^2 + 2(L_{\Phi}^{(\ell')}(M^{\text{train}}))^2 (L_{\Psi}^{(\ell')}(M^{\text{train}}))^2
				\Big).
			\end{split}
			\nonumber
		\end{align}
		
		This, we can derive the difference $\delta$ as
		\begin{align}
			\begin{split}
				\delta 
				:= \max_{i = 1,2,\cdots, |\mathcal{V}_s^{\text{test}}|}
				\Vert \Phi _i - \Psi_i \Vert_{\infty} 
				\leq
				(C_1 + C_2\lVert g \lVert_{\infty})
				\frac{\sqrt{\log(\nicefrac{2|\mathcal{V}_s^{\text{test}}|}{p})}}
				{\sqrt{|\mathcal{V}_s^{\text{test}}|}},
			\end{split}
			\nonumber
		\end{align}
		where $C_1, C_2$ depends on 
		$\{ L_{\Phi}^{(\ell)}(M^{\text{train}}) \}_{\ell=1}^L$
		and
		$\{ L_{\Psi}^{(\ell)}(M^{\text{train}}) \}_{\ell=1}^L$
		that
		\begin{align}
			\begin{split}
				\min(supp(|\mathcal{V}_s^{\text{test}}|)) \gg M^{\text{train}} = 
				\max(supp(|\mathcal{V}_s^{\text{train}}|)).
			\end{split}
			\nonumber
		\end{align}

		Then,
		consider any two test entities $u,v \! \in \! \mathcal{V}_s^{\text{test}}$,
		for which we can make a {prediction} decision of fact $(u, q, v)$,
		we have
		\begin{align}
			\begin{split}
				\Vert  \Phi_u - \Phi_v \Vert_{\infty} 
				\leq&
				\Vert  \Phi_u - \Psi_u \Vert_{\infty} + \Vert  \Psi_u - \Phi_v \Vert_{\infty}  \\
				=&
				\Vert  \Phi_u - \Psi_u \Vert_{\infty} + \Vert  \Psi_v - \Phi_v \Vert_{\infty}  
				\leq
				(C_1 + C_2\lVert g \lVert_{\infty})
				\frac{\sqrt{\log(\nicefrac{2|\mathcal{V}_s^{\text{test}}|}{p})}}
				{\sqrt{|\mathcal{V}_s^{\text{test}}|}}.
			\end{split}
			\nonumber
		\end{align}
		The first inequality holds by the triangle inequality.
		In contrast, 
		the second inequality holds since $\Psi_u \! = \! \Psi_v$.
		Note that $u$ and $v$ are isomorphic in the topology of the test graph~\citep{zhou2022ood}
		and thus with the same representations.
		
		Next, 
		with probability at least
		$1 \! - \! \sum_{\ell=1}^{L}2(|\bm{h}^{(\ell)}| \! + \! 1)p$,
		for arbitrary entity $v' \in \mathcal{V}_{s}^{\text{test}}$,
		we have
		\begin{align}
			\begin{split}
				\Vert  \Phi_v - \Phi_v' \Vert_{\infty} 
				\leq
				(C_1 + C_2\lVert g \lVert_{\infty})
				\frac{\sqrt{\log(\nicefrac{2|\mathcal{V}_s^{\text{test}}|}{p})}}
				{\sqrt{|\mathcal{V}_s^{\text{test}}|}}.
			\end{split}
			\nonumber
		\end{align}
		
		Then, when the size of the test graph is satisfied 
		and
		$\bm{p}_{uv} = f_{\bm w}(\mathcal{G}^{\text{test}}_s)_{uv} 
		= \texttt{READOUT}(\bm{h}^{(L)}_u, \bm{h}^{(L)}_v)$,
		we have
		\begin{align}
			\begin{split}
				\Vert \bm{p}_{uv} - \bm{p}_{uv’} \Vert_{\infty}
				\leq
				L(M^{\text{train}})\Vert  \Phi_v - \Phi_v' \Vert_{\infty} 
				\leq
				\Vert \bm{p}_{uv} - \tau \Vert_{\infty}.
			\end{split}
			\nonumber
		\end{align}
		
		Specifically,
		if $ \bm{p}_{uv} \ge \tau$,
		we have
		\begin{align}
			\begin{split}
				\bm{p}_{uv’} 
				\geq \bm{p}_{uv} - | \bm{p}_{uv} - \bm{p}_{uv’} |
				\ge    \bm{p}_{uv} - | \bm{p}_{uv} - \tau |
				=       \bm{p}_{uv}  - \bm{p}_{uv} + \tau = \tau.
			\end{split}
			\nonumber
		\end{align}
		
		if $ \bm{p}_{uv} \leq \tau$,
		we have
		\begin{align}
			\begin{split}
				\bm{p}_{uv’} 
				\leq   \bm{p}_{uv} + | \bm{p}_{uv} - \bm{p}_{uv’}|
				\le    \bm{p}_{uv}  +  | \bm{p}_{uv} - \tau |
				=       \bm{p}_{uv}  + \tau -  \bm{p}_{uv} = \tau.
			\end{split}
			\nonumber
		\end{align}

		Thus,
		whether $u,v$ are generated by the same or distinct $g$,
		where the underlying generative function 
		of graph signal $g \! \in \! L^{\infty}$
		is with the essential supreme norm
		as in \citet{maskey2022generalization, zhou2022ood},
		we have a probability at least 
		$1 \! - \! \sum_{\ell=1}^{L}2(|\bm{h}^{(\ell)}| \! + \! 1)p$
		that the same predictions are obtained.
	\end{proof}

	\clearpage
	\section{Implementation Details}
	\label{app:implementation}



	
	\textbf{$\text{MESS}(\cdot)$ and $\text{AGG}(\cdot)$ for GNN-based methods.}
	The two functions of the GNN-based methods
	are summarized in Tab.\ref{tab:GNN_functions}.
	The main differences between different methods
	are
	the combinators for entity and relation representations on the edges,
	the operators on the different representations,
	and the attention weights.
	Recent works \citep{vashishth2019composition,zhu2021neural} have shown 
	that the different combinators and operators only have a slight influence on the performance.
	Compared with GraIL and RED-GNN,
	even though the message functions and aggregation functions are similar,
	their empirical performances are quite different, with different propagation patterns.


	\begin{table}[ht]
		\centering
		\small
		\caption{Summary of the GNN functions for message propagation.
			The values in $\{\}$ represent different operation choices that are tuned as hyper-parameters 
			for different datasets.}
		\label{tab:GNN_functions}
		\resizebox{\textwidth}{!}{
			\begin{tabular}{c|C{170px}|C{150px}}
				\toprule
				method  & $\bm m_{(e_s, r, e_o)}^{(\ell)}:=\text{MESS}(\cdot)$ &  $\bm h_{e_o}^{(\ell)}:=\text{AGG}(\cdot)$ \\
				\midrule
				R-GCN~\citep{schlichtkrull2018modeling}  & $\bm{W}^{(\ell)}_r \bm{h}^{(\ell - 1)}_{e_s}$, where $\bm W_r$ depends on the relation $r$  & $\bm{W}^{(\ell)}_o \bm{h}^{(\ell - 1)}_{e_o} \!+\!\sum_{e_o\in\mathcal N(e_s)} \frac{1}{c_{o,r}} \bm m_{(e_s, r, e_o)}^{(\ell)}$  \\
				\midrule
				CompGCN~\citep{vashishth2019composition}  &  $\bm{W}^{(\ell)}_{\lambda(r)} \{-,*,\star \}(\bm{h}^{(\ell - 1)}_{e_s}, \bm{h}^{(\ell)}_{r})$, where $\bm{W}^{(\ell)}_{\lambda(r)} $ depends on the direction of $r$  &  $\sum_{e_o\in\mathcal N(e_s)} \bm m_{(e_s, r, e_o)}^{(\ell)}$ \\
				\midrule
				GraIL~\citep{teru2019inductive}  &  ${\alpha}^{(\ell)}_{(e_s, r, e_o)|r_q} (\bm{W}^{(\ell)}_1 \bm{h}^{(\ell - 1)}_{e_s}+\bm{W}^{(\ell)}_2\bm h_{e_o}^{(\ell - 1)})$, where  ${\alpha}^{(\ell)}_{(e_s, r, e_o)|r_q}$ is the attention weight.	&  $\bm{W}^{(\ell)}_o \bm{h}^{(\ell - 1)}_{e_o} \!+\!\sum_{e_o\in\mathcal N(e_s)} \frac{1}{c_{o,r}} \bm m_{(e_s, r, e_o)}^{(\ell)}$
				\\
				\midrule
				NBFNet~\citep{zhu2021neural}  &  $\bm W^{(\ell)} \{+,*,\circ \}(\bm h_{e_s}^{(\ell - 1)}, \bm w_q(r, r_q))$, where 
				$\bm w_q(r, r_q)$ is a query-dependent weight vector  &  
				$\!\!\{{\texttt{Sum,Mean,Max,PNA}} \}_{e_o\in\mathcal N(e_s)} \bm m_{(e_s, r, e_o)}^{(\ell)} $ \\
				\midrule
				RED-GNN~\citep{zhang2021knowledge}  &  $\alpha_{(e_s,r,e_o)|r_q}^{(\ell)} (\bm{h}^{(\ell - 1)}_{e_s} + \bm{h}^{(\ell)}_{r})$, where $\alpha_{(e_s,r,e_o)|r_q}^{(\ell)}$ is the attention weight  &   $ \sum_{e_o\in\mathcal N(e_s)} \bm m_{(e_s, r, e_o)}^{(\ell)}$ \\
				\bottomrule
		\end{tabular}}
	\end{table}

	\textbf{Design space of the predictor.}
	From the model's perspective,
	we build the configuration space of the predictor
	and further utilize the advantages
	of existing structural models introduced in Sec.~\ref{sec:couple}.
	Three query-dependent message functions \texttt{MESS($\cdot$)}
	are considered here, including
	DRUM~\citep{sadeghian2019drum} (denoted as $\texttt{M}_\text{DRUM}$), 
	NBFNet~\citep{zhu2021neural} ($\texttt{M}_\text{NBFNet}$),
	and RED-GNN~\citep{zhang2021knowledge} ($\texttt{M}_\text{REDGNN}$).
	The effective message is propagated from $u_q$
	to the entities in subgraph $\mathcal{G}_s$.
	
	Generally,
	the message propagation can be formulated as 
	\begin{equation}	
		\texttt{Predictor} \; f_{\bm{\theta}}: \;
		\bm{h}^{(\ell+1)}_{o} =
		\texttt{DROPOUT} 
		\biggl( \!
		\texttt{ACT} 		
		\Bigl( \!
		\texttt{AGG} 
		\big\{
		\texttt{MESS}
		(\bm{h}^{(\ell)}_{x}, \bm{h}^{(\ell)}_{r}, \bm{h}^{(\ell)}_{o}) \! : \!
		(x,r,o) \! \in \! \mathcal{E}_s 
		\! \big\} 
		\! \Bigr)
		\! \biggr).
	\nonumber
	\end{equation}

	Note the effective message is propagated from $u$
	to the entities in subgraph $\mathcal{G}_s$.
	The ranges for design dimensions
	of the configuration space are shown below,
	where the upper is intra-layer design
	while the lower is inter-layer design.

	\begin{table}[ht]
		\centering
		\small
		\setlength\tabcolsep{5.6pt}
		\begin{tabular}{ccccc}
			\textbf{{DROPOUT}}$(\cdot)$ & \textbf{{ACT}}$(\cdot)$ &  \textbf{{AGG}}$(\cdot)$  &  \textbf{{MESS}}$(\cdot)$ &
			\textbf{Dimension} \\ 
			\midrule
			(0, 0.5) &
			Identity, Relu, Tanh  &
			Max, Mean, Sum &
			$\texttt{M}_\text{DRUM}, \texttt{M}_\text{NBFNet}, \texttt{M}_\text{REDGNN}$ &
			16, 32, 64, 128 \\
		\end{tabular}
	\end{table}
	\begin{table}[ht]
		\centering
		\small
		\vspace{-10px}
		\setlength\tabcolsep{2.8pt}
		\begin{tabular}{ccccc}
			\textbf{No. layers ($L$)} &
			\textbf{Repre. initialization} &
			\textbf{Layer-wise shortcut} &
			\textbf{Repre. concatenation} &
			\textbf{READOUT}$(\cdot)$ 
			\\ \midrule
			\{4, 6, 8, 10\} &
			Binary, Relational &
			True, False  &
			True, False  &
			Linear, Dot product  \\
		\end{tabular}
	\end{table}

	\textbf{Head and tail prediction.}
	Note that predicting a missing head in KG 
	can also be formulated as tail prediction by adding inverse relations. 
	For example, predicting a missing head in KG $(?,q,v)$ 
	can also be formulated to tail prediction by adding inverse relations as $(v,q_\text{inverse},?)$. 
	This formulation involves augmenting the original KG with inverse relations, 
	following the approach adopted in other KG methods
	\citep{zhu2021neural,zhang2021knowledge,zhang2022learning,zhu2022learning, zhang2023emerging, galkin2023towards}.

	\section{Full Evaluations}
	\label{app:evaluation}

	\subsection{Setup}

	\textbf{Datasets.}	
	We use five benchmarks 
	with more than ten-thousand entities,
	including 
	{WN18RR}~\citep{dettmers2017convolutional}, 
	{NELL-995}~\citep{xiong2017deeppath},
	{YAGO3-10}~\citep{suchanek2007yago},
	OGBL-BIOKG
	and
	OGBL-WIKIKG2~\citep{hu2020open}.
	The statistics 
	are summarized in Tab.~\ref{tab:dataset}.

	\begin{table}[ht]
		\centering
		\caption{Statistics of the five KG datasets
			with more than ten-thousand entities.
			Fact triplets in $\mathcal E$ are used to build the graph,
			and
			$\mathcal E^{\text{train}}$,
			$\mathcal E^{\text{val}}$,
			$\mathcal E^{\text{test}}$
			are edge sets of training, validation, and test set.}
		\vspace{-8px}
		\label{tab:dataset}
		\fontsize{8}{8}\selectfont
		\setlength\tabcolsep{14pt}
		\begin{tabular}{c|cc|cccc}
			\toprule
			dataset                 &  $|\mathcal V|$ & $|\mathcal R|$ &  $|\mathcal E|$  & $|\mathcal E^{\text{train}}|$ & $|\mathcal E^{\text{val}}|$ & $|\mathcal E^{\text{test}}|$  \\ \midrule
			{WN18RR}  &  40.9k  &     11     &  65.1k   &  59.0k	&  3.0k  & 3.1k    \\
			NELL-995  &  74.5k  &    200     &  112.2k &	108.9k & 0.5k  & 2.8k   \\
			{YAGO3-10}   &   123.1k  &  37  & 1089.0k & 1079.0k  &  5.0k  & 5.0k  \\
			{OGBL-BIOKG}   &  93.7k   &  51  & 5088.4k	 &  4762.7k &  162.9k &  162.9k  \\
			{OGBL-WIKIKG2}   &  2500.6k   &  535  &  17137.1k  &  16109.2k  &  429.5k  & 598.5k  \\
			\bottomrule
		\end{tabular}
		\vspace{-8px}
	\end{table}

	\textbf{Baselines.}
	We compare our method with general {link prediction} methods,
	including
	(i) semantics models:
	ConvE~\citep{dettmers2017convolutional}, QuatE~\citep{zhang2019quaternion},
	and RotatE~\citep{sun2019rotate};
	and (ii)
	structural models:
	MINERVA~\citep{das2017go},
	DRUM~\citep{sadeghian2019drum},
	RNNLogic~\citep{qu2021rnnlogic},
	CompGCN~\citep{vashishth2019composition},
	DPMPN~\citep{xu2019dynamically}, 
	NBFNet~\citep{zhu2021neural},
	and
	RED-GNN~\citep{zhang2021knowledge}.
	The results of these baseline are taken from their papers
	or reproduced by their official codes.

	\subsection{Comparison With Other Efficient Methods}
	
	To reduce entity vocabulary to be much smaller than full entities, 
	NodePiece~\citep{galkin2022nodepiece} 
	represents an anchor-based approach that facilitates the acquisition of a fixed-size entity vocabulary. Note that NodePiece was originally designed for semantic models, which diverges from the primary focus of our investigation centered around structural models.
	
	As shown below, our {one-shot-subgraph link prediction} method 
	outperforms semantic models, NodePiece, and original RED-GNN in the MRR metric.
	Additionally, it's worth noting that NodePiece 
	can lead to substantial performance degradation 
	when tasked with reducing unique embeddings for a sizable number of entities. 
	The table below illustrates this phenomenon. 
	In contrast, the RED-GNN, when augmented with our one-shot-subgraph method, 
	showcases an ability to enhance performance even while learning and predicting 
	with only 10\% of entities. 
	Furthermore, the parameter count for our structural models remains significantly 
	lower than that of semantic models,
	even with the incorporation of NodePiece improvements.
	
	\begin{table*}[ht]
		\centering
		\caption{Comparison with NodePiece.}
		\vspace{-8px}
		\fontsize{8}{8}\selectfont
		\setlength\tabcolsep{5pt}
		\begin{tabular}{c|ccc|ccc}
			\toprule
			&   \multicolumn{3}{c|}{{WN18RR}}   &  \multicolumn{3}{c}{{YAGO3-10}} \\
			& MRR     & H@10 &  \#Params & MRR    & H@10 & \#Params \\ \midrule
			RotatE (100\% entities) & 0.476 & 57.1 & 41M & 0.495 & 67.0 & 123M \\
			RotatE + NodePiece (10\% entities) & 0.403 & 51.5 & 4.4M & 0.247 & 48.8 & 4.1M  \\
			RED-GNN (100\% entities) & 0.533 & 62.4 & 0.02M & 0.559 & 68.9 & 0.06M  \\
			RED-GNN \textbf{+ {one-shot-subgraph} (10\% entities)} & \textbf{0.567} & \textbf{66.6} & \textbf{0.03M} & \textbf{0.606} & \textbf{72.1} & \textbf{0.09M}  \\
			\bottomrule
		\end{tabular}
	\end{table*}

	Besides, the other efficient {link prediction} method,
	DPMPN~\citep{xu2019dynamically}, 
	contains two GNNs, one is a full-graph GNN that is similar to CompGCN, and the other one is pruned. 
	DPMPN is a mixture of GNN methods, where the pruned GNN already knows the global information and adopts a layer-wise sampling manner. DPMPN requires several propagation steps to prune the message passing and sample the subgraph, while our PPR sampler can efficiently extract the subgraph without learning.
	In comparison, our method is simpler but much more effective
	that it observably outperforms DPMPN. 
	Hence, considering the differences, 
	{one-shot-subgraph link prediction} is still a novel subgraph sampling-based method 
	for {link prediction} and achieves state-of-the-art {prediction} performance.

	As for other sampling-based methods,
	the extracted subgraph by~\citet{yasunaga2021qa}
	is equivalent to the Breadth-first-searching (BFS) that is compared in our work. 
	As it comprises all entities on the k-hop neighbors in its subgraph, the number of sampled entities could be quite large. For example, on the WN18RR dataset, the full 5-hop neighbors of the query entity take up $9.4\%$ of entities and $13.5\%$ of edges, while the full 8-hop neighbors can even take up $69.8\%$ of entities and $89.7\%$ of edges. As for the YAGO3-10 dataset, the full 5-hop neighbors take up $98.1\%$ of entities and $95.6\%$ of edges, almost equal to the entire KG. 
	It could still be expensive for {prediction} and thus not suitable for {prediction} 
	on large KGs like YAGO3-10.

	The sampling method of \citet{mohamed2023locality}
	aims to extract an enclosing subgraph to answer one given query $(u,q,?)$, which is consistent with the previous work GraIL~\citep{teru2019inductive}. 
	Note the extracted subgraphs are different for different given triples. 
	That is, for answering one query $(u,q,?)$, this method requires sampling subgraphs and predicting the score of $N$ potential links on each individual subgraph, where $N$ is the number of all entities. 
	Hence, this method is extremely expensive and also not suitable for large KGs.
	By contrast, our {one-shot-subgraph link prediction} method
	only requires extracting one subgraph 
	to answer one query rather than $N$ subgraphs. 
	Meanwhile, the extracted subgraph is much smaller than the full k-hop neighbor. 
	Thus, our method is more efficient and more suitable 
	for learning and {prediction} on large KGs.

	\subsection{The Edge Split Scheme}
	
	The edge split can be seen as a masking operation on KG, 
	\textit{i.e.}, removing the query edges from the observation edges (the inputs of a {prediction} model). 
	It is necessary for KG learning; otherwise, 
	the query edges for training can be found in the observation graph, which is not practical and not reasonable. 
	We further clarify the edge split scheme in the following three folds.
	
	\textbf{A general perspective of KG incompleteness.}
	Note that the KG datasets are generally and naturally incomplete, and the {link prediction} tasks aim to predict the missing links among entities. An ideal {link prediction} model should be robust to the intrinsic incompleteness of a KG, and the local evidence to be utilized for the prediction of a specific query is also generally incomplete. In this view, the edge split can be seen as a data augmentation method.
	
	\textbf{Training details about edge split.}
	Edge split indeed influences the local connectivity of a KG, where a lower split ratio of fact:train can lead to a sparser observation graph for training. However, we afresh split the query and observation edges in each epoch, and thus, all the edges in the train set can be recursively used as the query edges for training. Besides, all these fact/train edges can be used in the test phase as factual observations. Hence, all the edges in the training set are recursively used in training and explicitly used in testing.
	
	\textbf{The influence of edge split on training.}
	We conduct a further experiment with different split ratios 
	and constrain the experiments using the same amount of training time.
	As the data shown below, the split ratios greatly impact the training time for one epoch, and a higher fact:train ratio leads to short training time. Besides, various split ratios only slightly influence the converged result; however, they greatly influence the speed of convergence. Specifically, a higher split ratio brings a faster convergence speed, especially on the YAGO3-10. 
	
	\begin{table*}[ht]
		\centering
		\caption{Comparison of different split ratios.}
		\vspace{-8px}
		\fontsize{8}{8}\selectfont
		\setlength\tabcolsep{10pt}
		\begin{tabular}{c|cccc|cccc}
			\toprule
			&   \multicolumn{4}{c|}{{WN18RR}}   &  \multicolumn{4}{c}{{YAGO3-10}} \\
			& MRR     & H@1     & H@10 &  Time & MRR     & H@1     & H@10 & Time \\ \midrule
			Fact:Train = 0.70 & 0.566 & 51.1 & 66.7 & 22.9min & 0.552 & 46.9 & 69.9 & 55.6h \\
			Fact:Train = 0.80 & 0.561 & 50.8 & 66.0 & 16.2min & 0.563 & 48.4 & 70.7 & 42.0h \\
			Fact:Train = 0.90 & 0.562 & 51.1 & 66.1 & 8.7min & 0.586 & 51.4 & 71.8 & 23.2h \\
			Fact:Train = 0.95 & 0.567 & 51.4 & 66.6 & 4.5min & 0.587 & 51.7 & 71.1 & 12.1h \\
			Fact:Train = 0.99 & 0.563 & 51.2 & 66.3 & 0.9min & 0.598 & 52.9 & 71.9 & 2.5h \\
			\bottomrule
		\end{tabular}
		\vspace{-6px}
	\end{table*}

	\begin{table*}[ht]
		\centering
		\caption{Comparison of effectiveness with regard to subgraph sampling.}
		\label{tab:ablation-num-entity-edge}
		\vspace{-8px}
		\fontsize{8}{8}\selectfont
		\setlength\tabcolsep{9.2pt}
		\begin{tabular}{cc|ccc|ccc|ccc}
			\toprule
			\multirow{2}{*}{$r^q_{\mathcal{V}}$}   & \multirow{2}{*}{$r^q_{\mathcal{E}}$} &  
			\multicolumn{3}{c|}{{WN18RR}}   &  \multicolumn{3}{c|}{{NELL-995}}    &  \multicolumn{3}{c}{{YAGO3-10}}\\
			&   & MRR     & H@1     & H@10 & MRR     & H@1     & H@10 & MRR     & H@1     & H@10  \\ \midrule
			1.0 & 1.0 & 0.549 & 50.2 & 63.5 & 0.507  & 43.8 & 61.9  & 0.598 & 53.3 & 71.2 \\
			0.5 & 0.5 & 0.555 & 50.6 & 64.4 & 0.537 &47.3  & 64.2  & 0.599 & 53.4 & 71.4  \\
			0.2 & 0.2 & 0.563 & 51.0 & 66.0 & 0.541 & 47.9 & 63.9 & 0.603 & 53.8 & 71.8 \\
			0.1 & 0.1 & 0.567 & 51.4 & 66.4 & 0.539 & 48.0 & 63.0 & 0.599 & 53.6 & 70.8 \\
			\bottomrule
		\end{tabular}
		\vspace{-6px}
	\end{table*}

	\begin{figure}[ht]
		\centering
		\includegraphics[width=10cm]{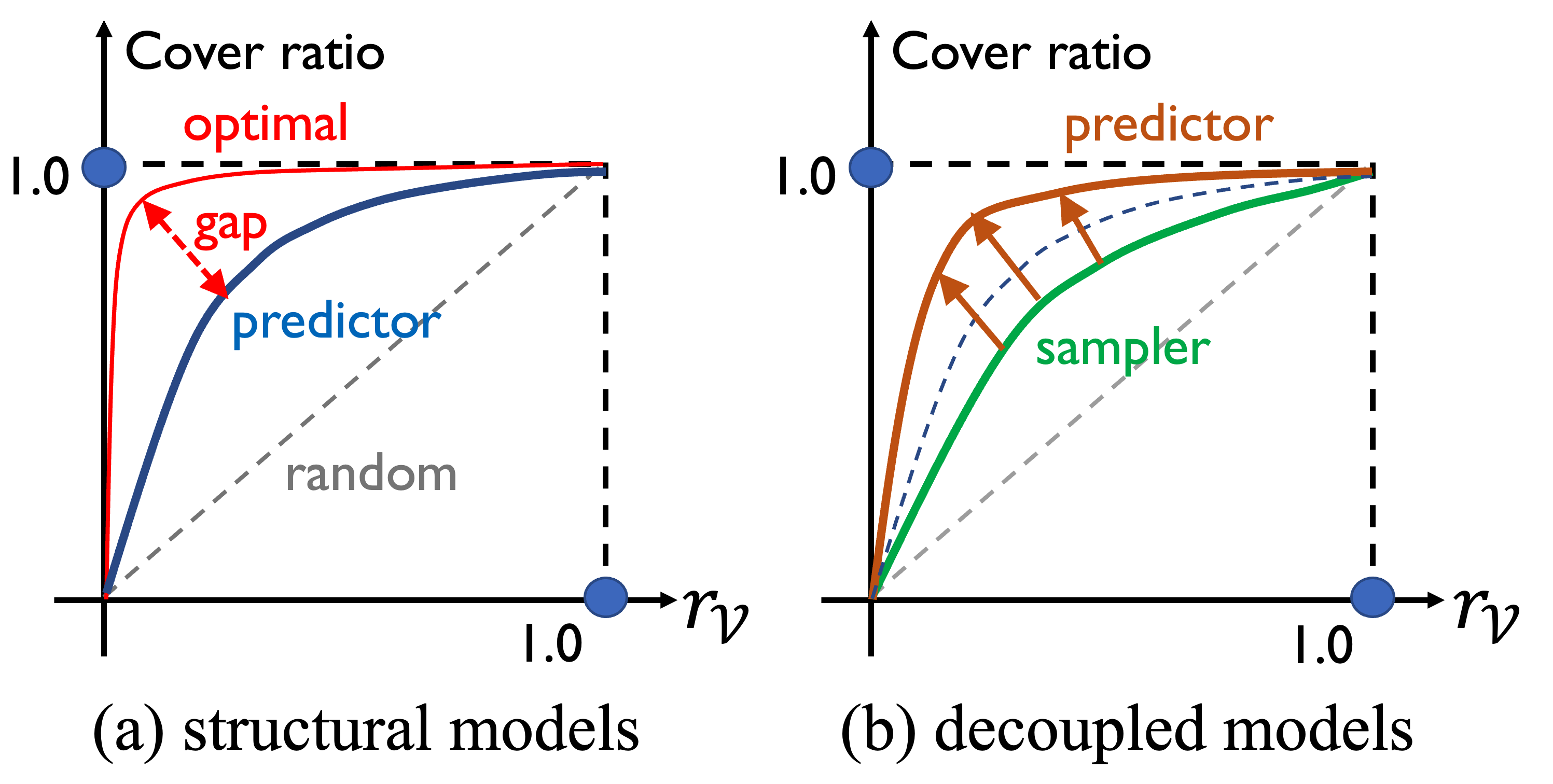}
		\caption{
			Illustrations of structural models (a)
			and decoupled subgraph models (b).
		}
		\label{fig: Comparison-structural-decoupled}
	\end{figure}

	\subsection{Robustness Against Data Perturbations}
	
	\textbf{Randomly adding noise.} We generate a noisy dataset by randomly adding noisy edges that are not observed from the dataset. Specifically, the entity set and relation set are kept the same. The noise ratio is computed as $\epsilon=\nicefrac{|\mathcal{E}^{\text{noise}}|}{|\mathcal{E}^{\text{train}}|}$. The $\mathcal{E}^{\text{val}}$ and $\mathcal{E}^{\text{test}}$ are clean to guarantee the accurate evaluation.
	
	As the results shown in Tab.~\ref{tab:added-noise}, such noise does not significantly influence the prediction performance. Even in the high-noise scenario ($\epsilon=50\%$), the MRR result is nearly identical to that on the clean dataset. This result shows the preliminary robustness of adding noise, and a further investigation should be conducted with more complex noise patterns, e.g., by adopting adversarial techniques.
	
	\begin{table*}[ht]
		\centering
		\caption{{Model performance (MRR) with randomly added noise.}}
		\label{tab:added-noise}
		\vspace{-8px}
		\fontsize{8}{8}\selectfont
		\setlength\tabcolsep{10pt}
		\begin{tabular}{c|cccccc}
			\toprule
			Dataset  &  $\epsilon=0\%$ (clean) & $\epsilon=10\%$ &  $\epsilon=20\%$ &  $\epsilon=30\%$ &  $\epsilon=40\%$ &  $\epsilon=50\%$ \\ \midrule
			WN18RR                           & 0.567      & 0.567 & 0.566 & 0.566 & 0.564 & 0.564 \\
			NELL-995                         & 0.547      & 0.547 & 0.547 & 0.547 & 0.546 & 0.547 \\ 
			YAGO3-10                         & 0.606      & 0.605 & 0.603 & 0.601 & 0.602 & 0.601 \\
			\bottomrule
		\end{tabular}
		\vspace{-6px}
	\end{table*}
	
	\textbf{Randomly deleting facts.} We create a more incomplete KG by randomly deleting the facts in the training set. The delete ratio is computed as $r=\nicefrac{|\mathcal{E}^{\text{delete}}|}{|\mathcal{E}^{\text{train}}|}$. 
	The $\mathcal{E}^{\text{val}}$ and $\mathcal{E}^{\text{test}}$ are kept the same as the original dataset.
	As shown in Tab.~\ref{tab:delete-noise}, this kind of input perturbation greatly degenerates the performance, as the link prediction could rely heavily on the correct edges in the original dataset. Here, a sparser graph with more deleted edges can not sufficiently support the prediction, and the message propagation from the query entity is also hindered. Hence, combating graph incompleteness, 
	data heterogeneity~\citep{tang2022virtual, GossipFL},
	noisy annotations~\citep{zhou2023combating}, or adversarial attacks~\citep{zhang2022adversarial, chen2022understanding, zhou2023mcgra, zhangs2023EPSAD, li2023learning}, would be a valuable direction.
		
	\begin{table*}[ht]
		\centering
		\caption{{Model performance (MRR) with randomly added noise.}}
		\label{tab:delete-noise}
		\vspace{-8px}
		\fontsize{8}{8}\selectfont
		\setlength\tabcolsep{10pt}
		\begin{tabular}{c|cccccc}
			\toprule
			Dataset  &  $r=0\%$ (full) & $r=10\%$ &  $r=20\%$ &  $r=30\%$ &  $r=40\%$ &  $r=50\%$ \\ \midrule
			WN18RR                     & 0.567     & 0.507 & 0.451 & 0.394 & 0.338 & 0.278 \\
			NELL-995                   & 0.547     & 0.515 & 0.483 & 0.446 & 0.412 & 0.365 \\
			YAGO3-10                   & 0.606     & 0.553 & 0.489 & 0.434 & 0.385 & 0.329 \\
			\bottomrule
		\end{tabular}
		\vspace{-6px}
	\end{table*}
	
	\section{Further Discussion}
	
	
	\subsection{Design Principles}
	Note that
	it is non-trivial to achieve the one-shot-subgraph link prediction,
	where three-fold questions are required to be answered: 
	\textbf{(i)} From the \textit{data}'s perspective,
	what kind of sampler is suitable here?
	\textbf{(ii)} from the \textit{model}'s perspective,
	how to build up the predictor's architecture 
	to be expressive on subgraphs?
	\textbf{(iii)} from the \textit{optimization}'s perspective,
	how to optimize the sampler and predictor jointly and efficiently?
	
	The major technical challenge lies in the 
	design and implementation of 
	an efficient and query-dependent sampler.
	Given a query $(u, q, ?)$,
	the sampler should not only 
	preserve the local neighbors of $u$
	that contain the potential answers and supporting facts,
	but also be able to distinguish different $q$
	as the essential information could be different.
	Accordingly, the two design principles are as follows.
	
	{
	\textbf{Principle-1: Local-structure-preserving.}
	Since the target entity $v$ is unknown,
	freely sampling from all the entities in original $\mathcal{G}$
	will inevitably discard the structural connection 
	between $u$ and promising $v$.
	This manner of existing node-wise or layer-wise sampling methods
	can deteriorate the message flow started from $u$,
	and thus degenerate the whole {prediction} process.
	Here, the local structure of $u$ is expected to be preserved in $\mathcal{G}_s$,
	and each sampled entity should be reachable from $u$.}
	
	{
	\textbf{Principle-2: Query-relation-aware.}
	The objective of learning relation-aware sampling here 
	is to sample the promising targets $v$ 
	according to the query relation $q$.
	Note that, in common homogeneous graphs, the relation may be unnecessary, 
	as all relations between nodes share the same semantics. 
	However, the relation here does matter, as it
	presents the distinct characteristics of a KG.
	Thus, in our design, the sampler is expected to be query-relation-aware
	in order to adapt to KGs efficiently.}
	
	\subsection{Contributions}
	
	Here, we further explain the novelty and contributions in the following three folds.
	
	\textbf{A valuable research problem.}
	Our investigation delves into the limitations of structural models, which confront a pronounced scalability challenge. These models rely on {prediction} over the entire Knowledge Graph, encompassing all entities and edges, with the additional burden of scoring all entities as potential answers. This approach proves to be highly inefficient, impeding their optimization when applied to large-scale KGs. As a consequence, we present an open question, pondering how to conduct {prediction} on knowledge graphs efficiently and effectively, seeking avenues to overcome this hindrance.
	
	\textbf{A conceptual framework with several practical instantiations.}
	In response to the prevalent scalability challenges faced by existing KG methods, we present a conceptual solution, \textit{i.e.}, the {one-shot-subgraph link prediction} {manner}. This novel approach promises enhanced flexibility and efficacy in design. Specifically, instead of conducting direct {prediction} on the complete original KG, we advocate a two-step {prediction} process involving subgraph sampling and {subgraph-based prediction}. In this regard, our proposed {prediction} {manner} comprises a sampler and a predictor. Leveraging the efficiency gains derived from subgraph-based {prediction}, we introduce an optimization technique for subgraph-based searching, incorporating several well-crafted technical designs. These innovations aim to strike a balance between {prediction} efficiency and the complexity associated with identifying optimal configurations within both data and model spaces.
	
	\textbf{Several important discoveries from experiments.}
	Through comprehensive experimentation on three prevalent KGs, we demonstrate that our framework achieves state-of-the-art performance. Particularly noteworthy are the substantial advancements we achieve in both efficiency and effectiveness, a trend that is particularly evident in the case of the large-scale dataset YAGO3-10. Our quest for optimal configurations leads us to the intriguing revelation that utilizing the entire KG for {prediction} is unnecessary. Instead, simple heuristics can efficiently identify a small proportion of entities and facts essential for answering specific queries without the need for additional learning. These compelling findings hold significant meaning and are poised to pique the interest of the KG community.

	\subsection{Explanation for the Improvement in Performance}
	\label{app:explanation}
	
	Here, we provide a two-fold explanation for better performance
	as follows.
	
	\textbf{Data perspective:} 
	Extracting subgraphs can remove irrelevant information for {link prediction}. 
	Conventional structural models explicitly take all the entities and edges into {prediction}, ignoring the correlation between the entities and the query relation. 
	As delineated in Sec.~\ref{sec:exp} of this paper, our {one-shot-subgraph link prediction} approach effectively discerns and excludes irrelevant entities while retaining the proper answers. This strategic refinement, in turn, contributes to simplifying the learning problem, thereby amplifying {prediction} performance. Our findings demonstrate that a mere fraction (\textit{i.e.}, 10\%) of entities suffices for answering specific queries, \textit{i.e.}, a subset efficiently identified by the heuristic Personalized PageRank mechanism without the need for learnable sampling.
	
	Another supporting material is that only relying on a subgraph for {prediction} can also boost the test-time performance~\citep{miao2022interpretable}. 
	The proposed GAST method~\citep{miao2022interpretable}, aims to extract a subgraph $G_s$ as the interpretation of a GNN. It inherits the same spirit of information bottleneck in building its optimization objective, \textit{i.e.}, $\min -I(G_s;Y) + \beta \cdot I(G_s;G)$. The integrated subgraph sampler can explicitly remove the spurious correlation or noisy information in the entire graph $G$, which is similar to our {one-shot-subgraph link prediction} framework.
	
	\textbf{Model perspective: }
	The higher learning efficiency on subgraphs can further boost hyper-parameter optimization~\citep{zhang2022kgtuner}.
	Generally, the task of hyperparameter optimization on extensive graph datasets poses substantial challenges due to the inherent inefficiencies in model training. However, the implementation of our subgraph-based method results in a substantial improvement in training efficiency. 
	
	Consequently, we can rapidly obtain evaluation feedback for configurations sampled from the hyperparameter space. Sec.~\ref{ssec: optimization} introduces an optimization technique for subgraph-based searching that features meticulously crafted technical designs. 
	The searched configuration usually leads to a deeper GNN (\textit{i.e.}, 8 layers, while previous studies are usually limited to 5 or 6 layers) that increases the expressiveness of {prediction}. These innovations aim to balance {prediction} efficiency and the complexity associated with identifying optimal configurations within data and model spaces, which also contributes to improved {prediction} performance.

	\subsection{Extension}
	
	Note that KG learning usually focuses more on link-level tasks, \textit{e.g.}, link prediction. 
	However, we firmly believe that {one-shot-subgraph link prediction} has the potential 
	to be extended and adapted for various graph learning tasks. 
	One general direction is to 
	adapt the {one-shot-subgraph link prediction} framework
	to other kinds of graph learning tasks,
	\textit{e.g.}, the node-level or graph-level tasks.
	
	For instance, with the PPR sampler, one can sample 
	a single-source "local" subgraph for node classification 
	or a multi-source "global" subgraph for graph classification, 
	where the rationale of first sampling and then {prediction} remains applicable. 
	Besides, 
	enhancing the {one-shot-subgraph link prediction} 
	with instance-wise adaptation
	is also a promising direction.
	That is, 
	sampling a subgraph of suitable scale for each given query,
	which can potentially improve the upper limit of {prediction}.
	
	{
	Furthermore, conducting link prediction with new relations or new entities is also a frontier topic. Improving the generalization or extrapolation power of GNN can be vital in practice. Considering the significant few-shot in-context learning of the large language model (LLM), an appropriate synergy between the latest LLM and current GNN will be a promising direction. Improving the efficiency and scalability of predicting with large graphs is also of great importance here.}
	
	{
	In addition, from a broader perspective of trustworthy machine learning, one should also consider the intrinsic interpretability and the robustness problem. These trustworthy properties can help users understand the model better and also keep it in a safe and controllable way.}

	\newpage	
	\section{Case Study}
	\label{app:case}

	In this section, we show the sampled subgraph of 
	different scales
	on three datasets as follows.

	\begin{figure*}[ht]
		\centering
		\hfill
		\includegraphics[width=6.8cm]{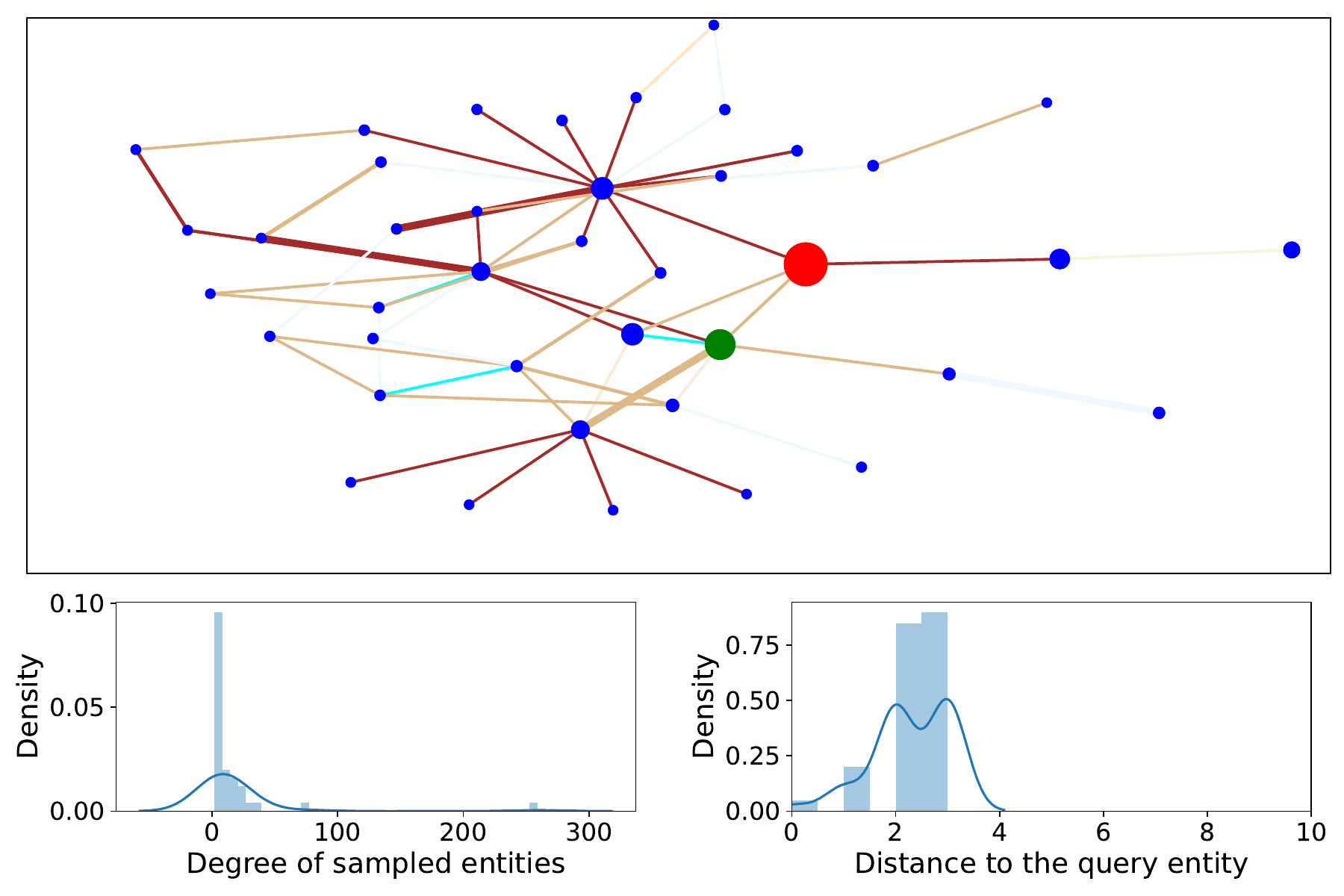}
		\hfill
		\includegraphics[width=6.8cm]{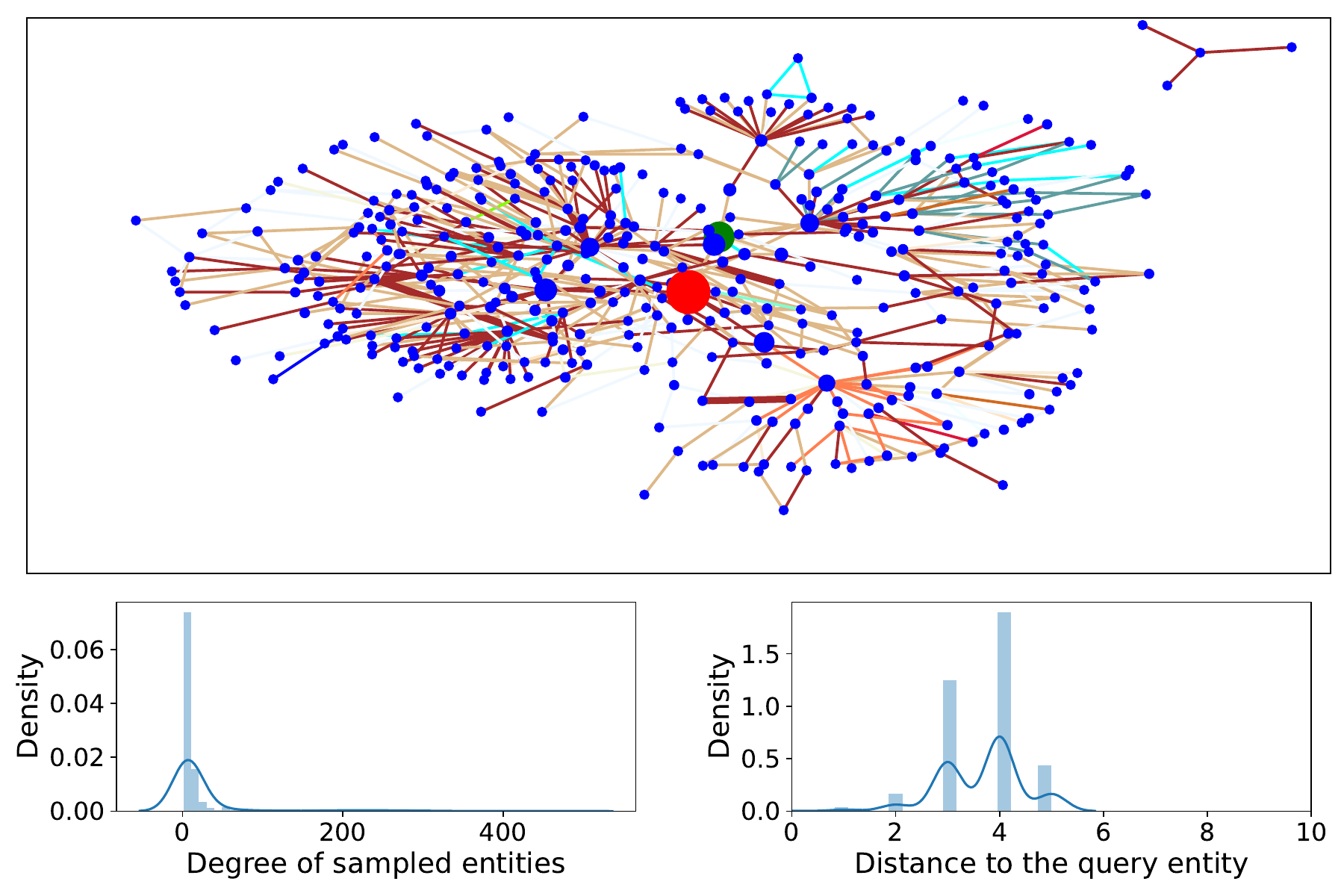}
		\hfill
		\vspace{-8px}
		\caption{
			Subgraphs (0.1\% and 1\%) from WN18RR:
			$u \! = \! 1, q \! = \! 12, v \! = \! \{ 5305 \}$.
		}
		\vspace{-4px}
	\end{figure*}
	
	\begin{figure*}[ht]
		\centering
		\hfill
		\includegraphics[width=6.8cm]{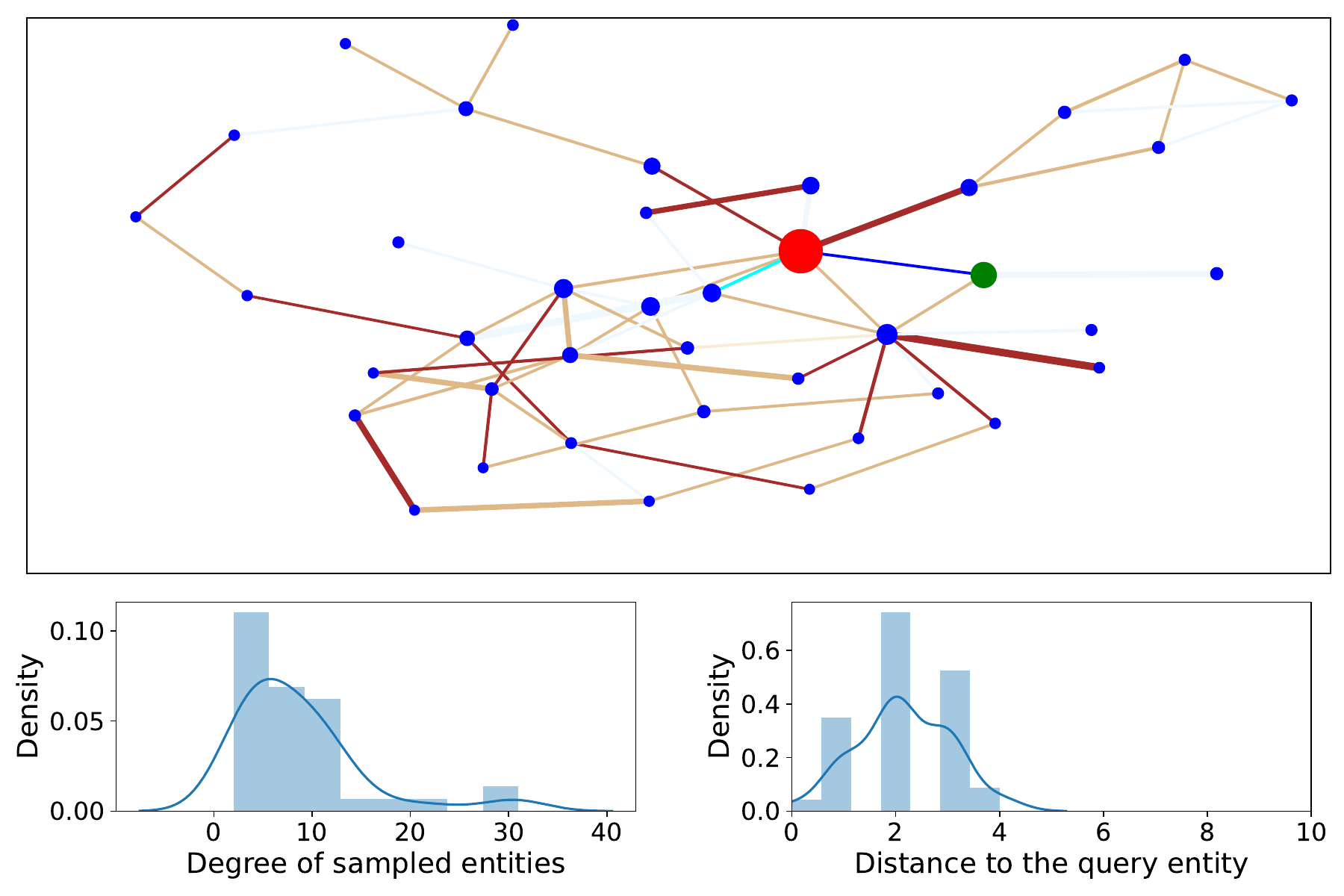}
		\hfill
		\includegraphics[width=6.8cm]{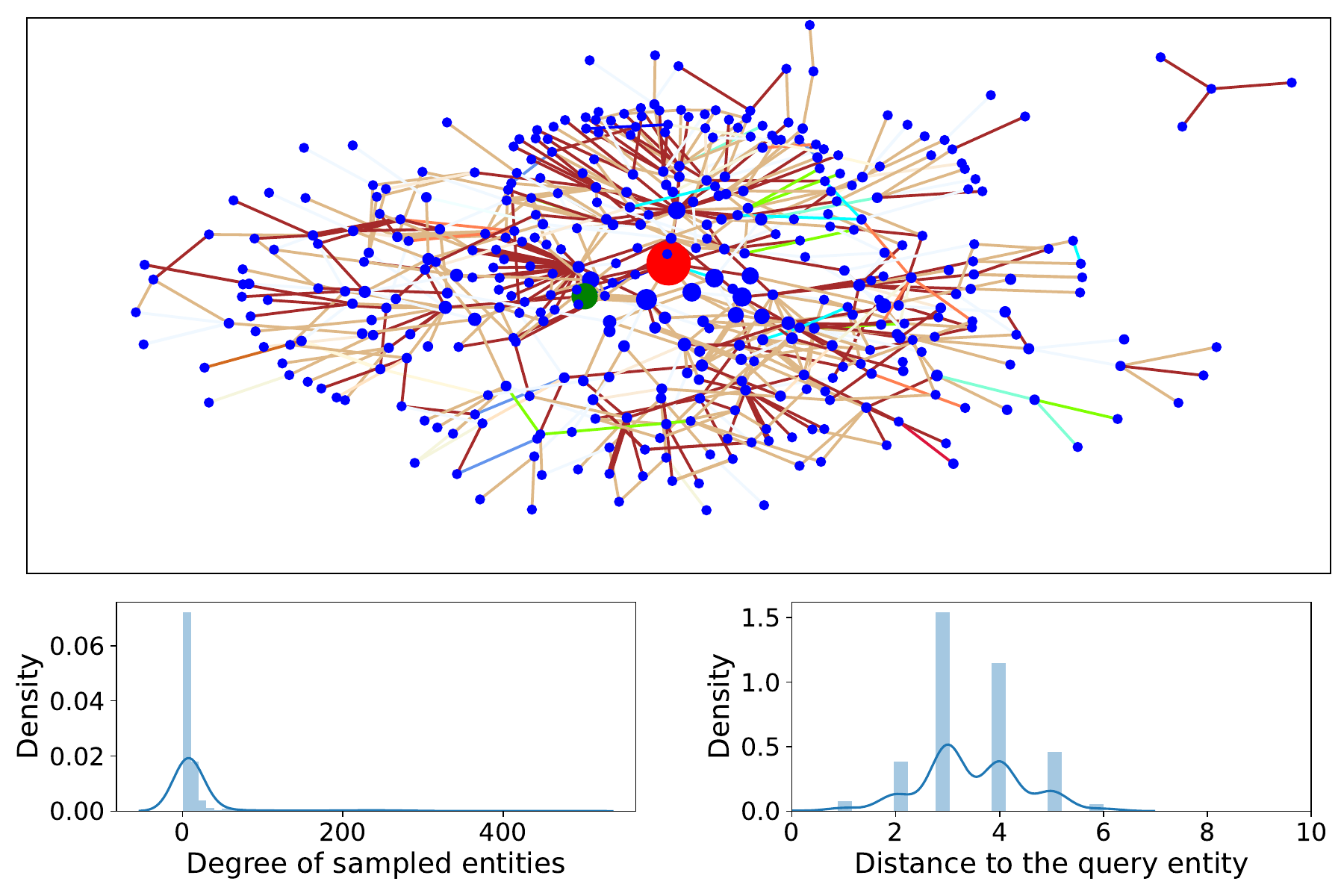}
		\hfill
		\vspace{-8px}
		\caption{
			Subgraphs (0.1\% and 1\%) from WN18RR:
			$u \! = \! 9, q \! = \! 20, v \! = \! \{ 38116 \}$.
		}
		\vspace{-4px}
	\end{figure*}

	\begin{figure*}[ht]
		\centering
		\hfill
		\includegraphics[width=6.8cm]{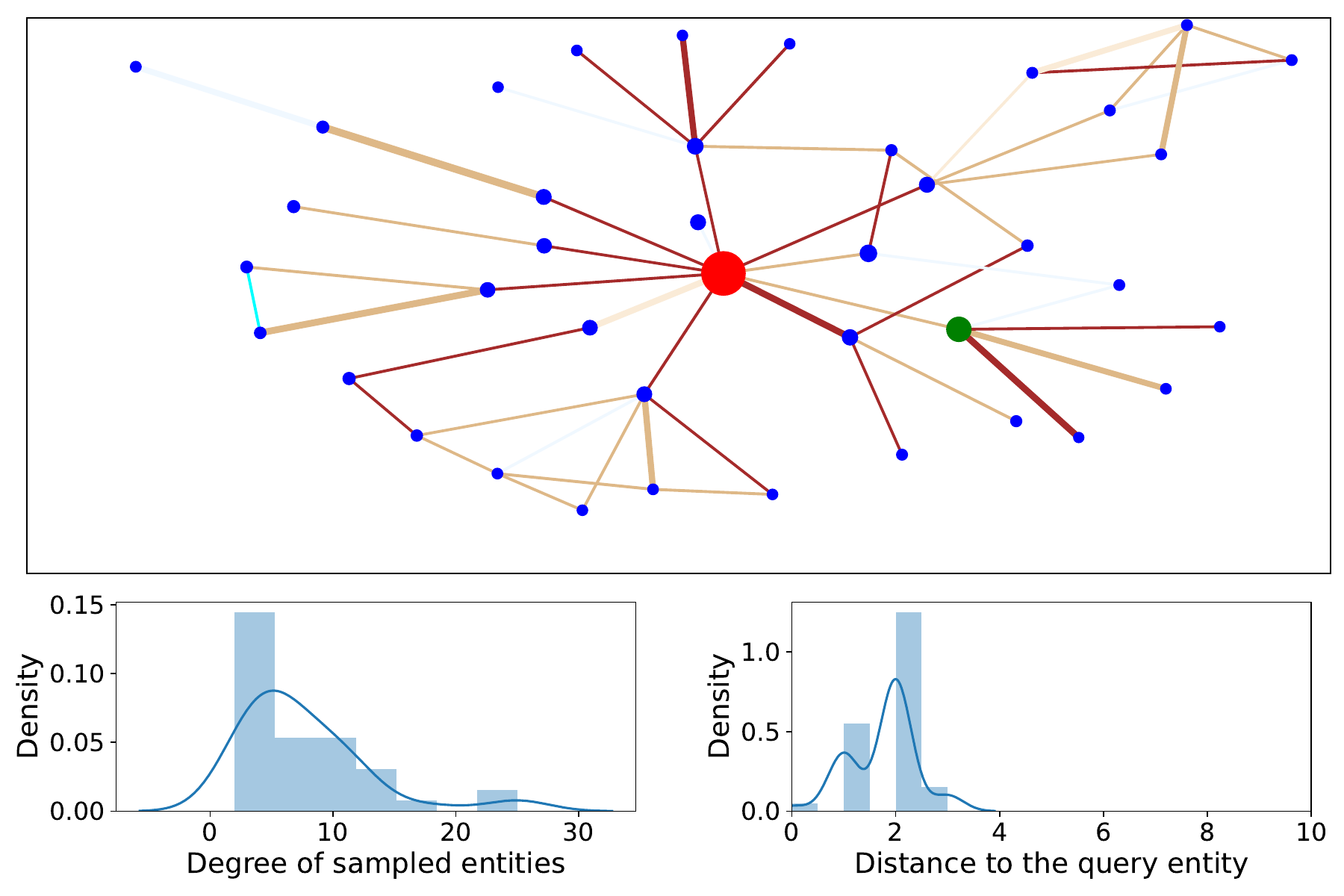}
		\hfill
		\includegraphics[width=6.8cm]{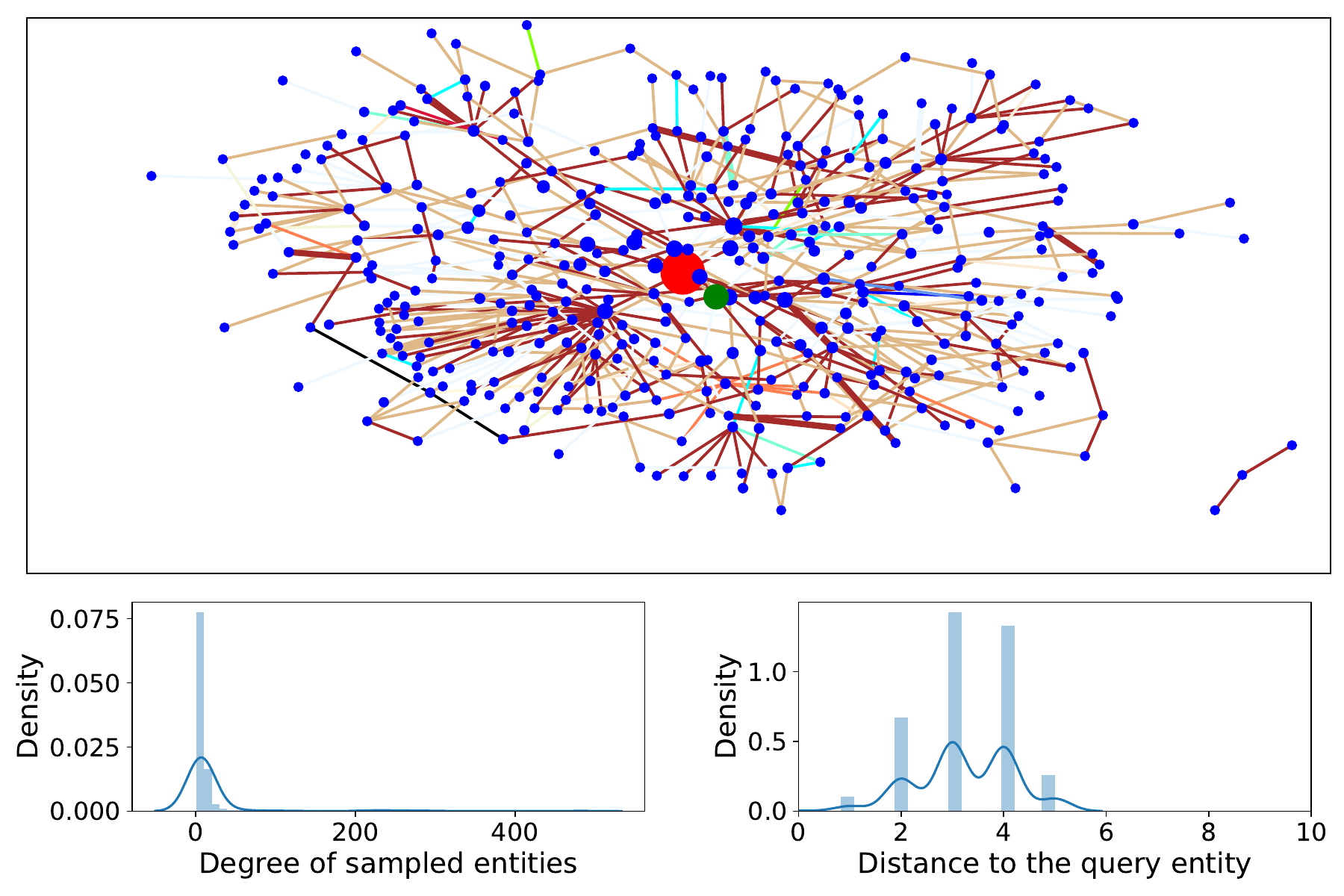}
		\hfill
		\vspace{-8px}
		\caption{
			Subgraphs (0.1\% and 1\%) from WN18RR:
			$u \! = \! 29, q \! = \! 1, v \! = \! \{ 11186 \}$.
		}
		\vspace{-4px}
	\end{figure*}
	
	\begin{figure*}[ht]
		\centering
		\hfill
		\includegraphics[width=6.8cm]{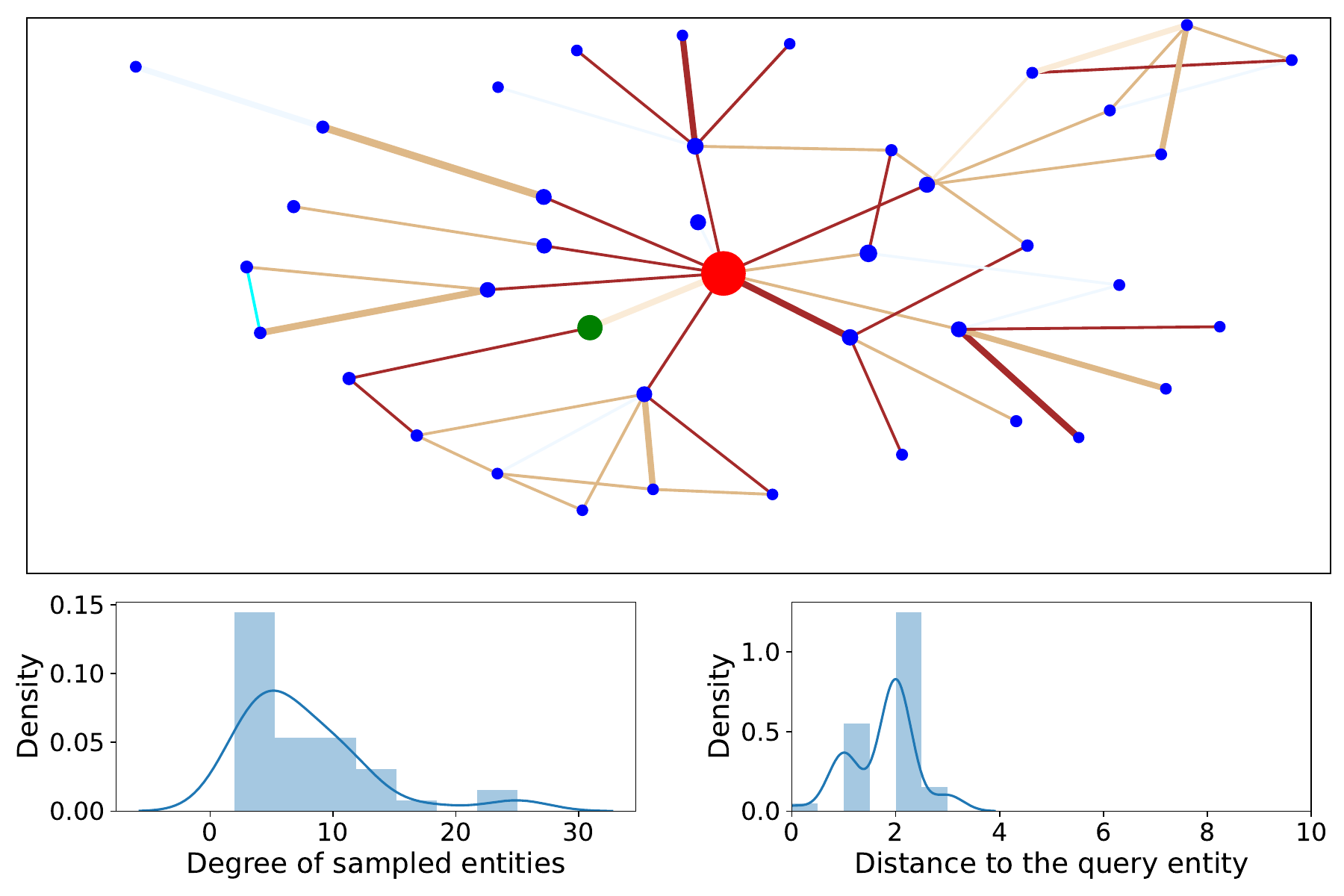}
		\hfill
		\includegraphics[width=6.8cm]{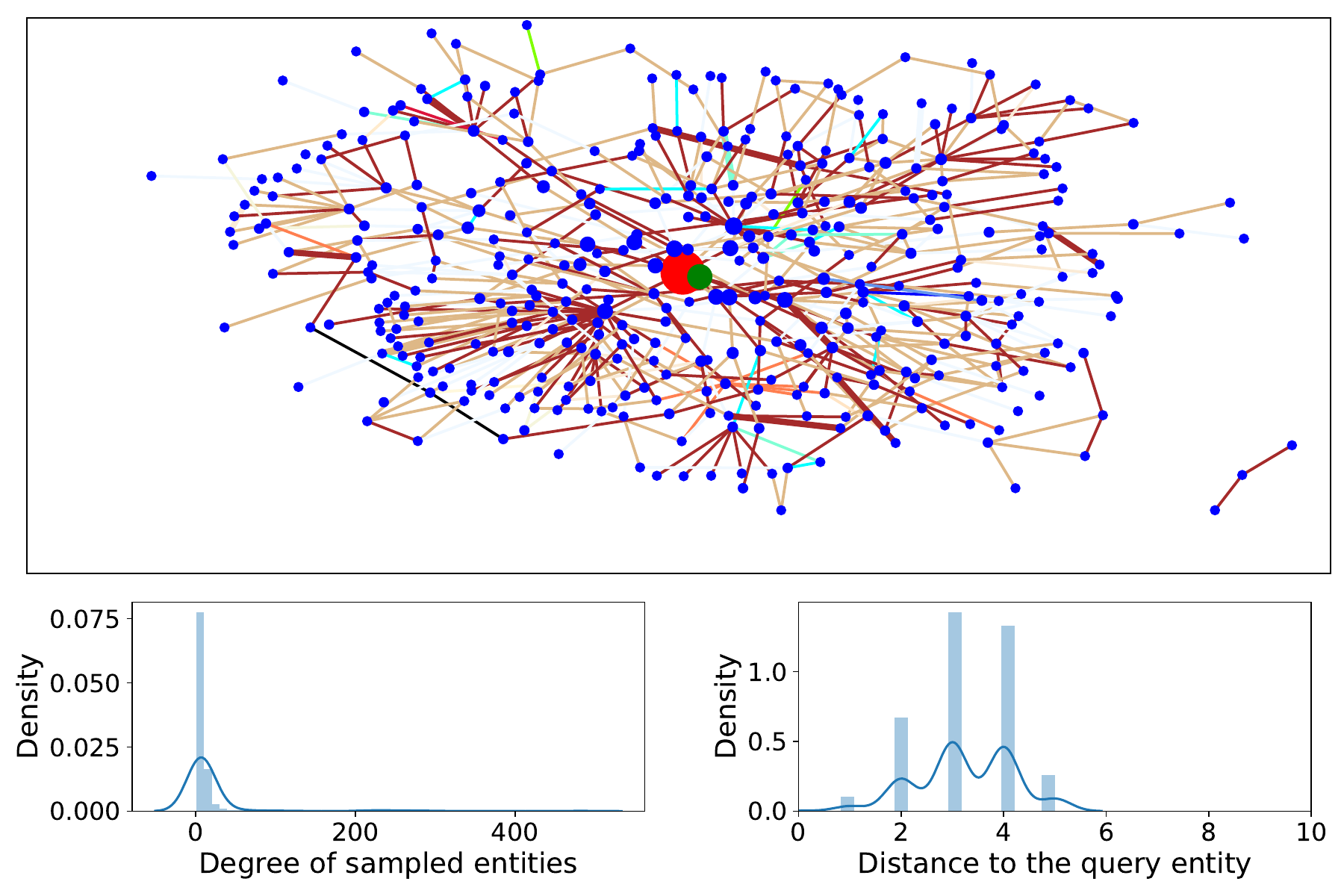}
		\hfill
		\vspace{-8px}
		\caption{
			Subgraphs (0.1\% and 1\%) from WN18RR:
			$u \! = \! 29, q \! = \! 12, v \! = \! \{ 6226 \}$.
		}
		\vspace{-4px}
	\end{figure*}

	\begin{figure*}[ht]
		\centering
		\hfill
		\includegraphics[width=6.8cm]{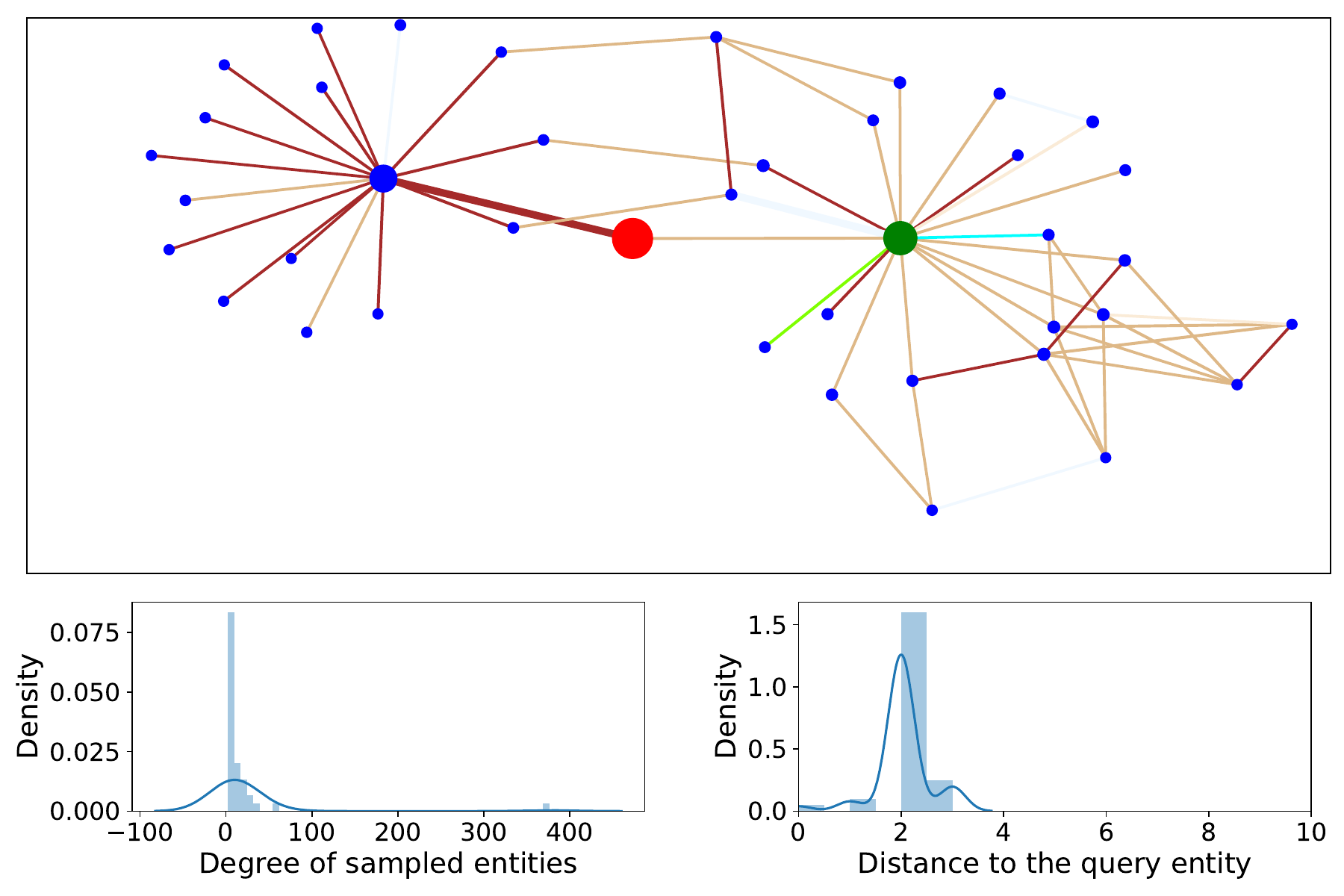}
		\hfill
		\includegraphics[width=6.8cm]{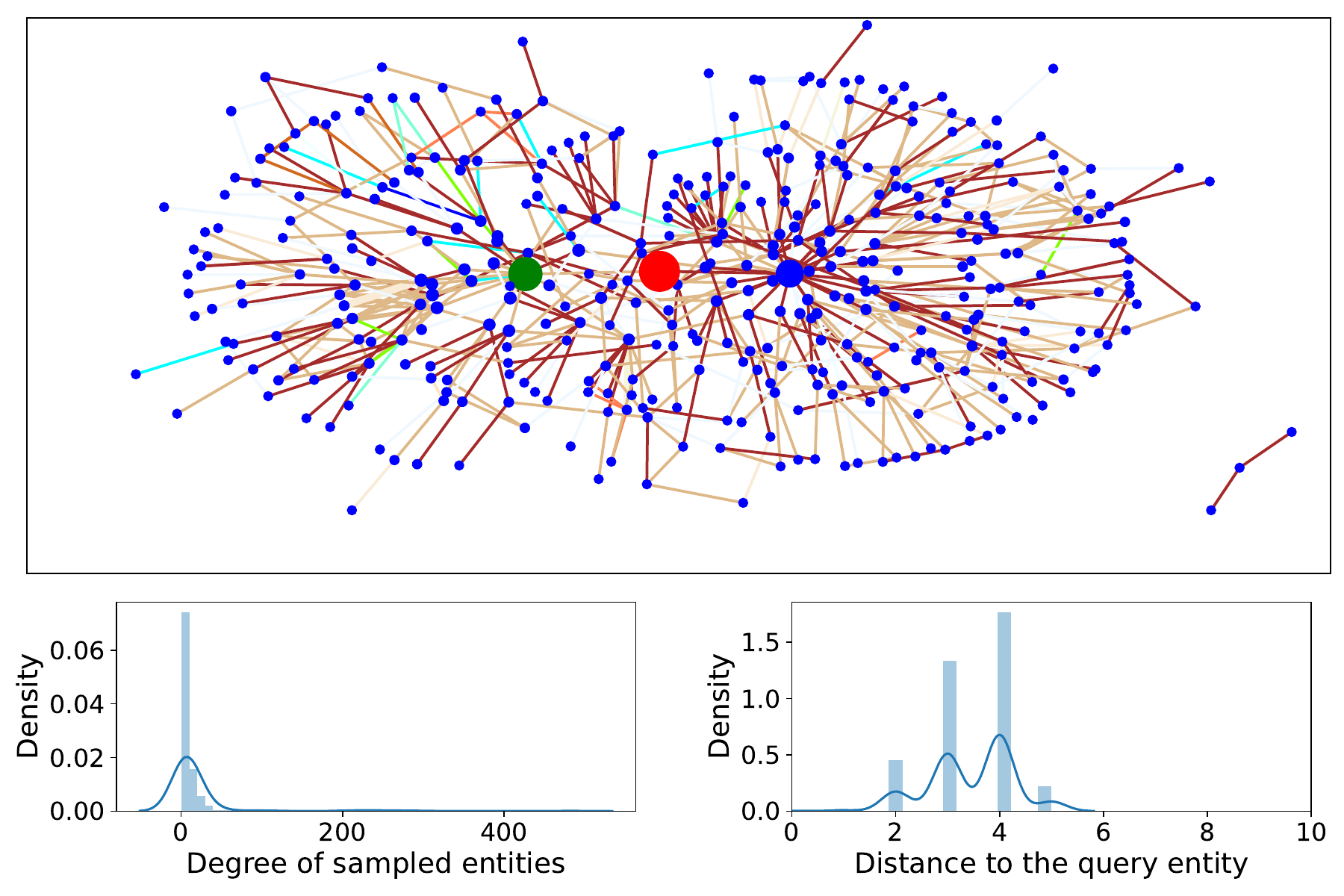}
		\hfill
		\vspace{-8px}
		\caption{
			Subgraphs (0.1\% and 1\%) from WN18RR:
			$u \! = \! 44, q \! = \! 12, v \! = \! \{ 45 \}$. 
		}
		\vspace{-4px}
	\end{figure*}
	
	\begin{figure*}[ht]
		\centering
		\hfill
		\includegraphics[width=6.8cm]{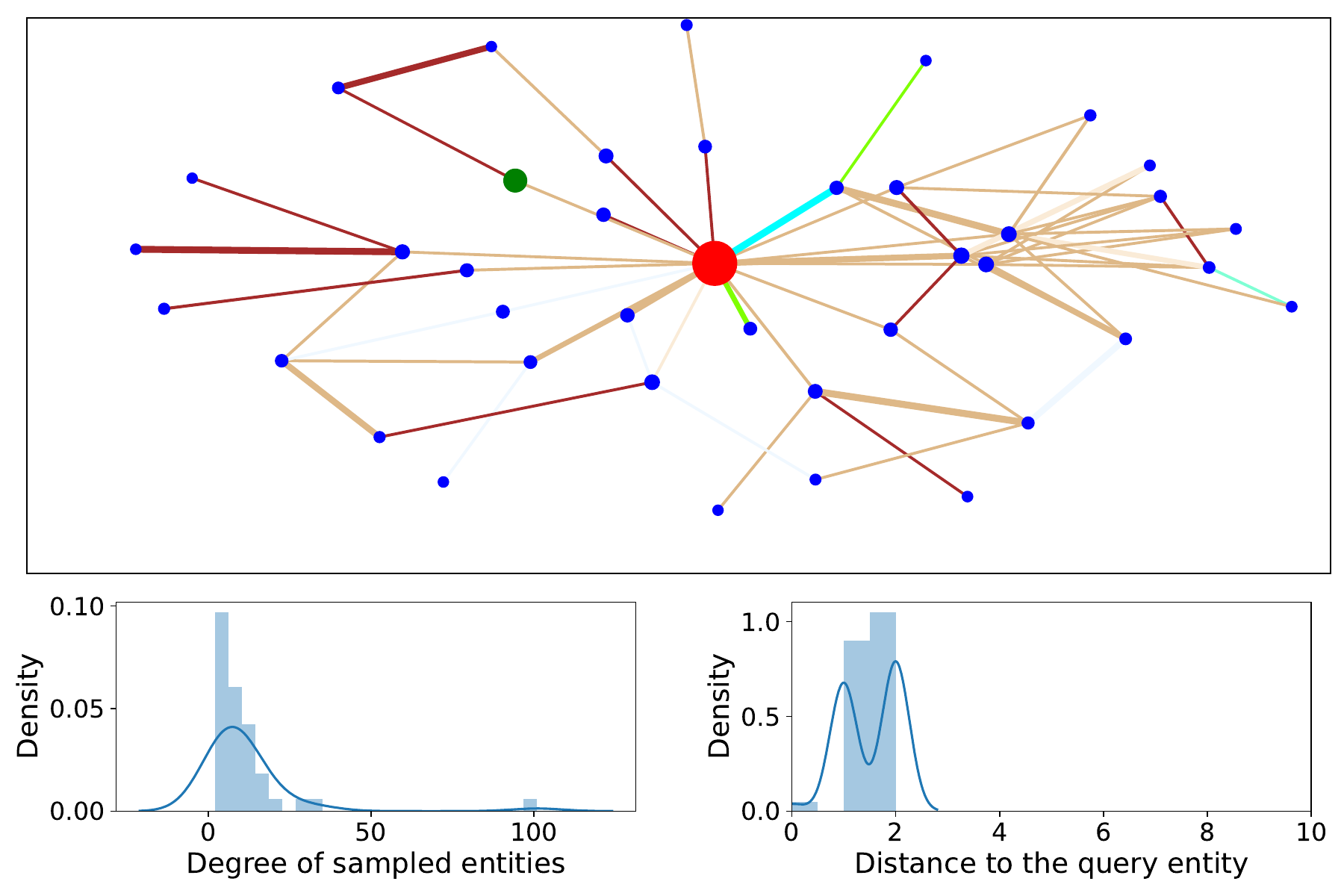}
		\hfill
		\includegraphics[width=6.8cm]{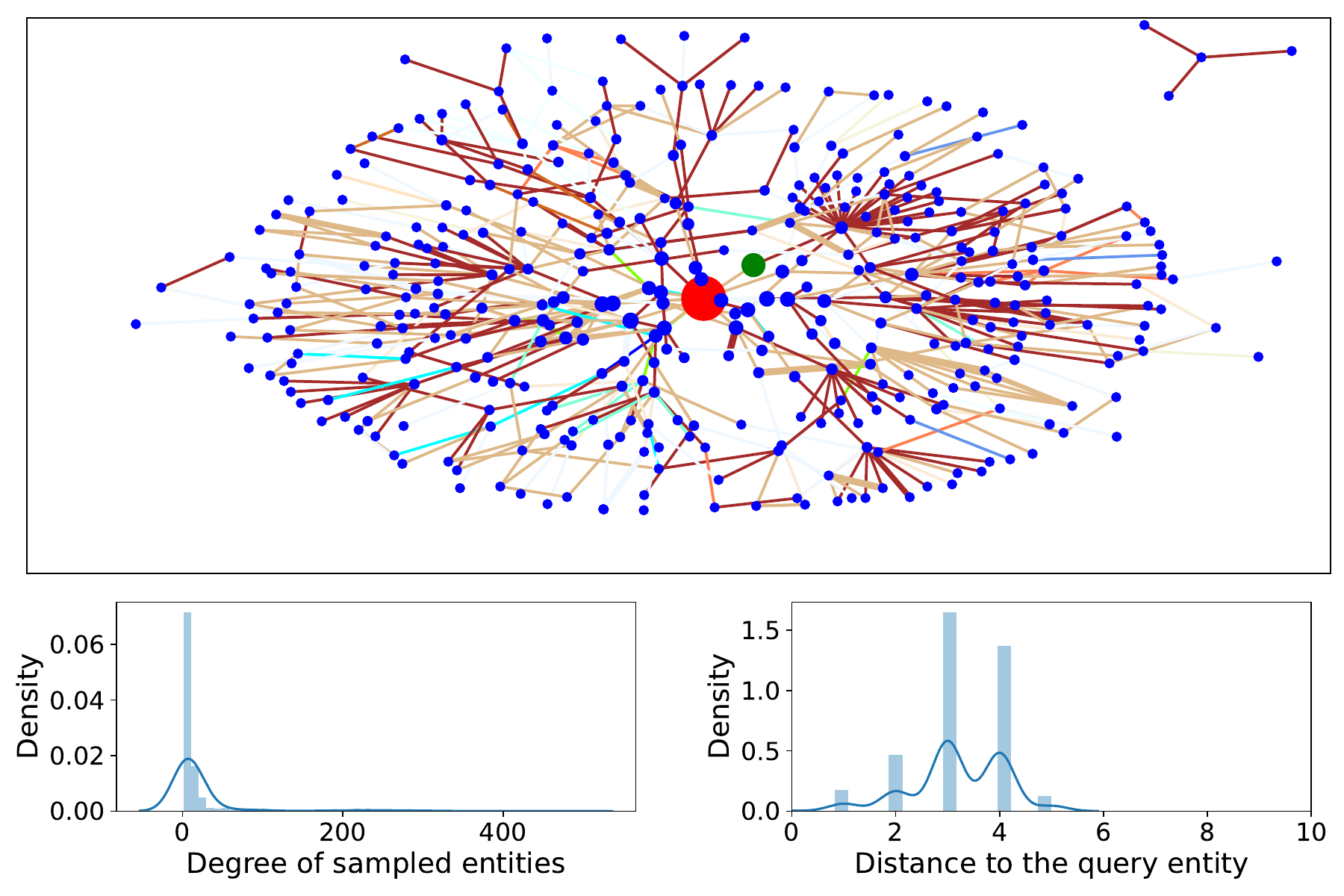}
		\hfill
		\vspace{-8px}
		\caption{
			Subgraphs (0.1\% and 1\%) from WN18RR:
			$u \! = \! 45, q \! = \! 1, v \! = \! \{ 44 \}$. 
		}
		\vspace{-4px}
	\end{figure*}

	\begin{figure*}[ht]
		\centering
		\hfill
		\includegraphics[width=6.8cm]{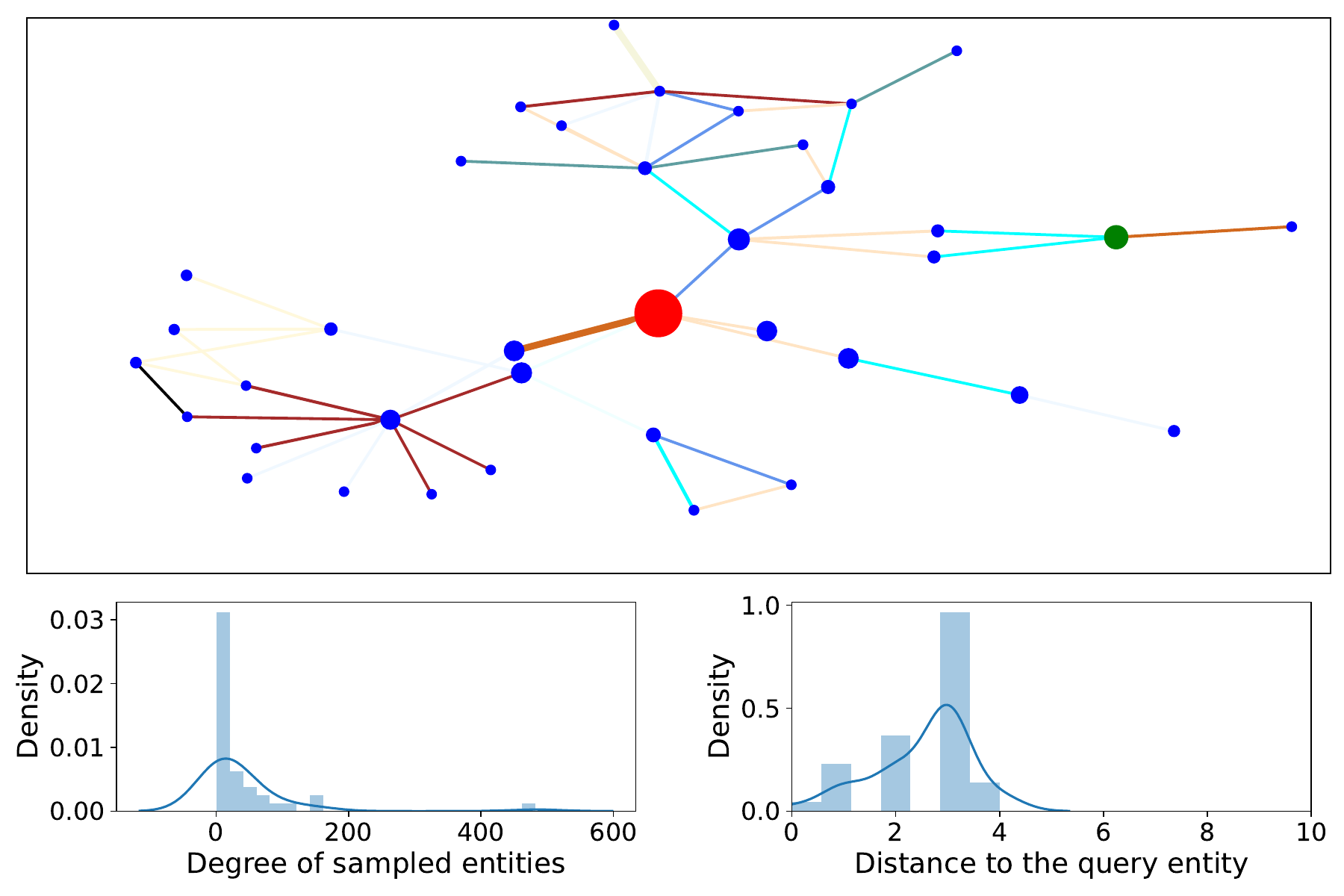}
		\hfill
		\includegraphics[width=6.8cm]{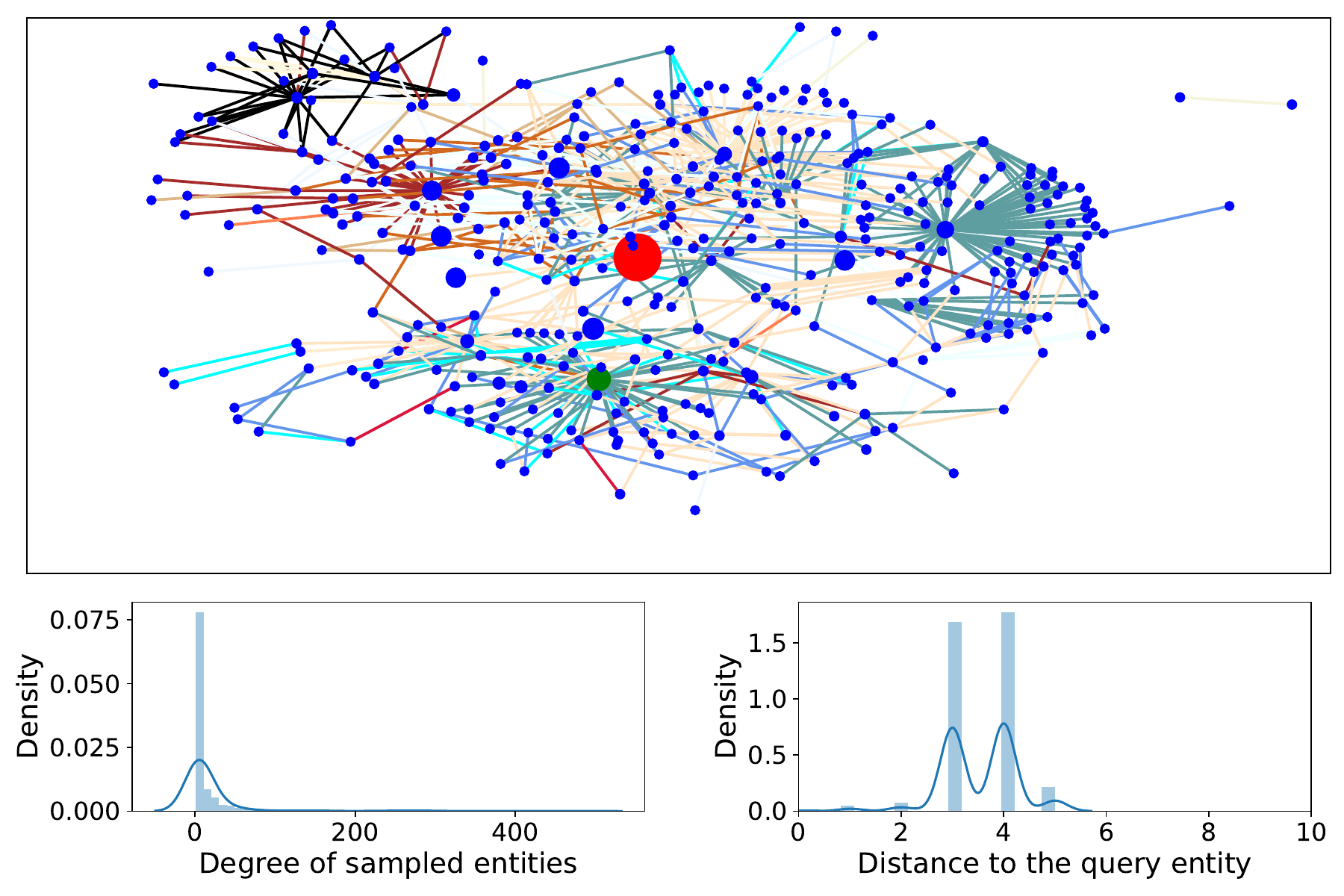}
		\hfill
		\vspace{-8px}
		\caption{
			Subgraphs (0.1\% and 1\%) from WN18RR:
			$u \! = \! 60, q \! = \! 2, v \! = \! \{ 6577 \}$.
		}
		\vspace{-4px}
	\end{figure*}
	
	\begin{figure*}[ht]
		\centering
		\hfill
		\includegraphics[width=6.8cm]{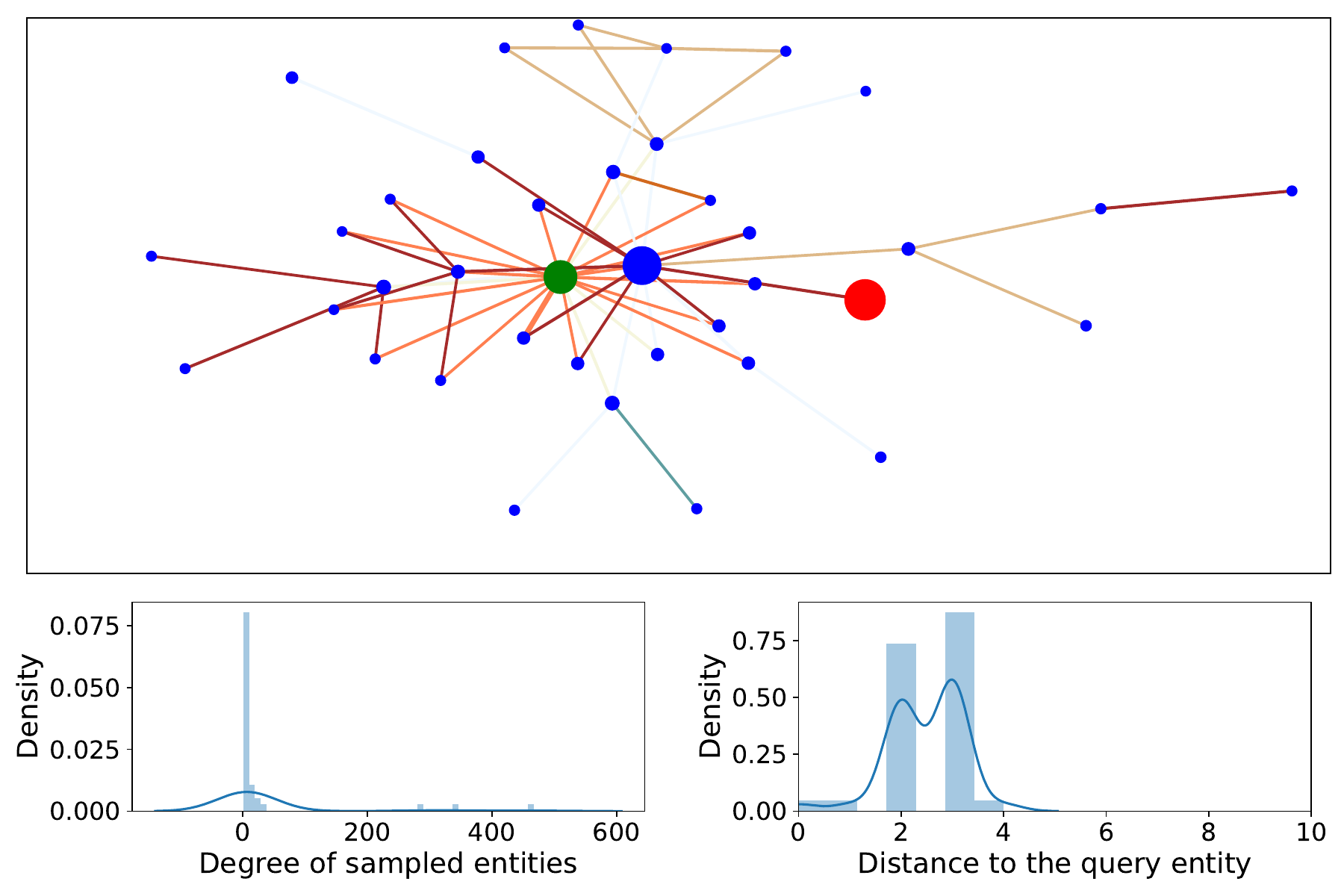}
		\hfill
		\includegraphics[width=6.8cm]{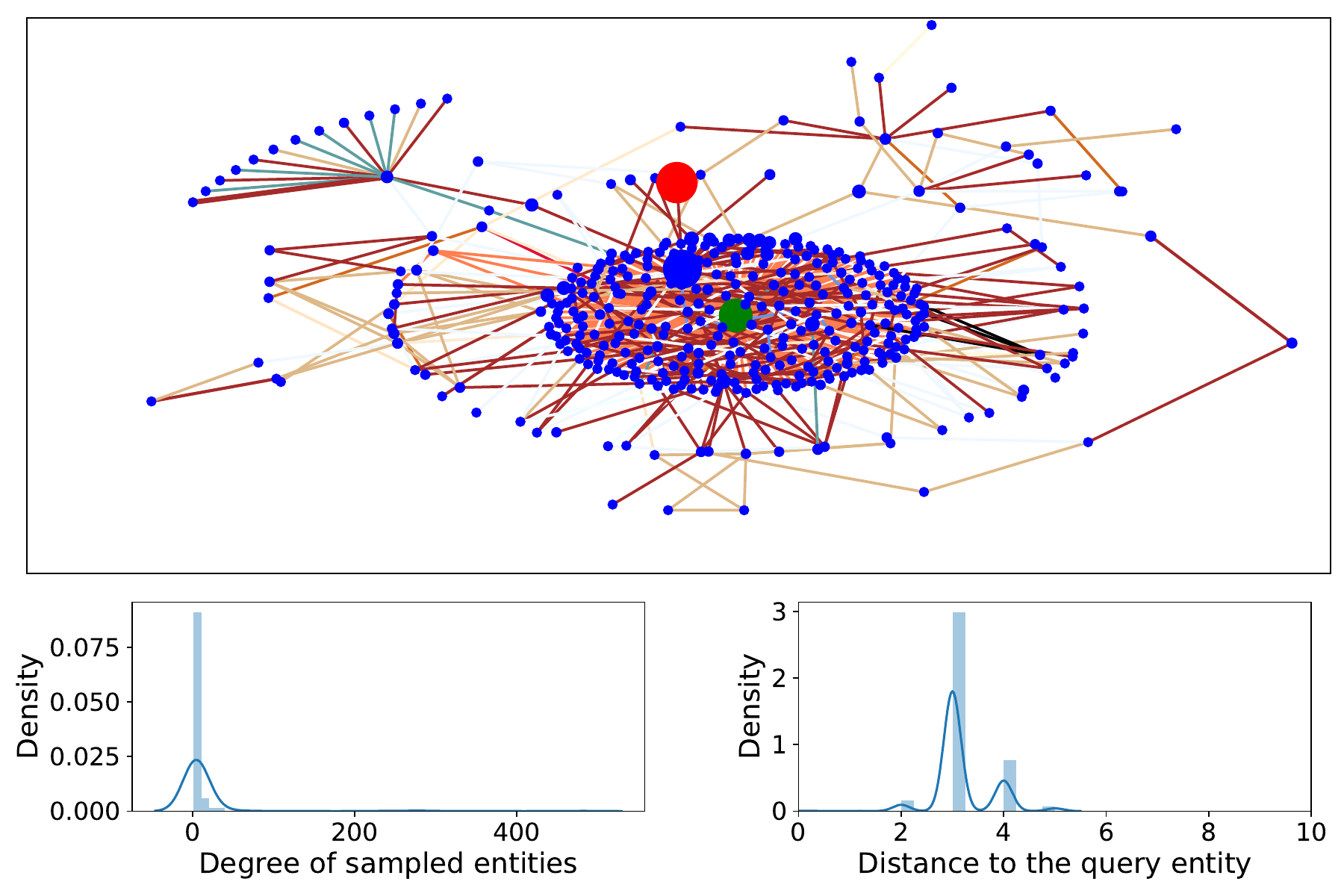}
		\hfill
		\vspace{-8px}
		\caption{
			Subgraphs (0.1\% and 1\%) from WN18RR:
			$u \! = \! 78, q \! = \! 5, v \! = \! \{ 172 \}$. 
		}
		\vspace{-4px}
	\end{figure*}
	
	\begin{figure*}[ht]
		\centering
		\hfill
		\includegraphics[width=6.8cm]{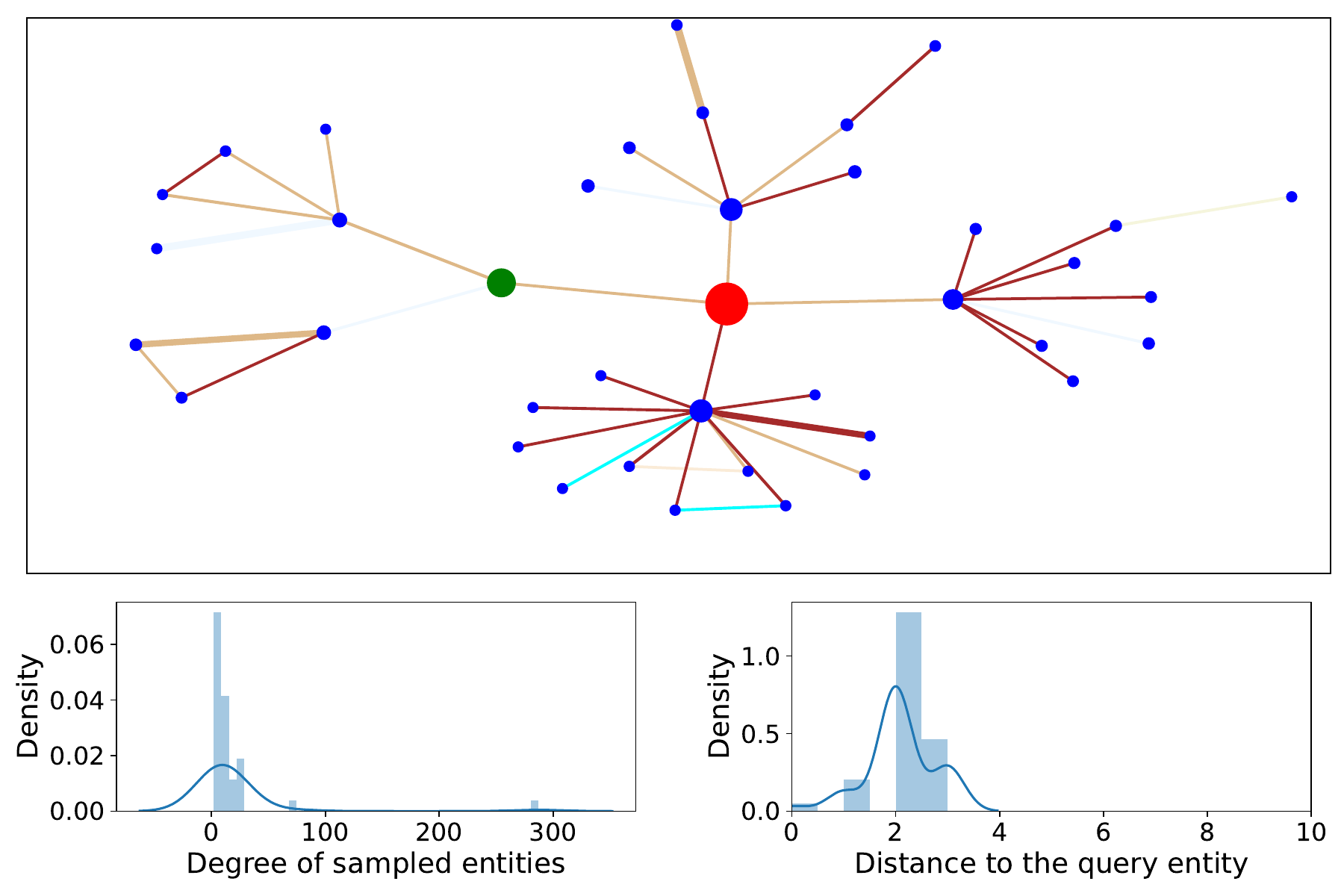}
		\hfill
		\includegraphics[width=6.8cm]{figures/subgraph_data/WN18RR_queryidx_28_ratio_0.01}
		\hfill
		\vspace{-8px}
		\caption{
			Subgraphs (0.1\% and 1\%) from WN18RR:
			$u \! = \! 88, q \! = \! 12, v \! = \! \{ 4621 \}$.
		}
		\vspace{-4px}
	\end{figure*}
	
	\begin{figure*}[ht]
		\centering
		\hfill
		\includegraphics[width=6.8cm]{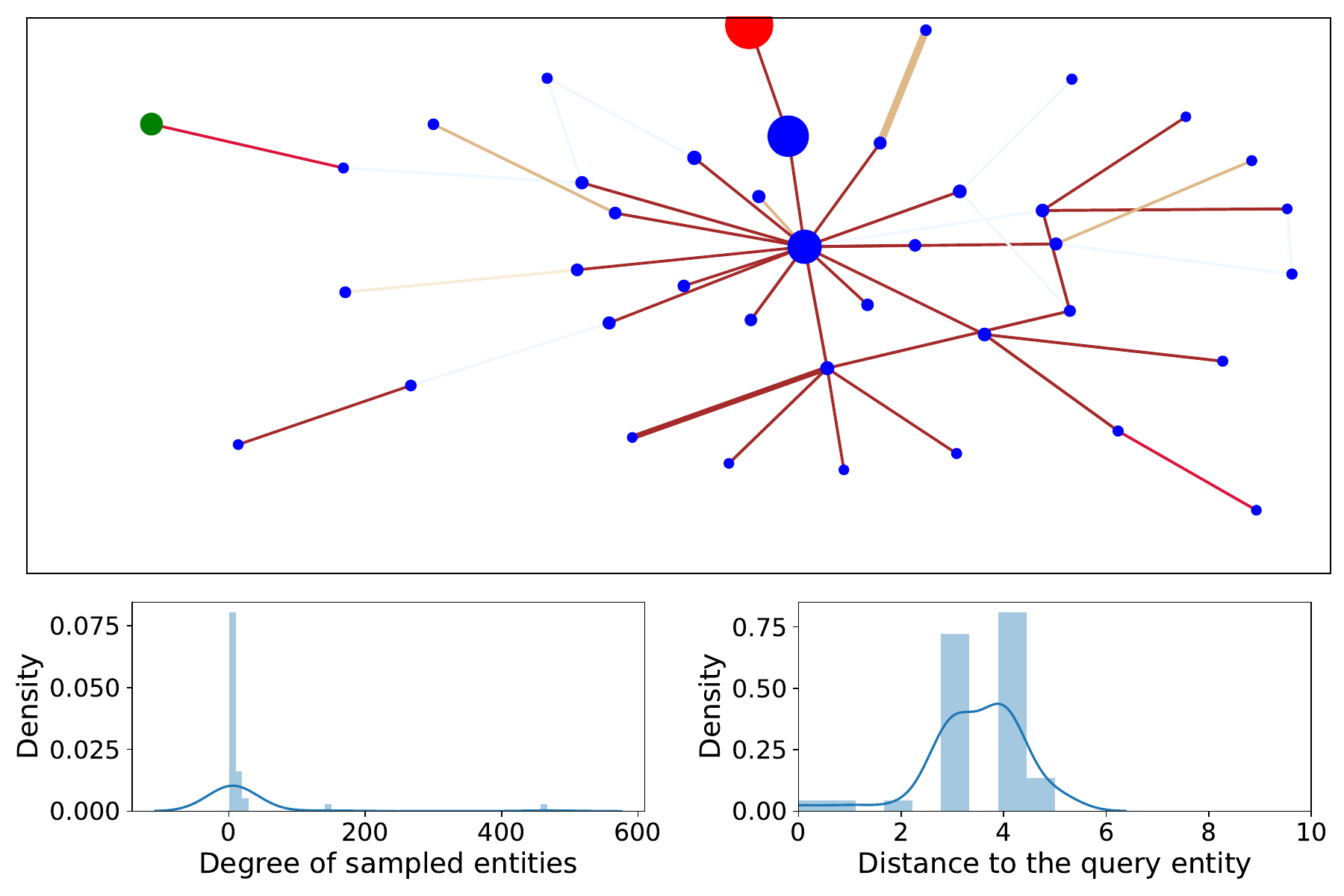}
		\hfill
		\includegraphics[width=6.8cm]{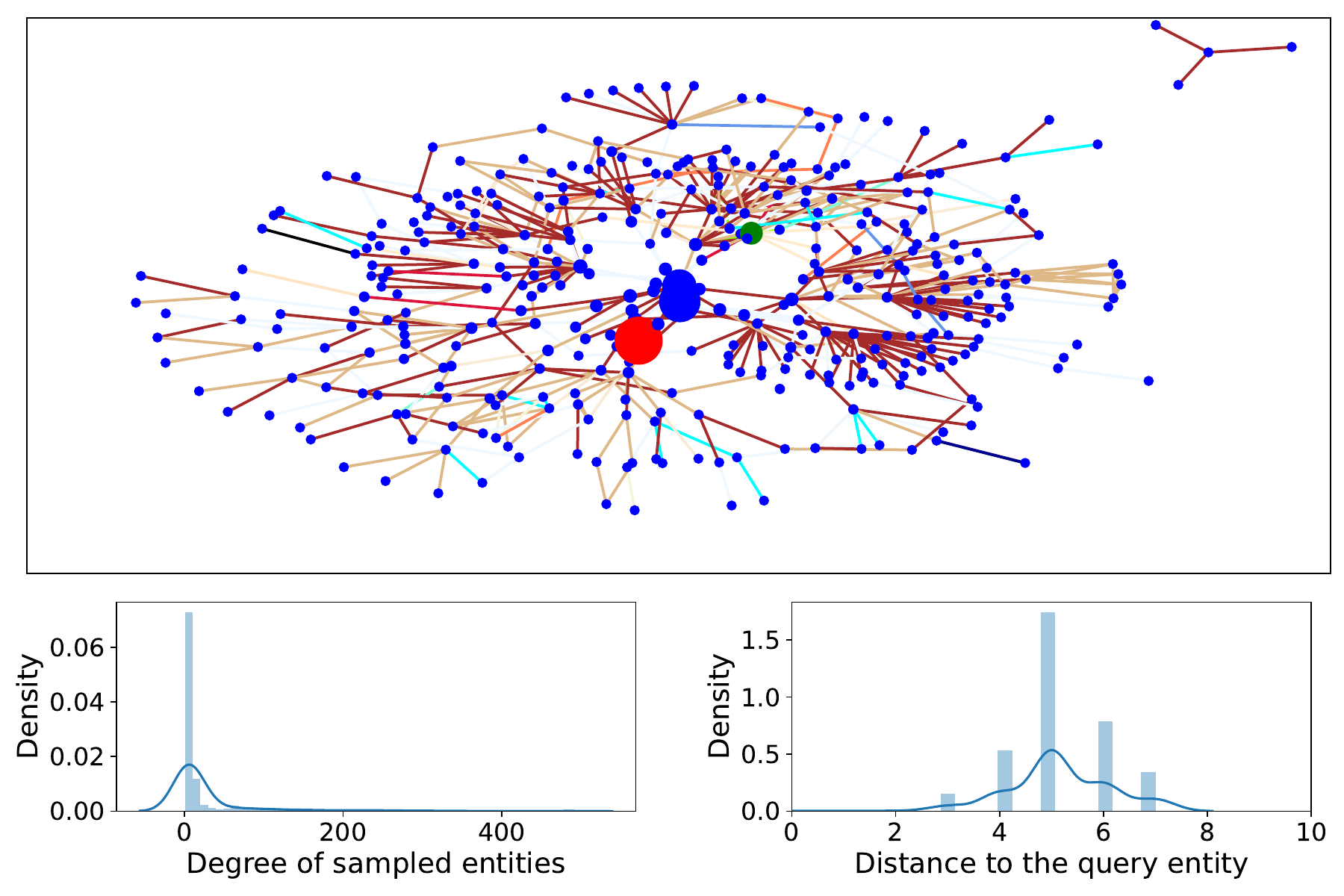}
		\hfill
		\vspace{-8px}
		\caption{
			Subgraphs (0.1\% and 1\%) from WN18RR:
			$u \! = \! 155, q \! = \! 19, v \! = \! \{ 785 \}$. 
		}
		\vspace{-4px}
	\end{figure*}
	
	\begin{figure*}[ht]
		\centering
		\hfill
		\includegraphics[width=6.8cm]{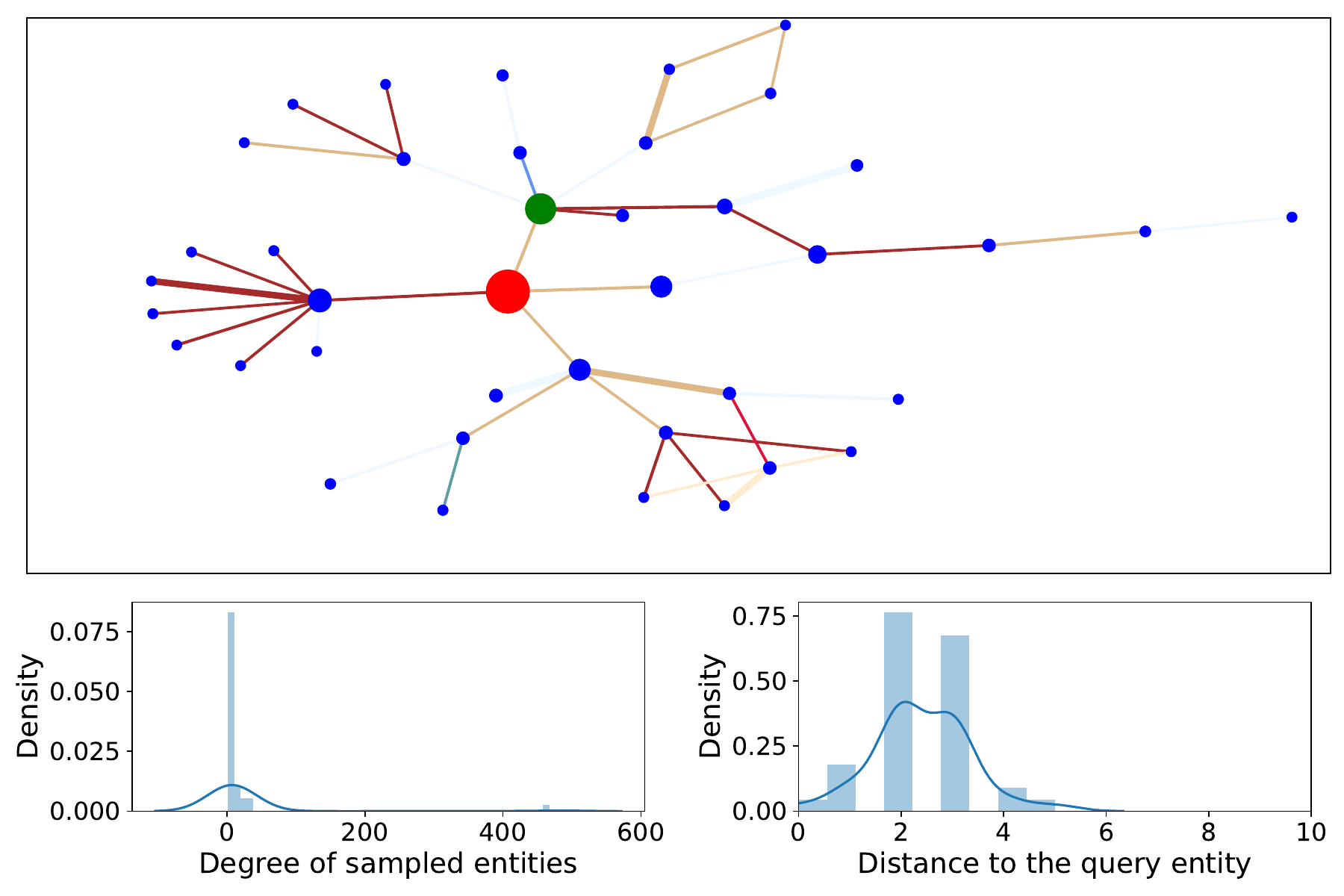}
		\hfill
		\includegraphics[width=6.8cm]{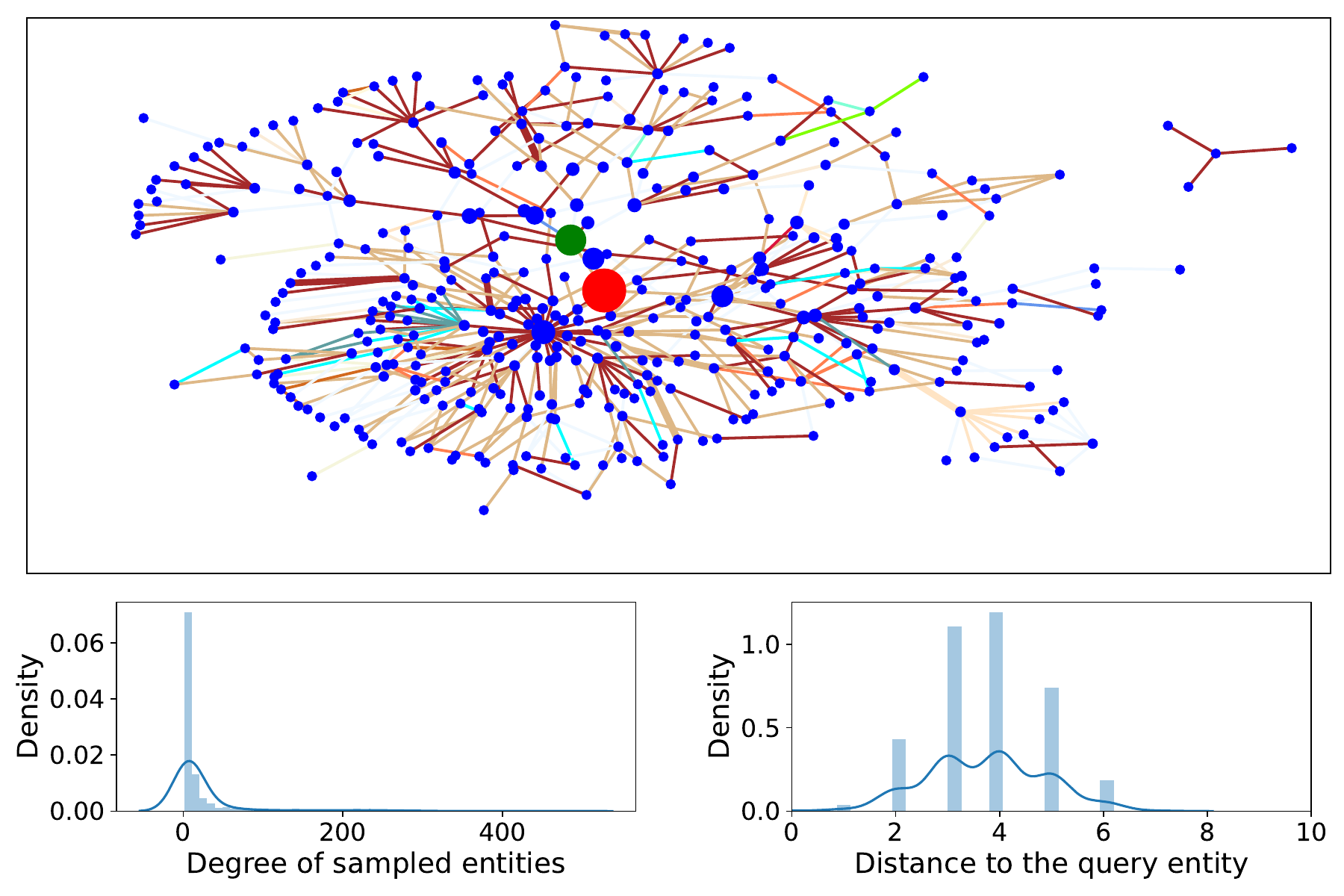}
		\hfill
		\vspace{-8px}
		\caption{
			Subgraphs (0.1\% and 1\%) from WN18RR:
			$u \! = \! 3297, q \! = \! 1, v \! = \! \{ 2037 \}$. 
		}
		\vspace{-4px}
	\end{figure*}
	
	
	\begin{figure*}[ht]
		\centering
		\hfill
		\includegraphics[width=6.8cm]{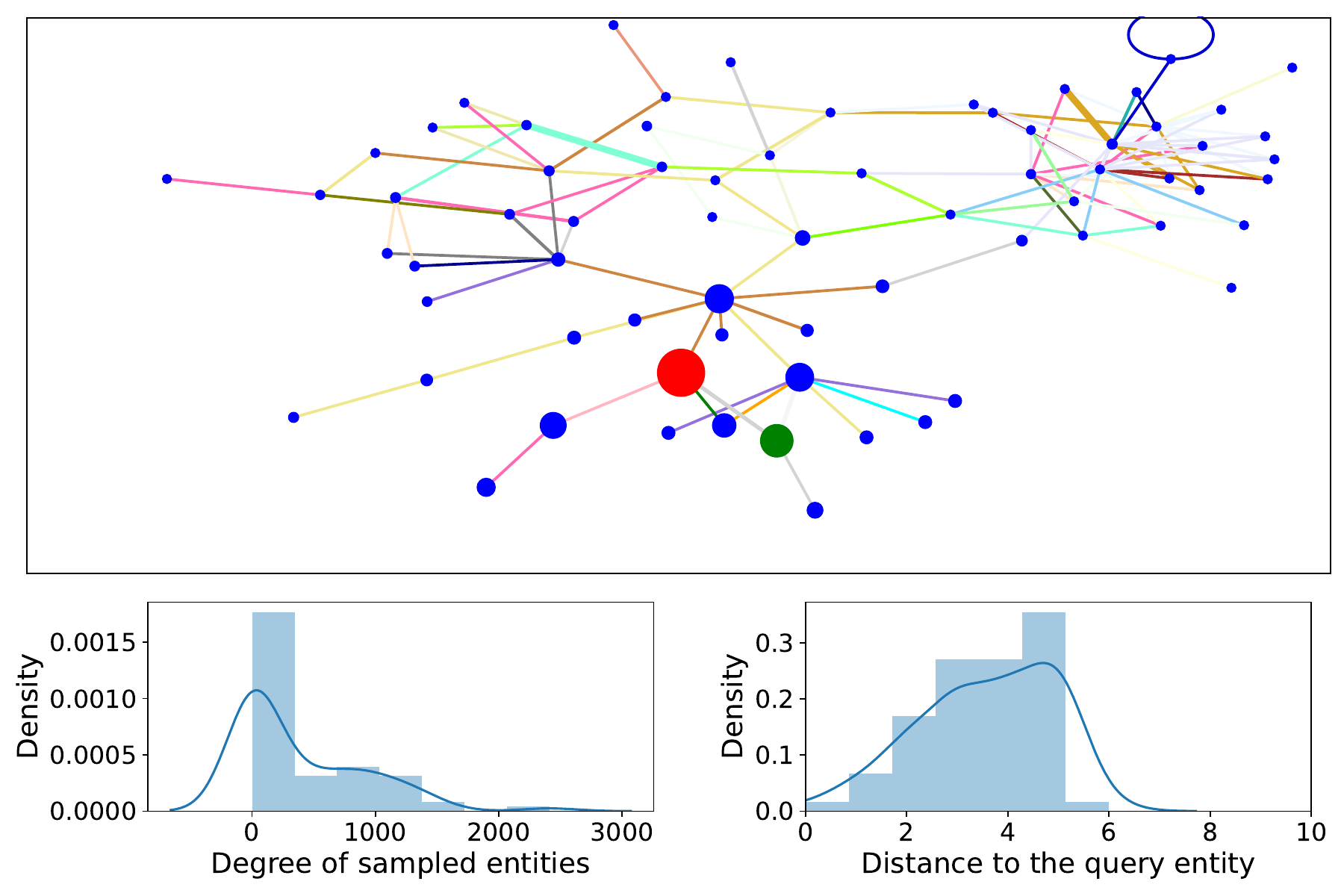}
		\hfill
		\includegraphics[width=6.8cm]{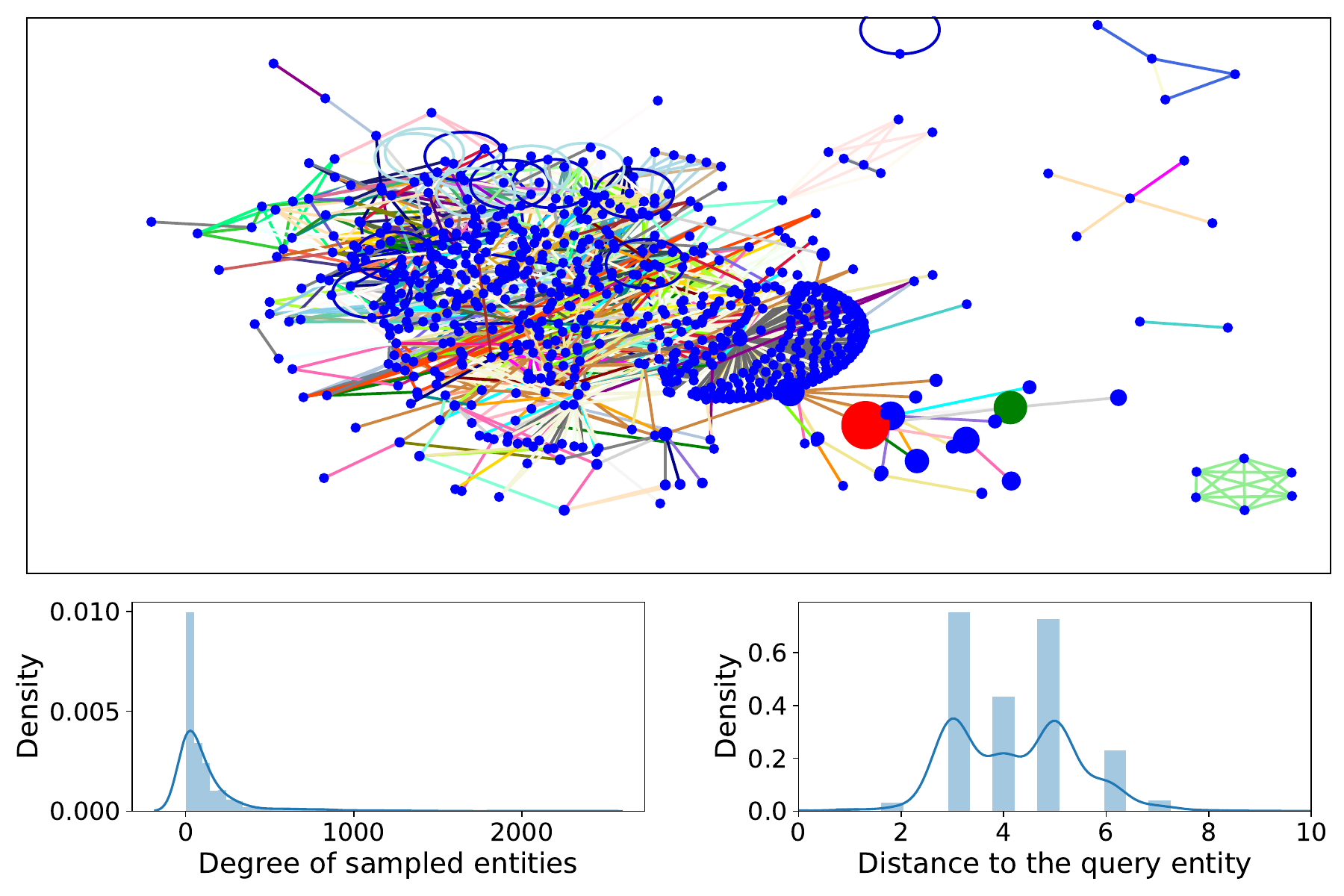}
		\hfill
		\vspace{-8px}
		\caption{
			Subgraphs (0.1\% and 1\%) from NELL-995:
			$u \! = \! 4, q \! = \! 238, v \! = \! \{22677\}$.
		}
		\vspace{-4px}
	\end{figure*}

	\begin{figure*}[ht]
		\centering
		\hfill
		\includegraphics[width=6.8cm]{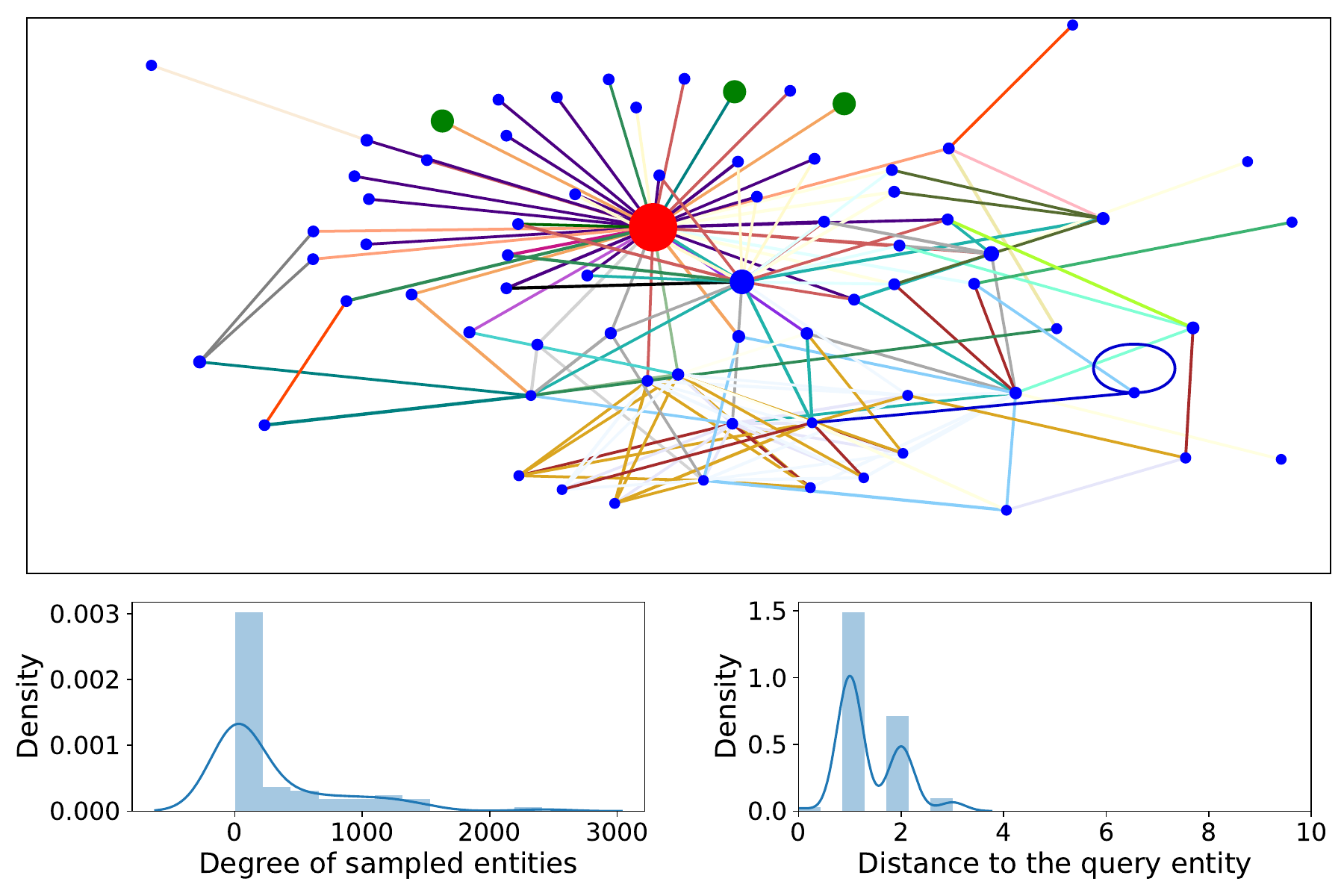}
		\hfill
		\includegraphics[width=6.8cm]{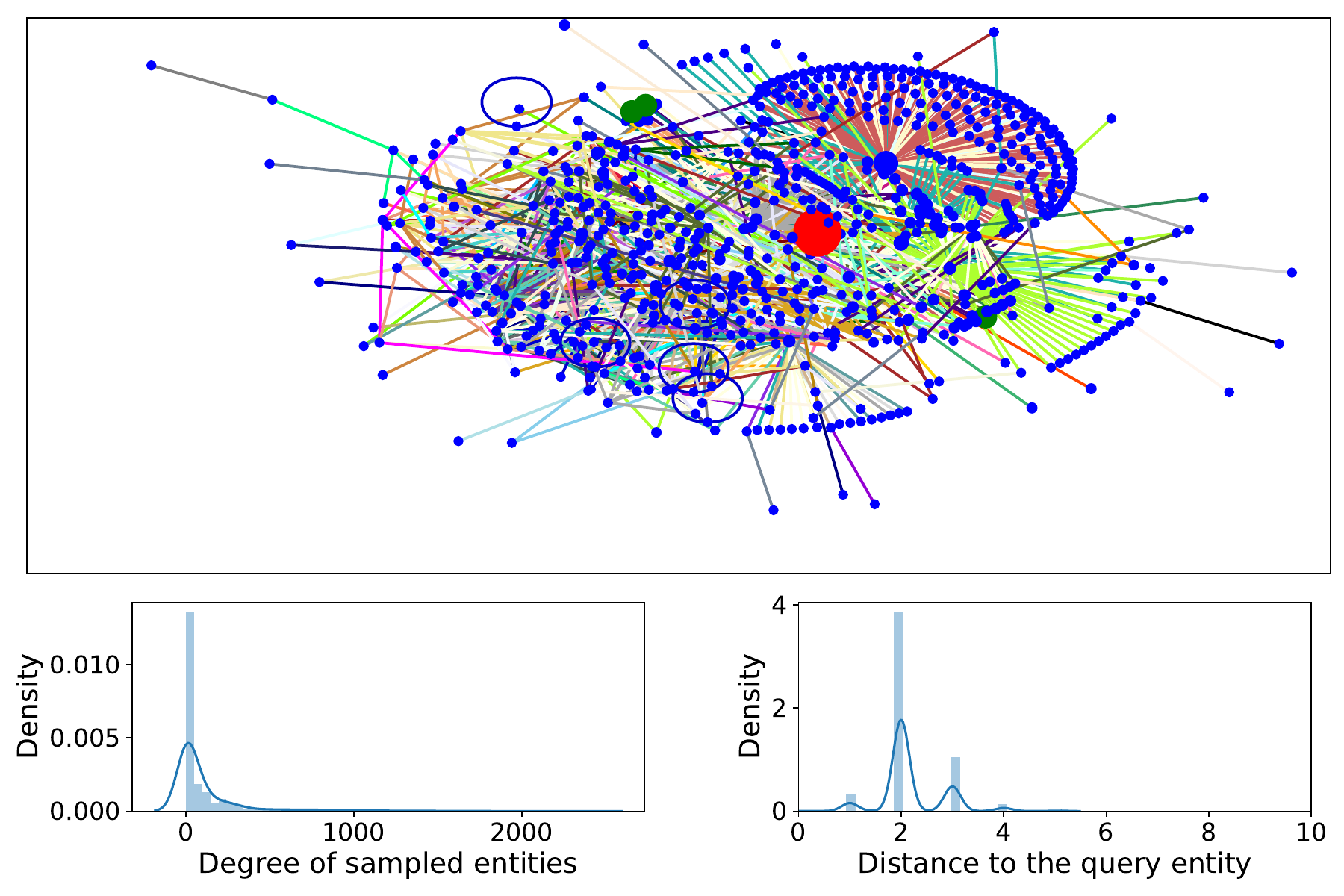}
		\hfill
		\vspace{-8px}
		\caption{
			Subgraphs (0.1\% and 1\%) from NELL-995:
			$u \! = \! 17, q \! = \! 274, v \! = \! \{ 57735, 61381, 63044 \}$.
		}
		\vspace{-4px}
	\end{figure*}

	\begin{figure*}[ht]
		\centering
		\hfill
		\includegraphics[width=6.8cm]{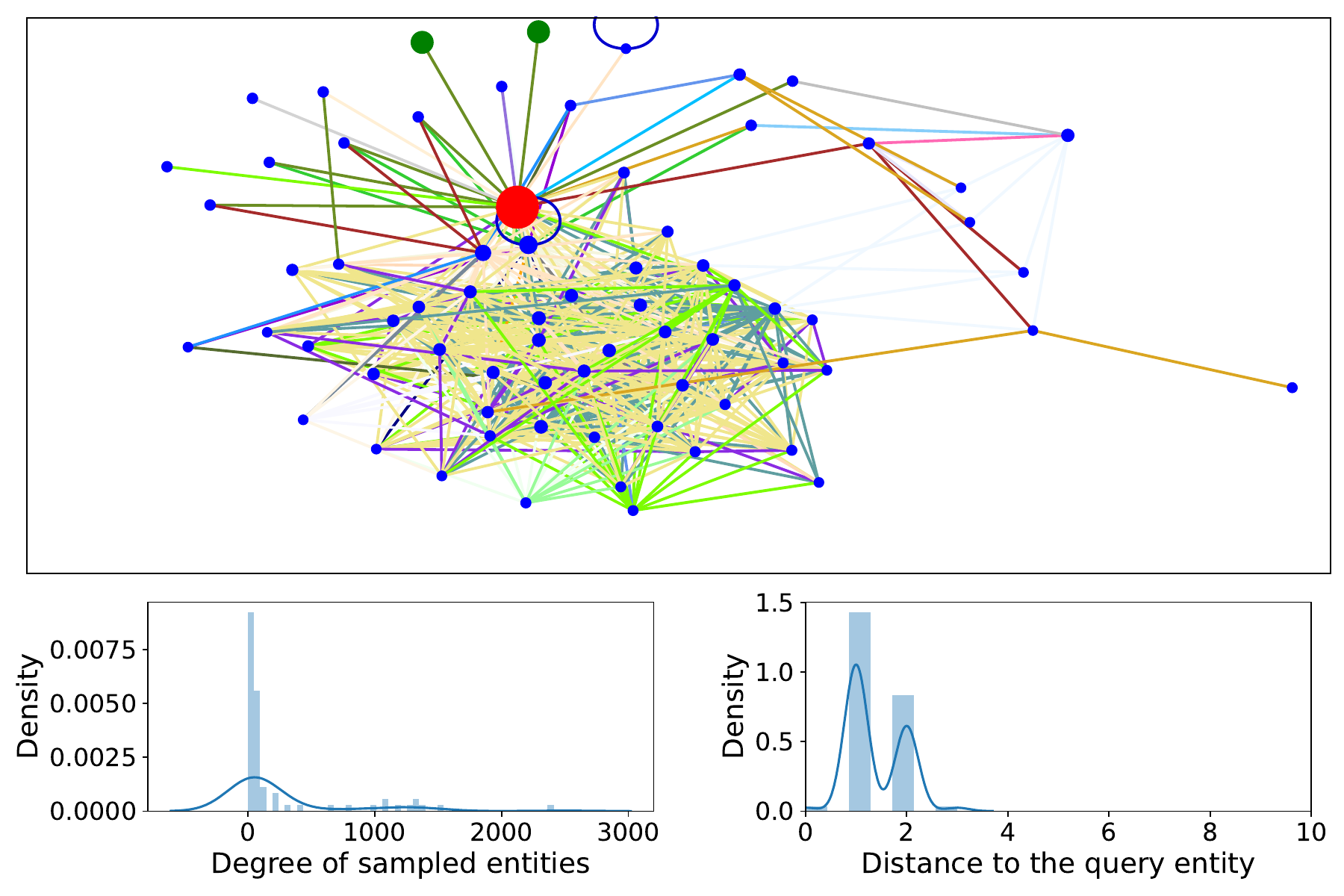}
		\hfill
		\includegraphics[width=6.8cm]{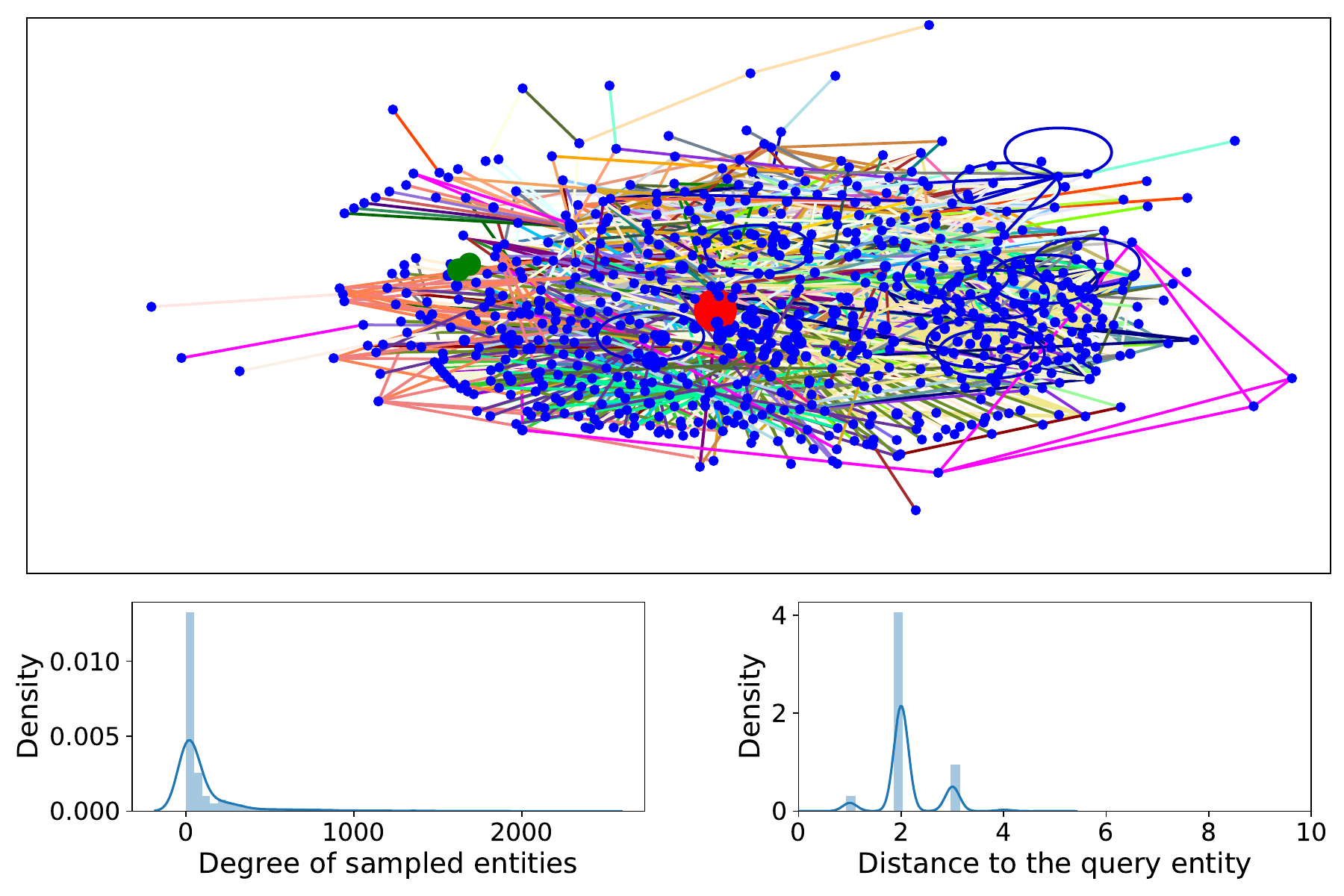}
		\hfill
		\vspace{-8px}
		\caption{
			Subgraphs (0.1\% and 1\%) from NELL-995:
			$u \! = \! 29, q \! = \! 260, v \! = \! \{ 27725, 73985 \}$.
		}
		\vspace{-4px}
	\end{figure*}
	
	\begin{figure*}[ht]
		\centering
		\hfill
		\includegraphics[width=6.8cm]{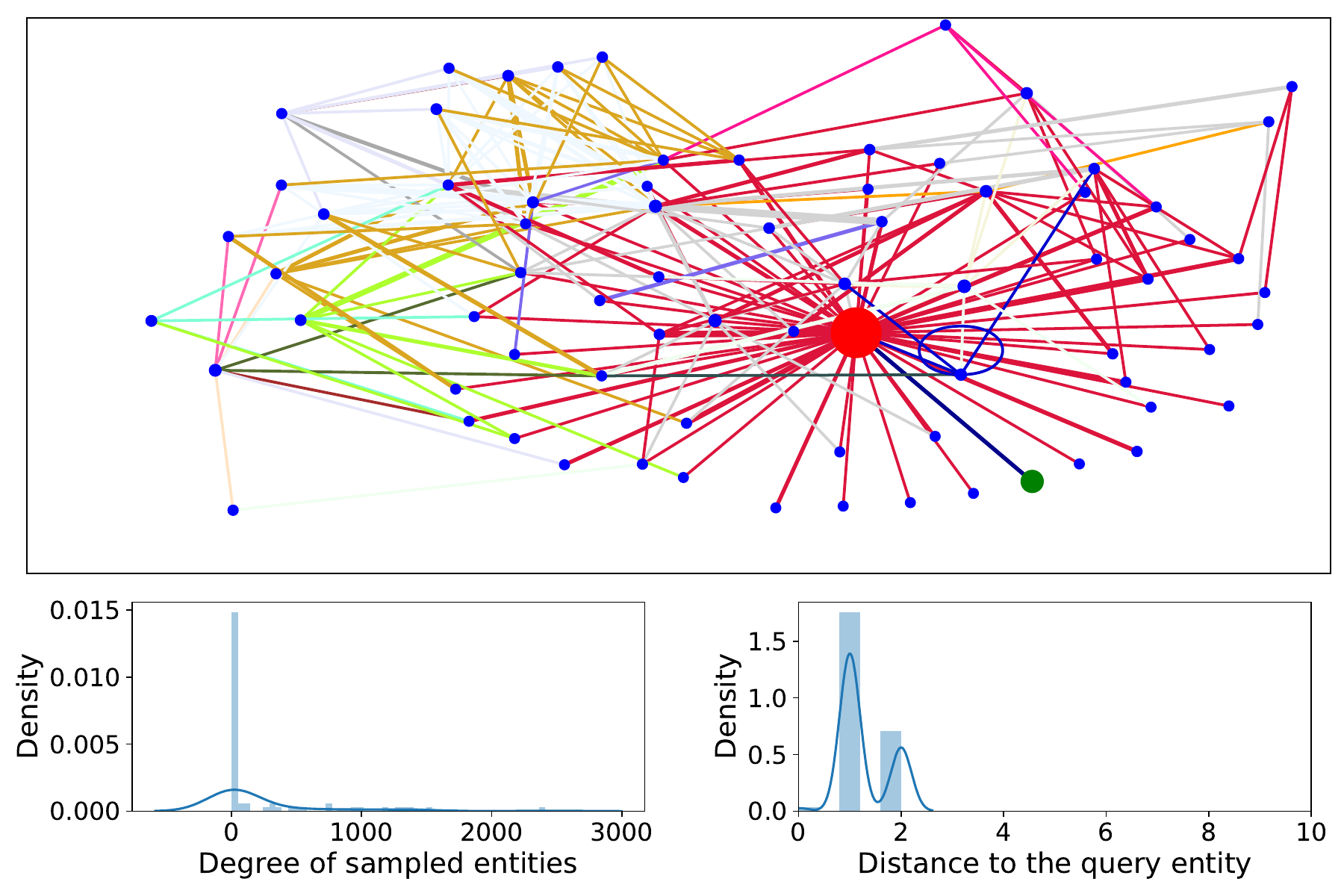}
		\hfill
		\includegraphics[width=6.8cm]{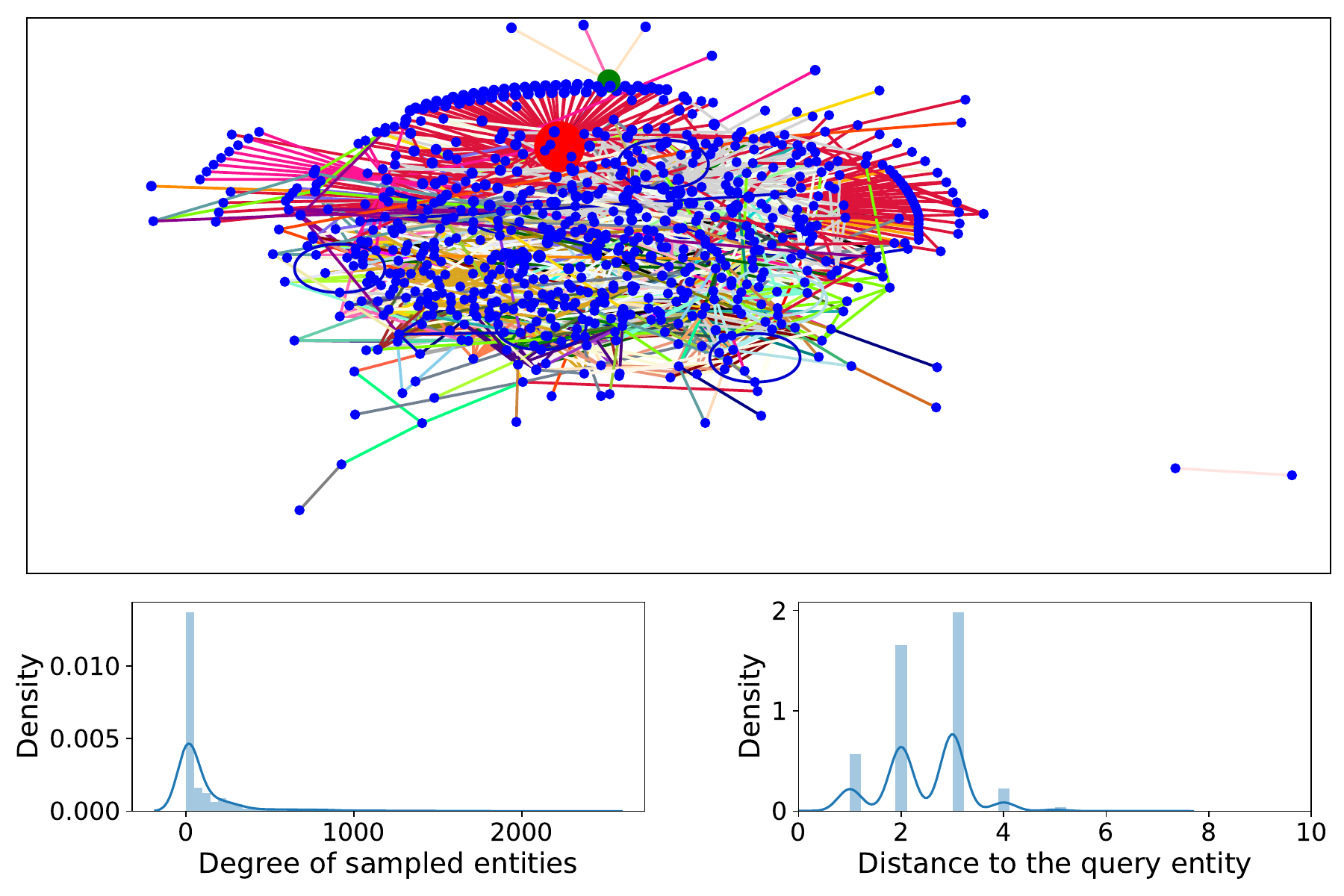}
		\hfill
		\vspace{-8px}
		\caption{
			Subgraphs (0.1\% and 1\%) from NELL-995:
			$u \! = \! 44, q \! = \! 222, v \! = \! \{ 11669 \}$.
		}
		\vspace{-4px}
	\end{figure*}

	\begin{figure*}[ht]
		\centering
		\hfill
		\includegraphics[width=6.8cm]{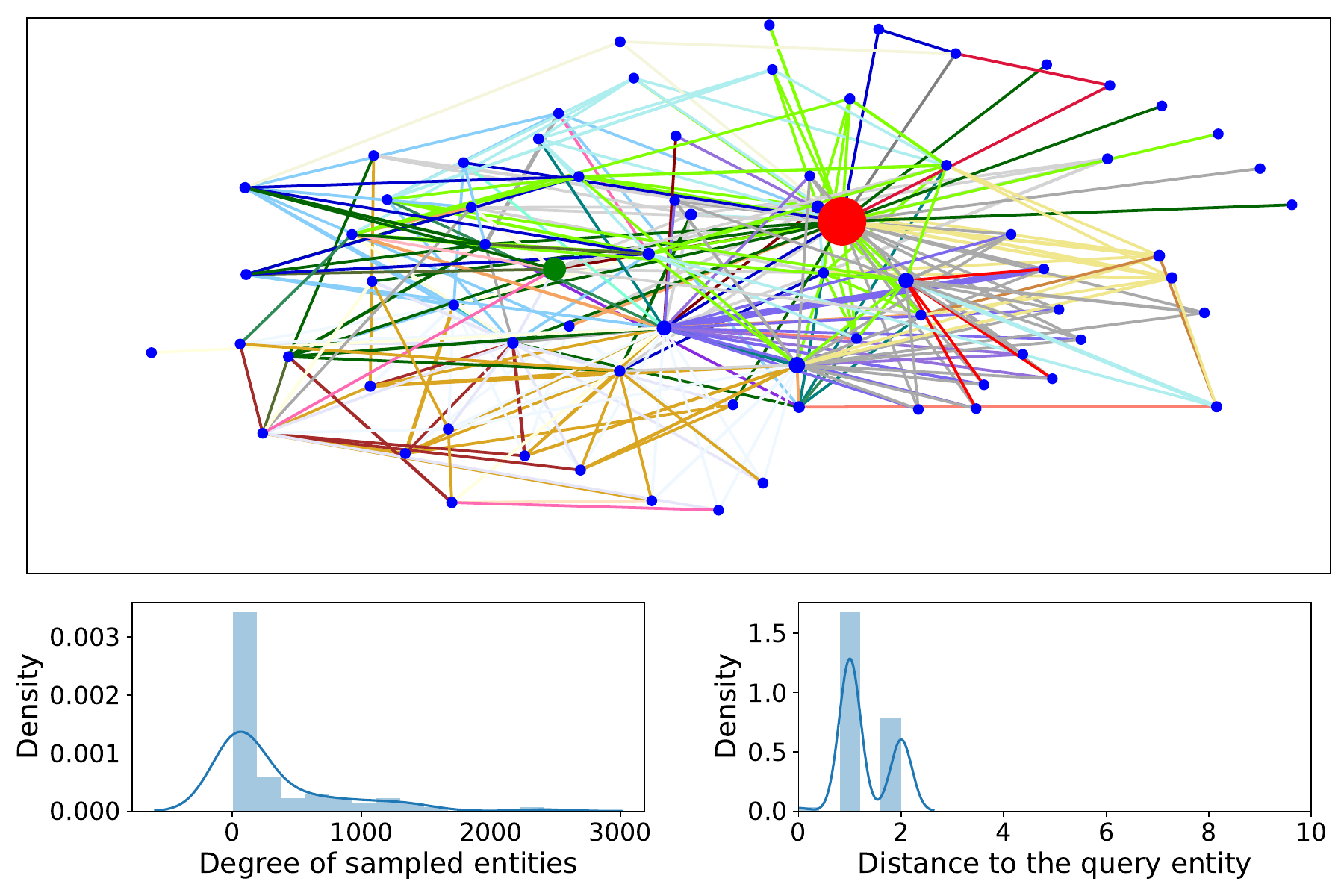}
		\hfill
		\includegraphics[width=6.8cm]{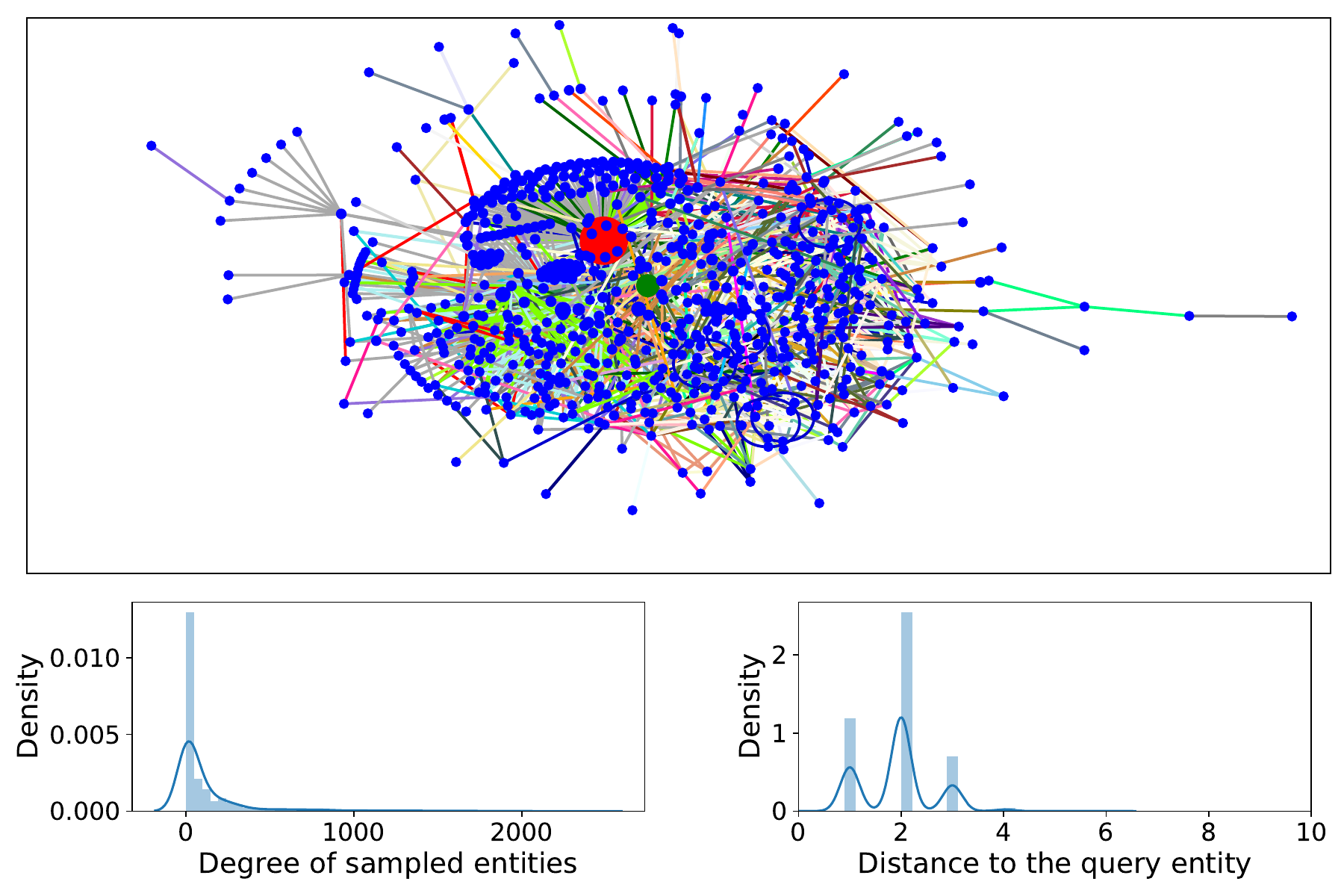}
		\hfill
		\vspace{-8px}
		\caption{
			Subgraphs (0.1\% and 1\%) from NELL-995:
			$u \! = \! 60, q \! = \! 74, v \! = \! \{ 164 \}$.
		}
		\vspace{-4px}
	\end{figure*}

	\begin{figure*}[ht]
		\centering
		\hfill
		\includegraphics[width=6.8cm]{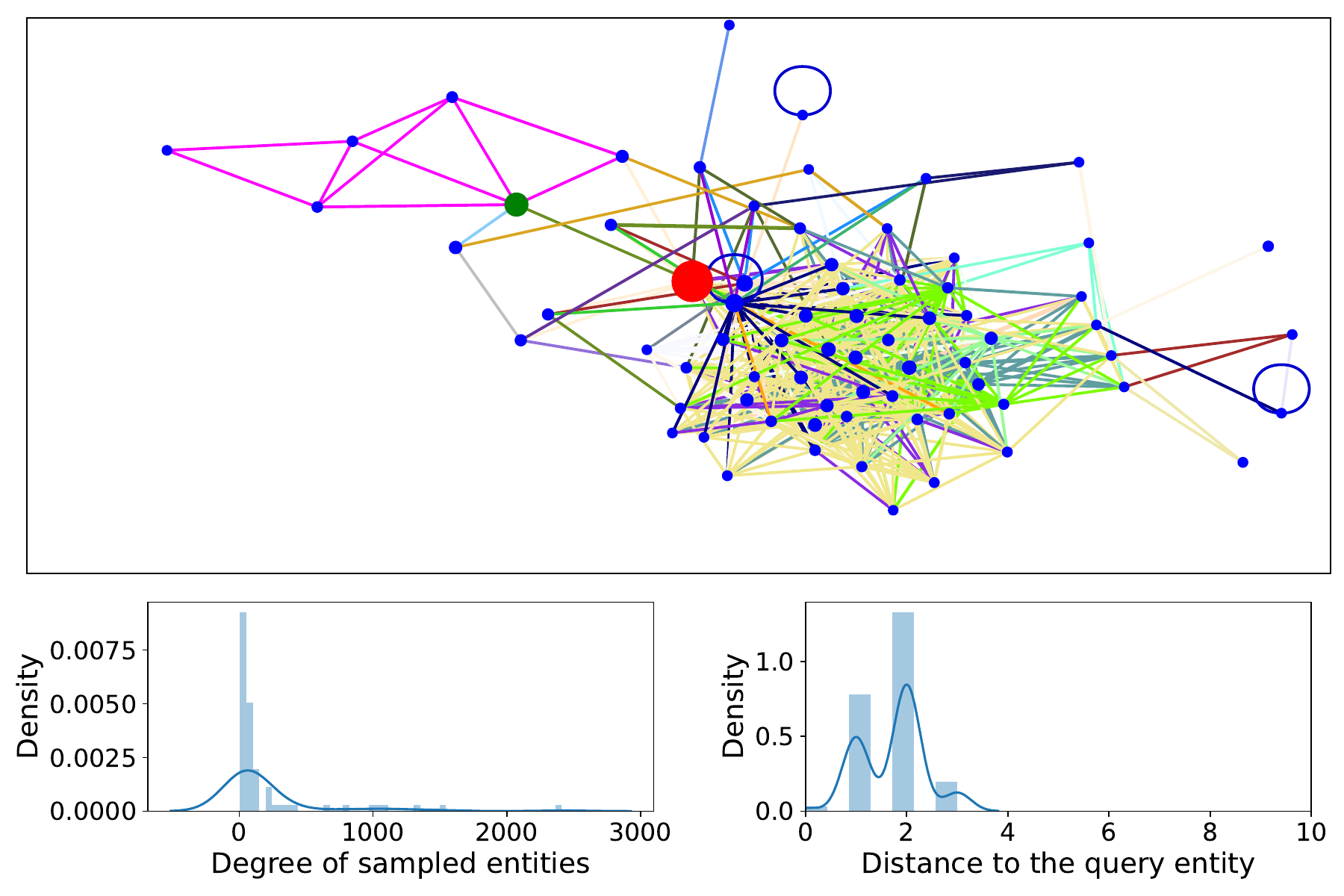}
		\hfill
		\includegraphics[width=6.8cm]{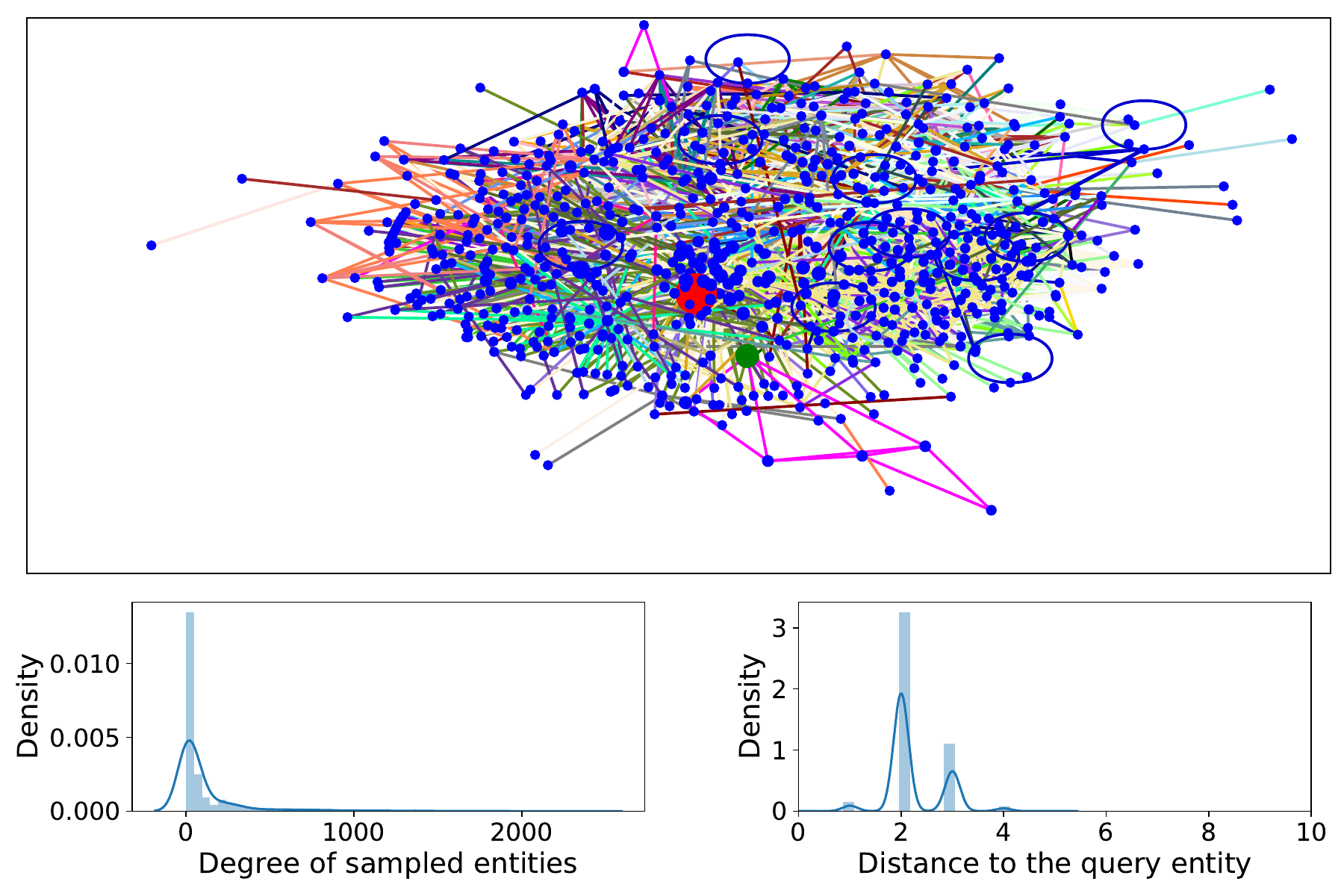}
		\hfill
		\vspace{-8px}
		\caption{
			Subgraphs (0.1\% and 1\%) from NELL-995:
			$u \! = \! 166, q \! = \! 260, v \! = \! \{ 6364 \}$.
		}
		\vspace{-4px}
	\end{figure*}

	\begin{figure*}[ht]
		\centering
		\hfill
		\includegraphics[width=6.8cm]{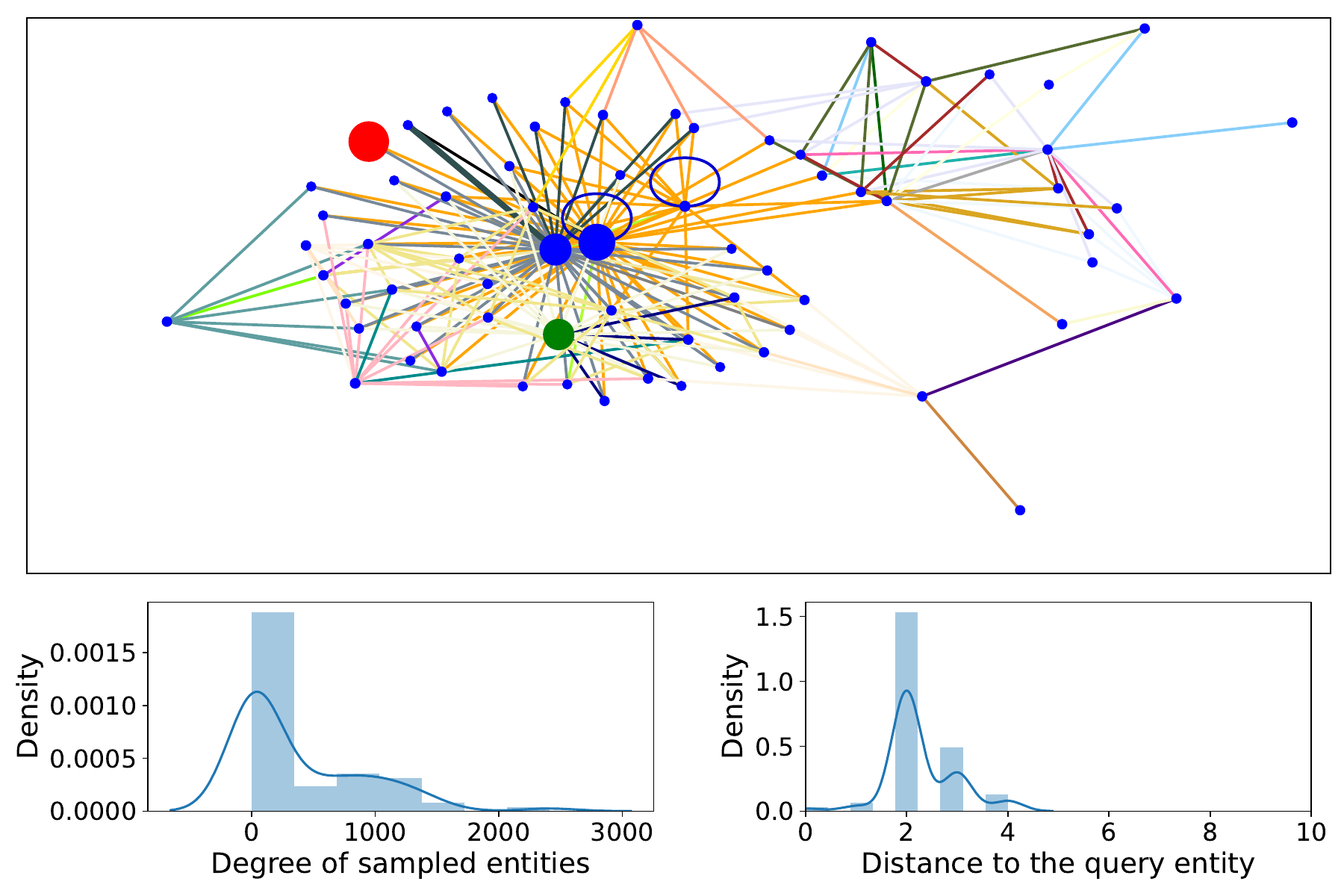}
		\hfill
		\includegraphics[width=6.8cm]{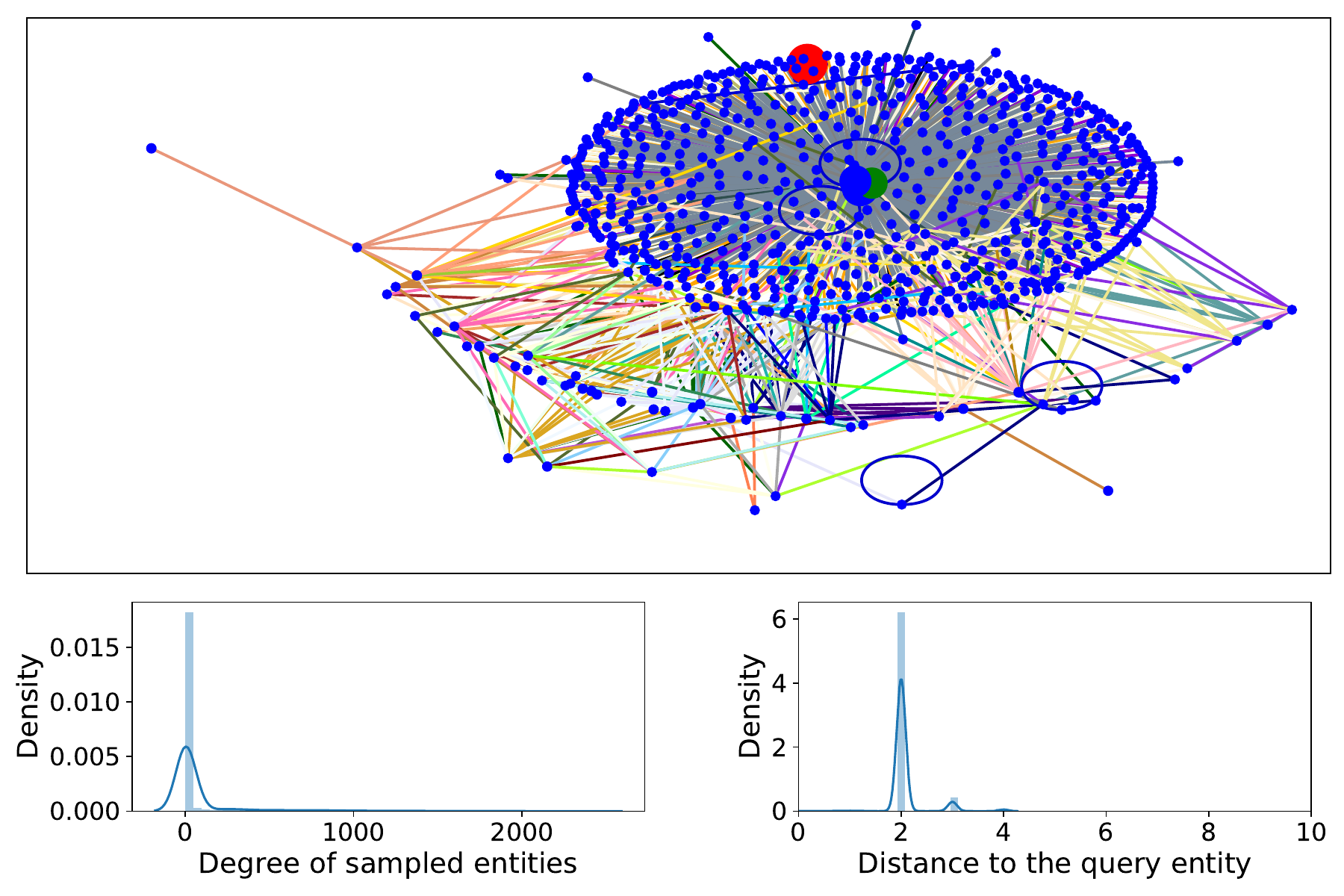}
		\hfill
		\vspace{-8px}
		\caption{
			Subgraphs (0.1\% and 1\%) from NELL-995:
			$u \! = \! 202, q \! = \! 101, v \! = \! \{ 399 \}$.
		}
		\vspace{-4px}
	\end{figure*}
	
	\begin{figure*}[ht]
		\centering
		\hfill
		\includegraphics[width=6.8cm]{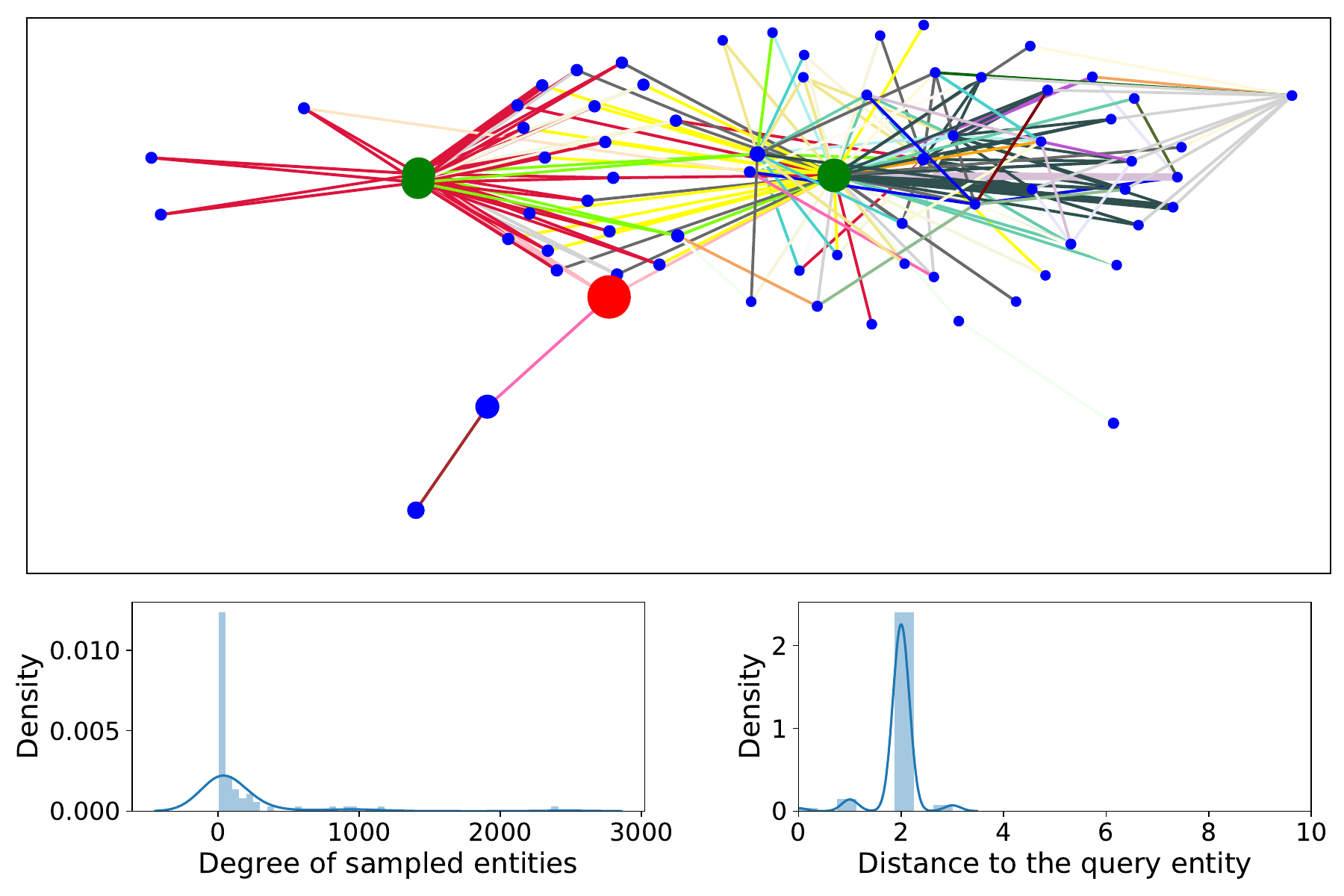}
		\hfill
		\includegraphics[width=6.8cm]{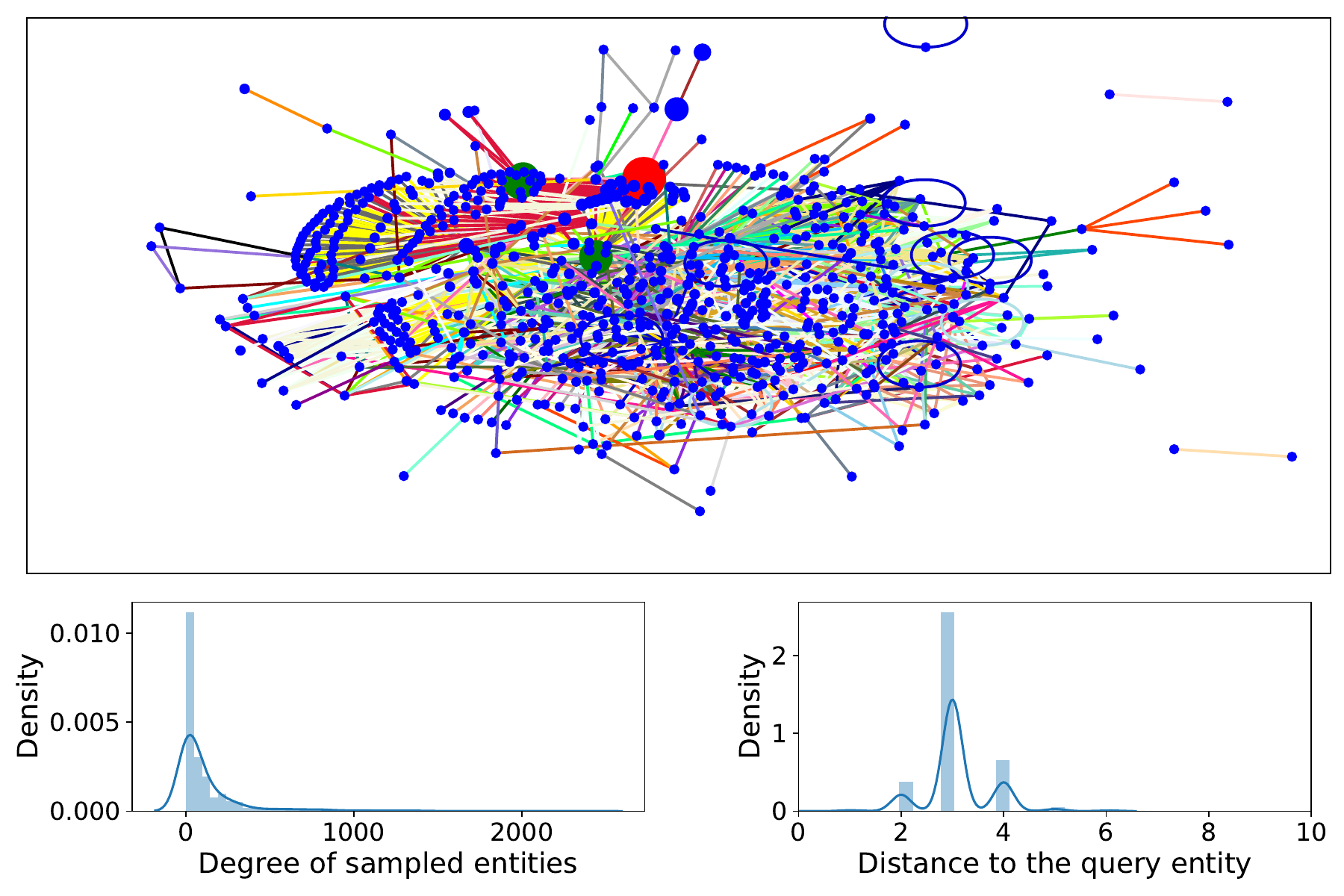}
		\hfill
		\vspace{-8px}
		\caption{
			Subgraphs (0.1\% and 1\%) from NELL-995:
			$u \! = \! 255, q \! = \! 232, v \! = \! \{ 1631, 9925, 11229 \}$.
		}
		\vspace{-4px}
	\end{figure*}
	
	\begin{figure*}[ht]
		\centering
		\hfill
		\includegraphics[width=6.8cm]{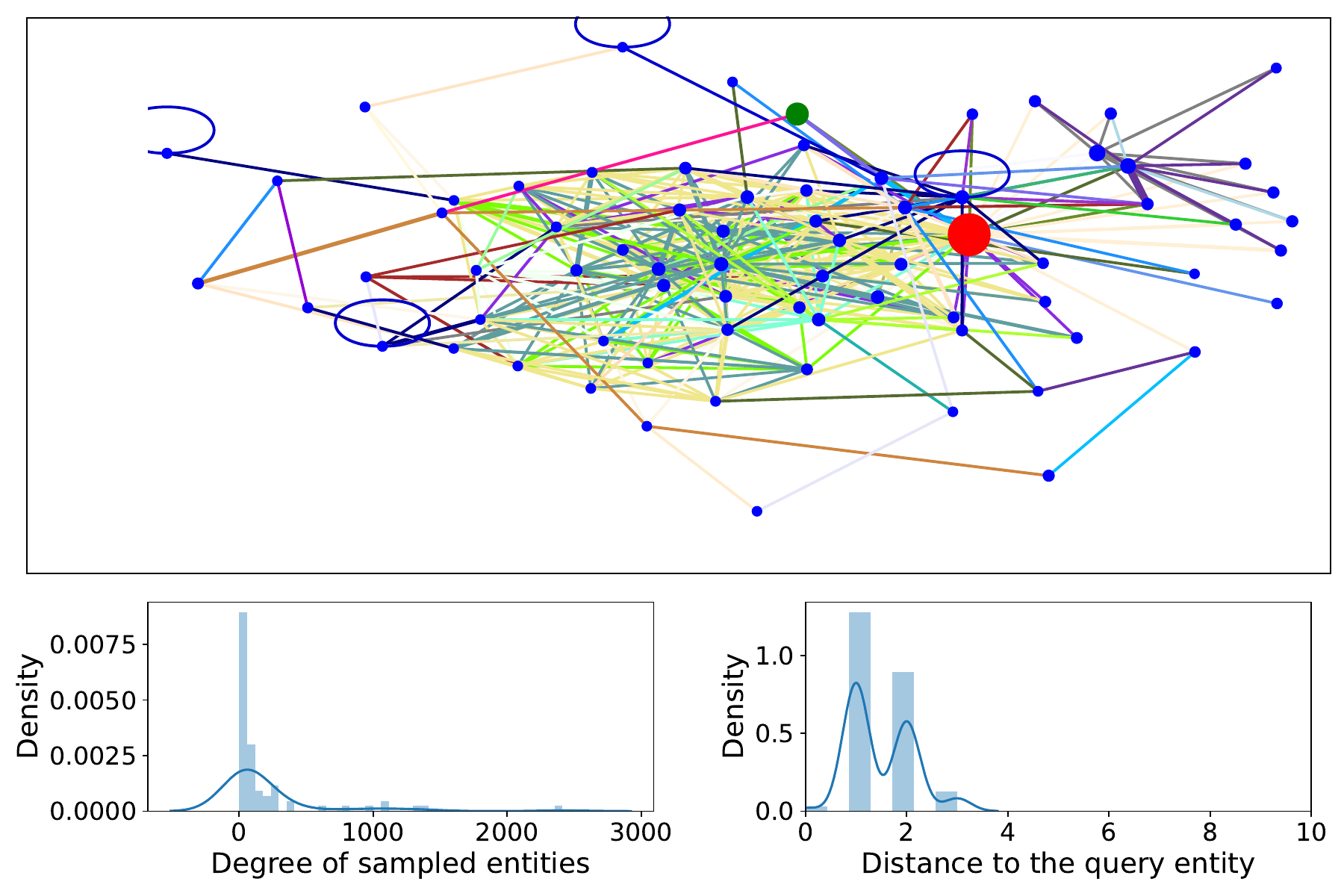}
		\hfill
		\includegraphics[width=6.8cm]{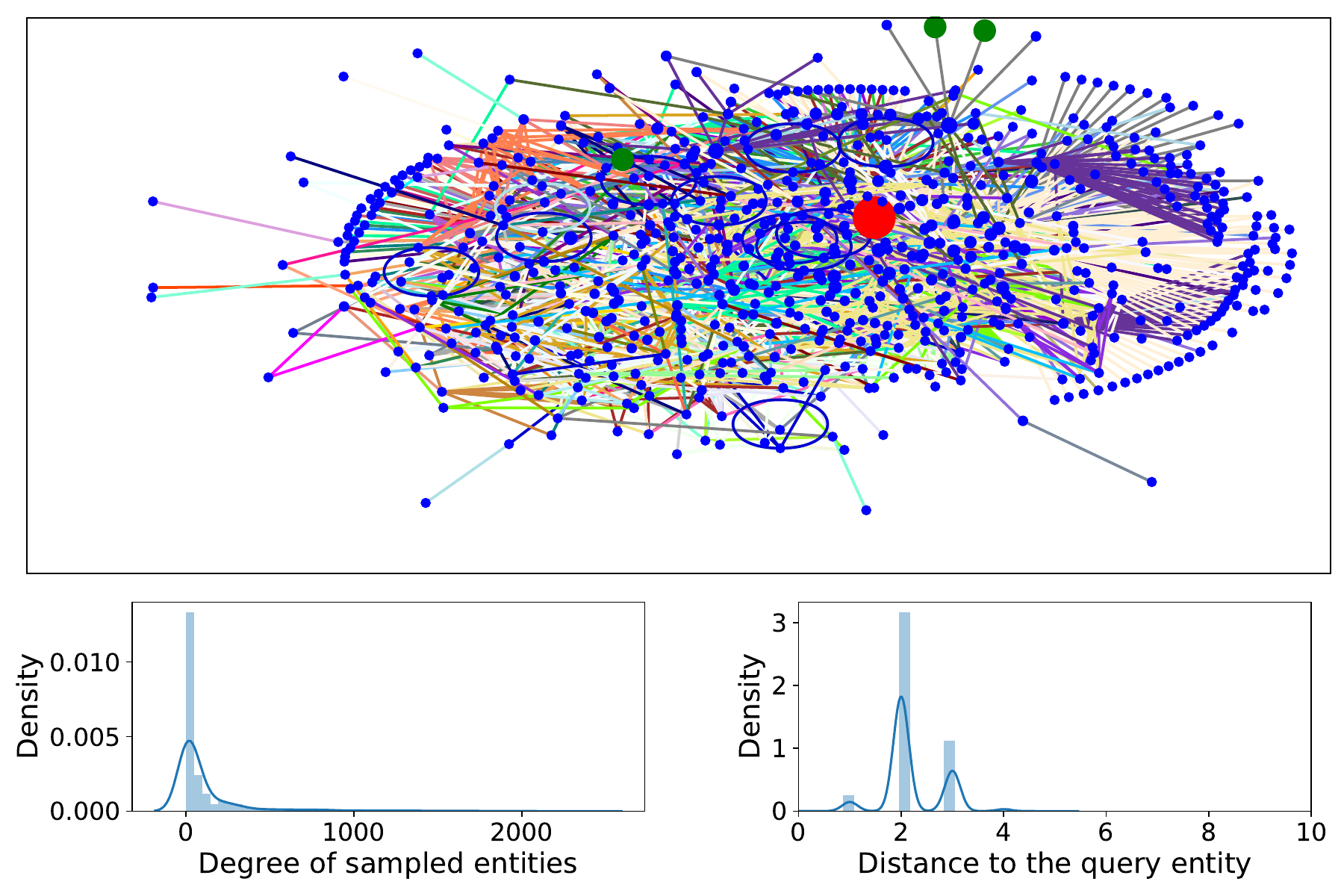}
		\hfill
		\vspace{-8px}
		\caption{
			Subgraphs (0.1\% and 1\%) from NELL-995:
			$u \! \! = \! \! 1371, q \! \! = \!  \!  260, v \! \! = \! \! \{ 24193, 50385, 60718 \}$.
		}
		\vspace{-4px}
	\end{figure*}
	
	\begin{figure*}[ht]
		\centering
		\hfill
		\includegraphics[width=6.8cm]{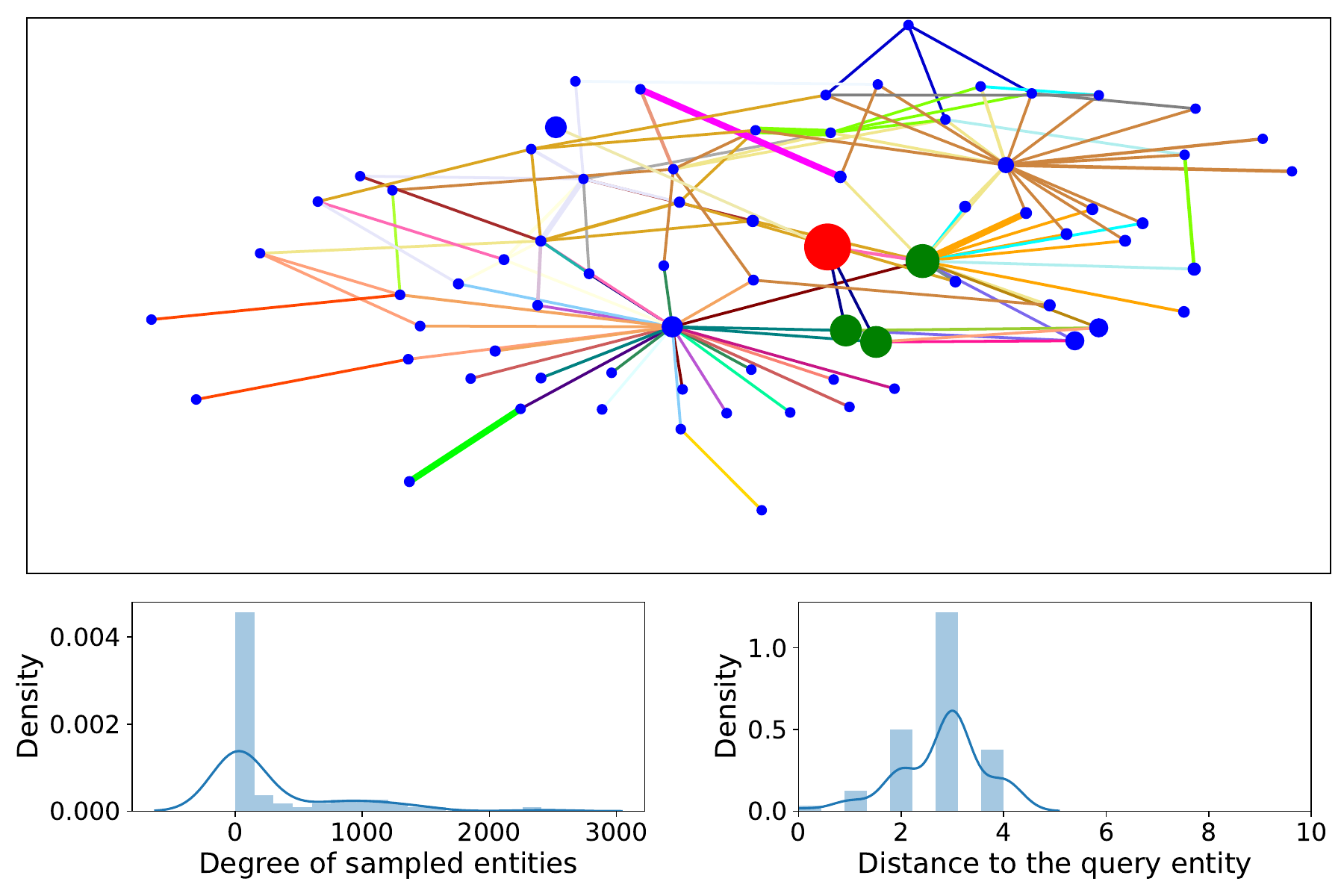}
		\hfill
		\includegraphics[width=6.8cm]{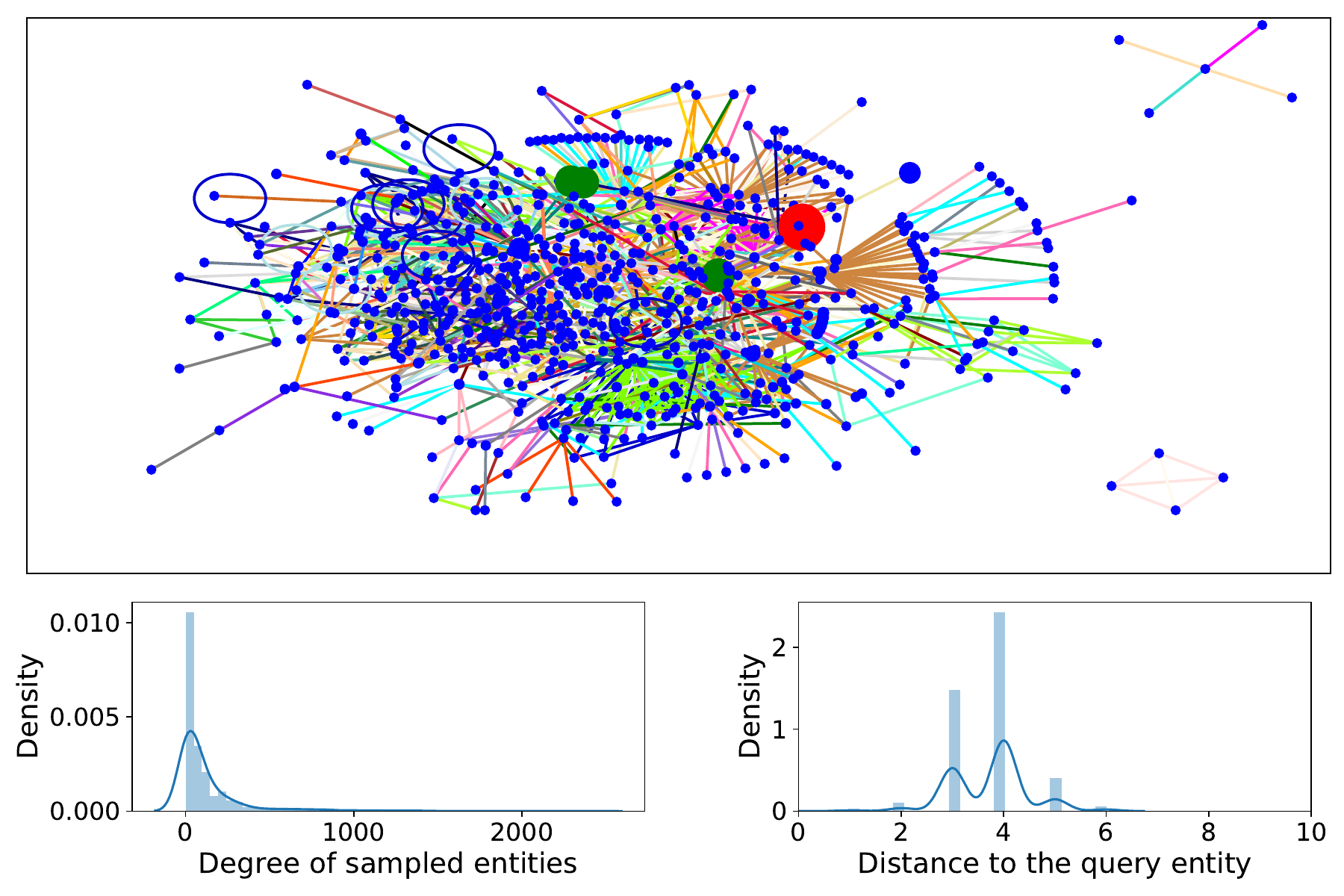}
		\hfill
		\vspace{-8px}
		\caption{
			Subgraphs (0.1\% and 1\%) from NELL-995:
			$u \! = \! 11200, q \! = \! 38, v \! = \! \{ 5737, 7292, 11199 \}$. 
		}
		\vspace{-4px}
	\end{figure*}
	
	
	\begin{figure*}[ht]
		\centering
		\hfill
		\includegraphics[width=6.8cm]{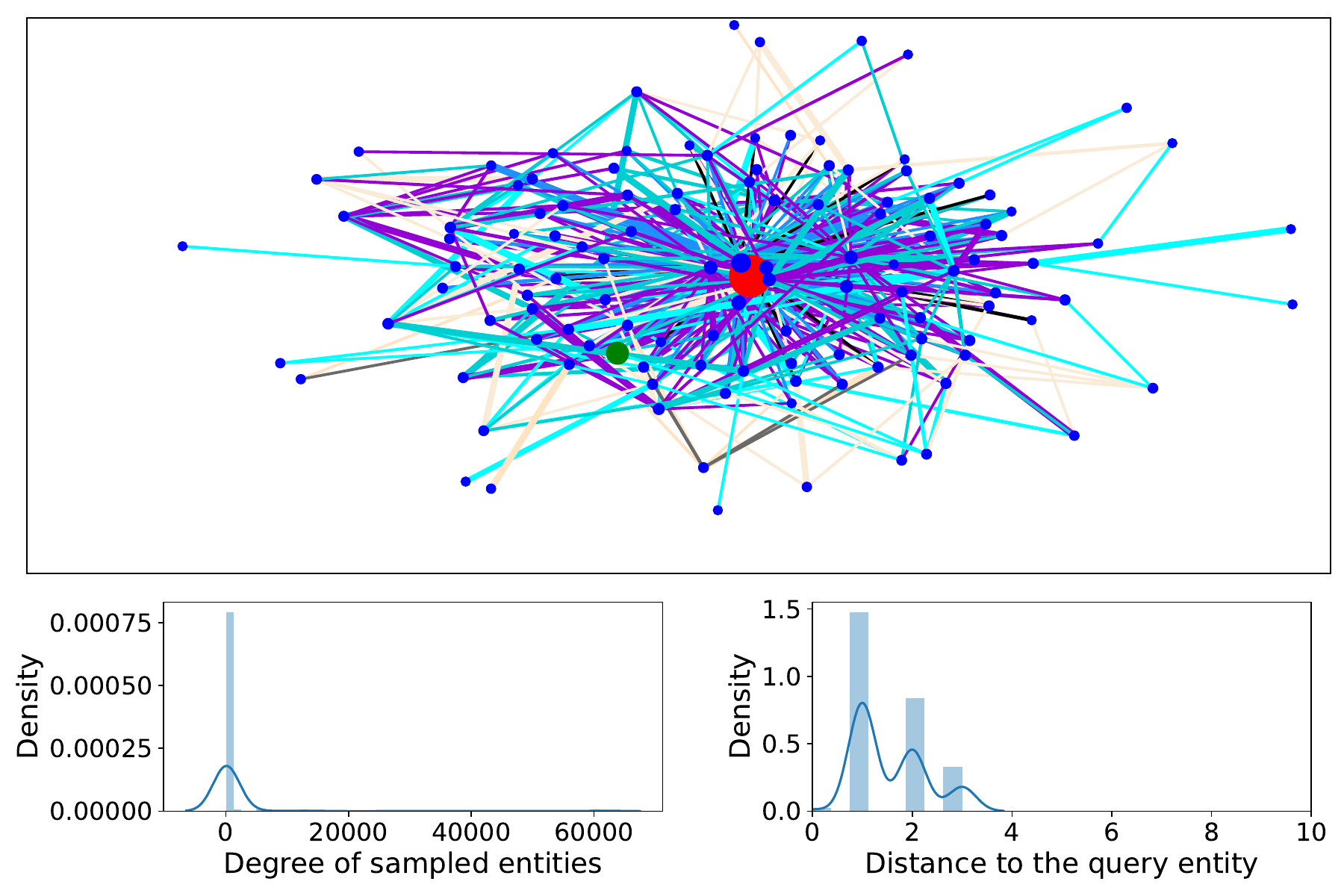}
		\hfill
		\includegraphics[width=6.8cm]{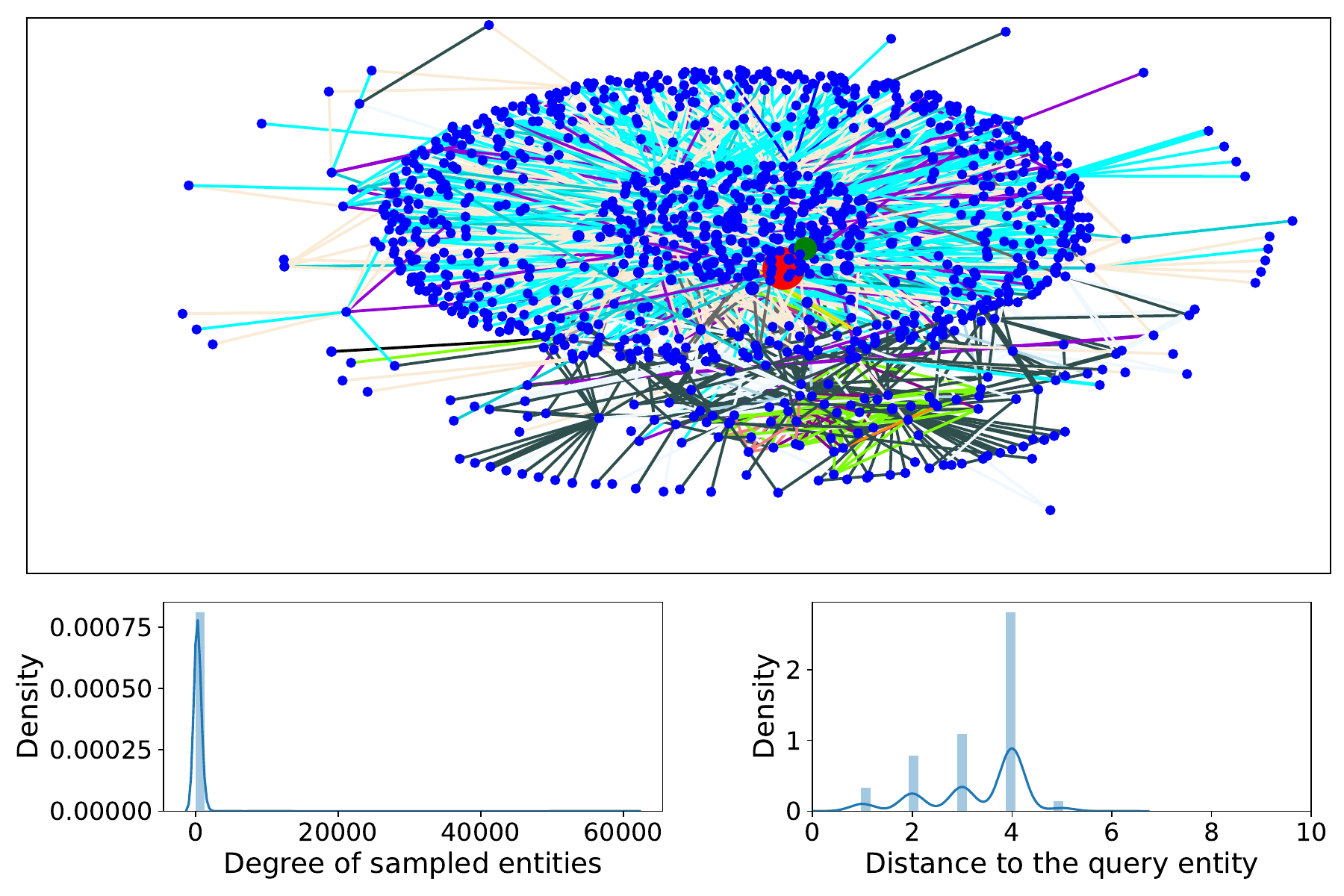}
		\hfill
		\vspace{-8px}
		\caption{
			Subgraphs (0.1\% and 1\%) from YAGO3-10:
			$u \! = \! 17, q \! = \! 39, v \! = \! \{ 54968 \}$. 
		}
		\vspace{-4px}
	\end{figure*}

	\begin{figure*}[ht]
		\centering
		\hfill
		\includegraphics[width=6.8cm]{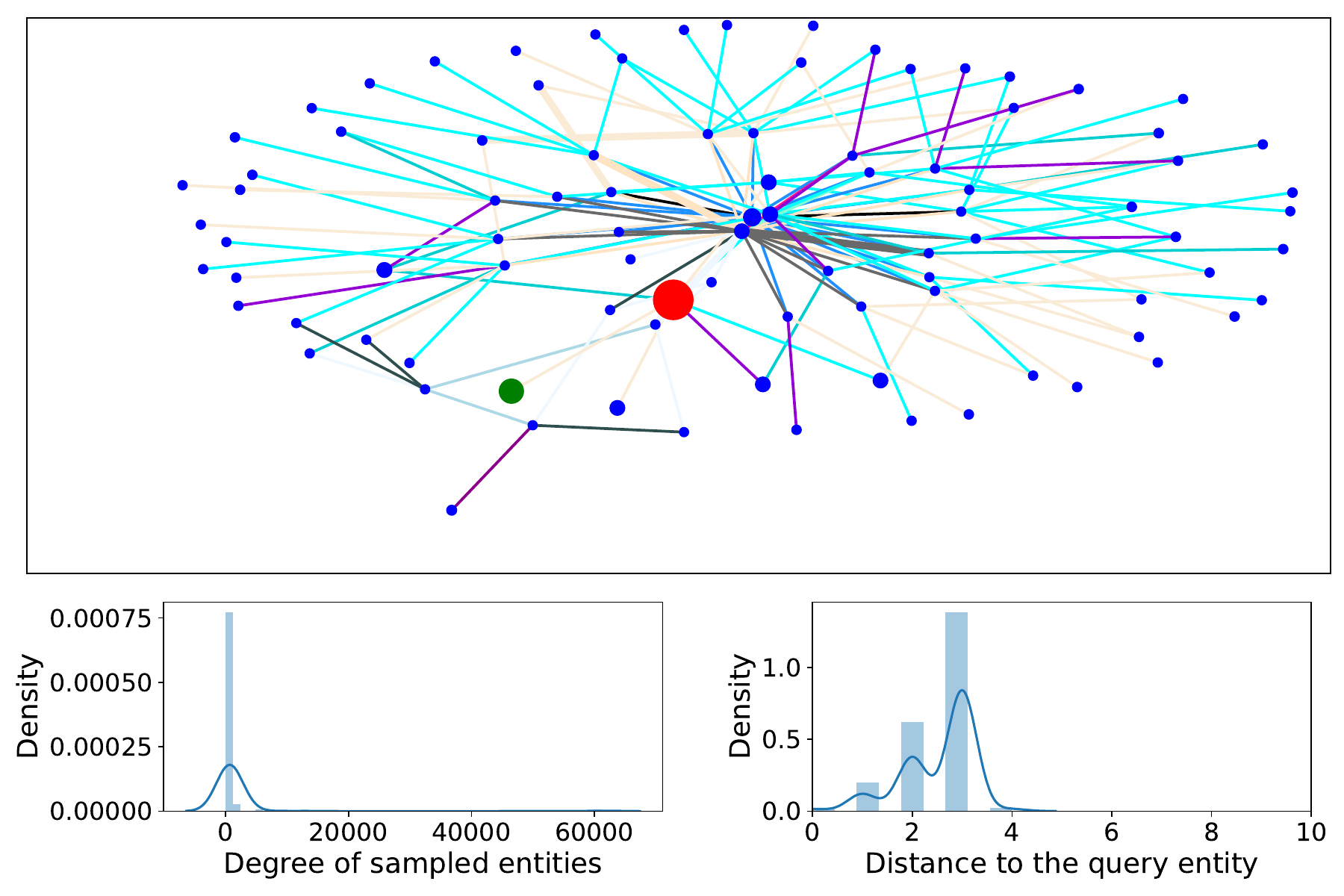}
		\hfill
		\includegraphics[width=6.8cm]{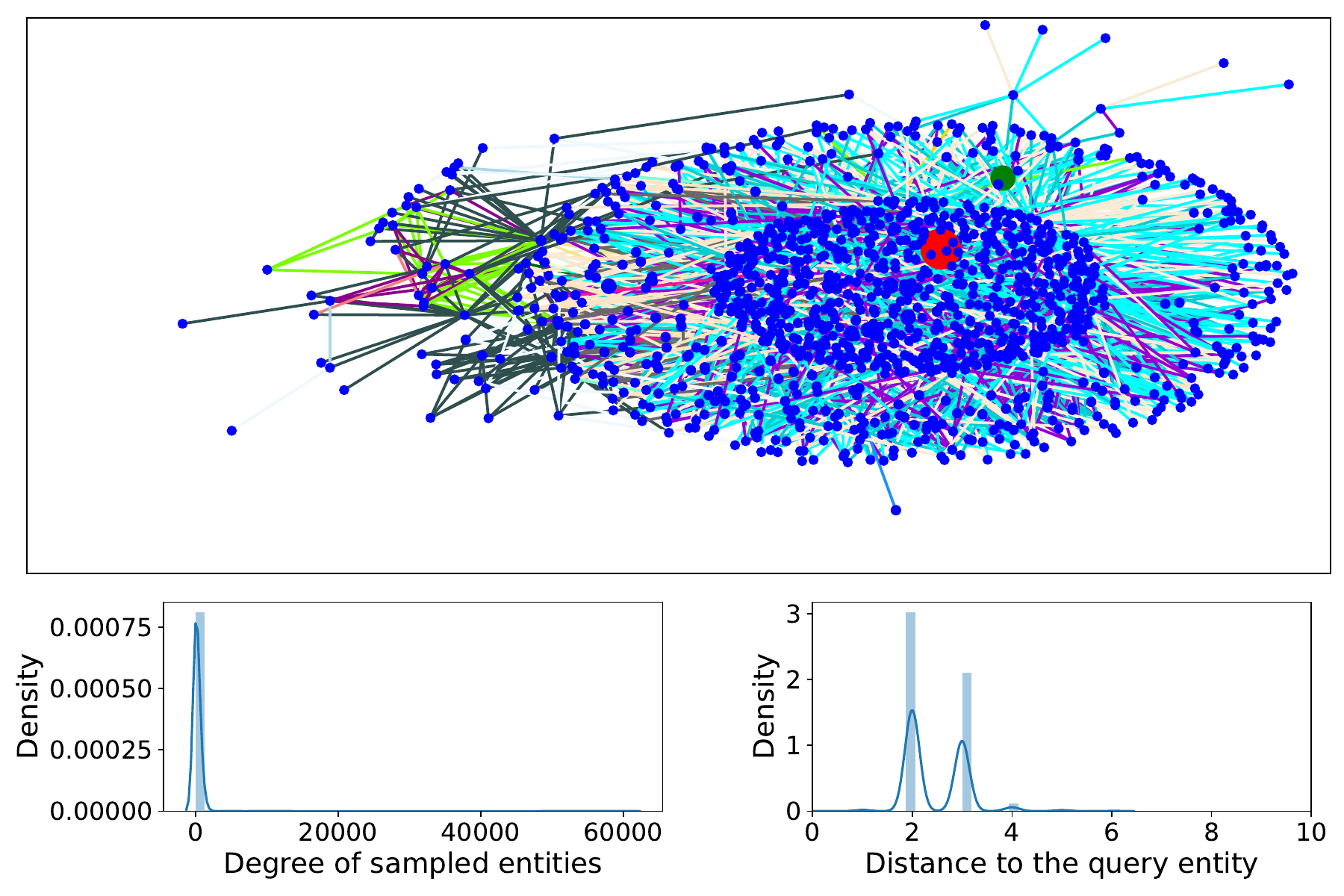}
		\hfill
		\vspace{-8px}
		\caption{
			Subgraphs (0.1\% and 1\%) from YAGO3-10:
			$u \! = \! 20, q \! = \! 2, v \! = \! \{ 34580 \}$. 
		}
		\vspace{-4px}
	\end{figure*}
	
	\begin{figure*}[ht]
		\centering
		\hfill
		\includegraphics[width=6.8cm]{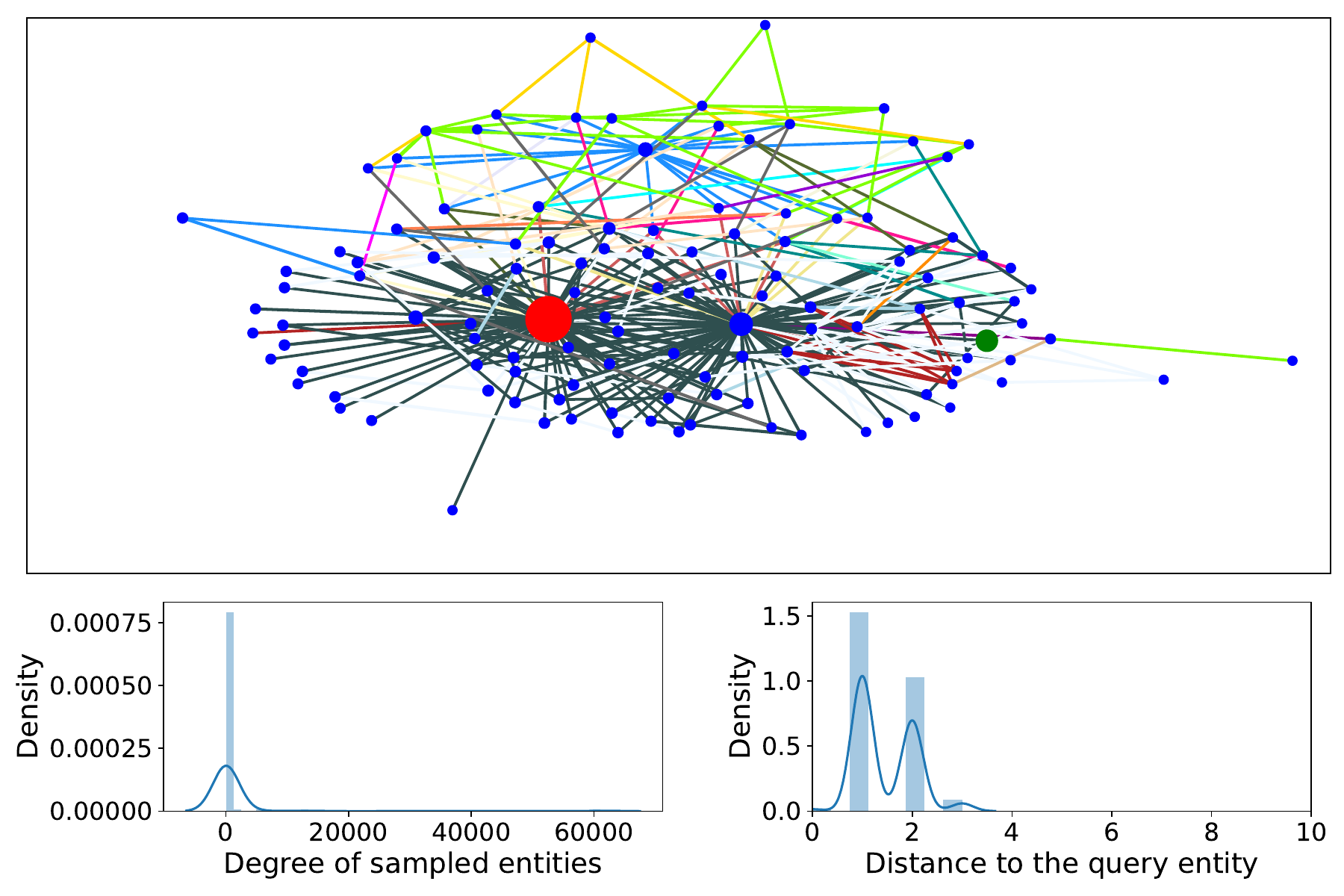}
		\hfill
		\includegraphics[width=6.8cm]{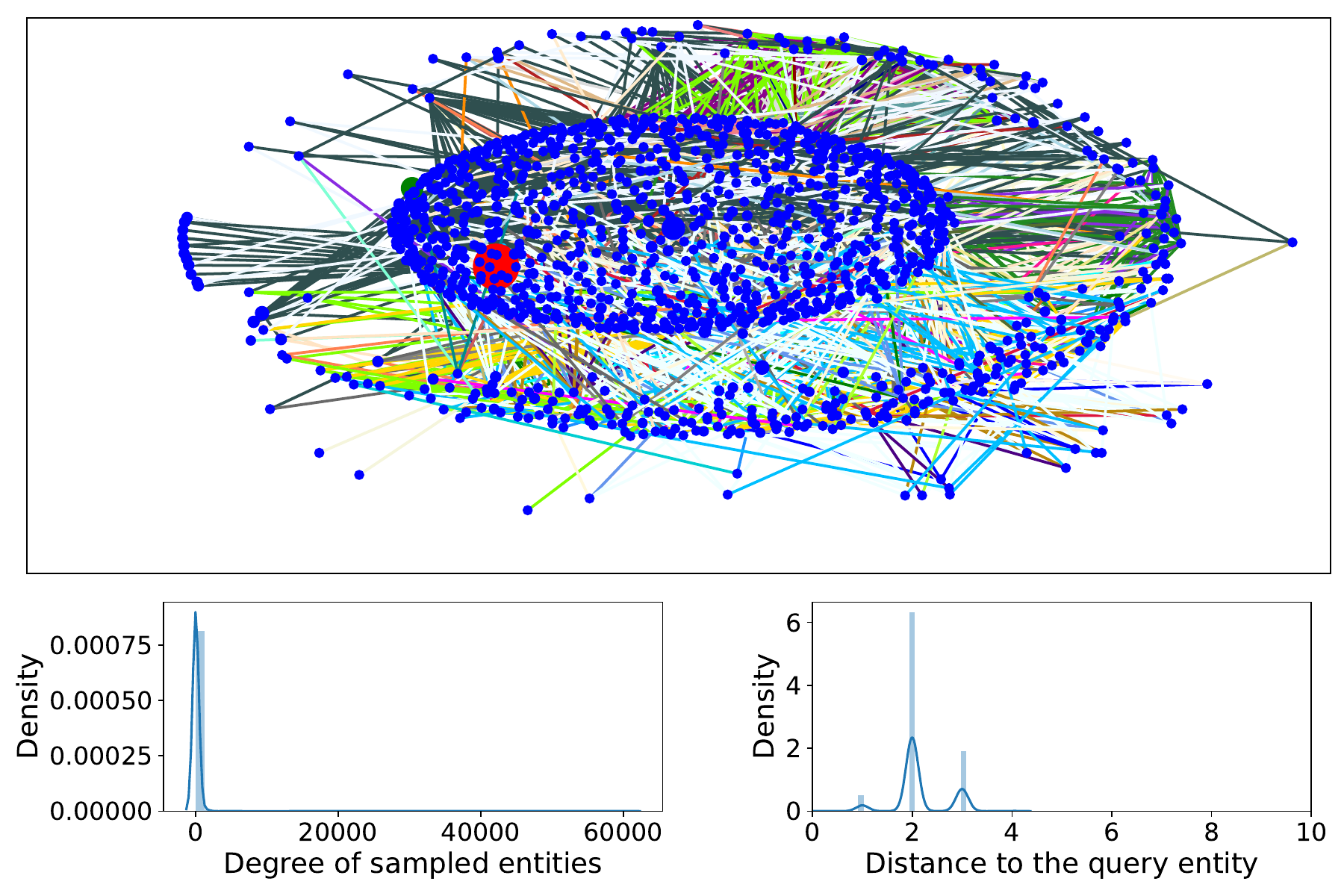}
		\hfill
		\vspace{-8px}
		\caption{
			Subgraphs (0.1\% and 1\%) from YAGO3-10:
			$u \! = \! 25, q \! = \! 37, v \! = \! \{ 40490 \}$. 
		}
		\vspace{-4px}
	\end{figure*}

	\begin{figure*}[ht]
		\centering
		\hfill
		\includegraphics[width=6.8cm]{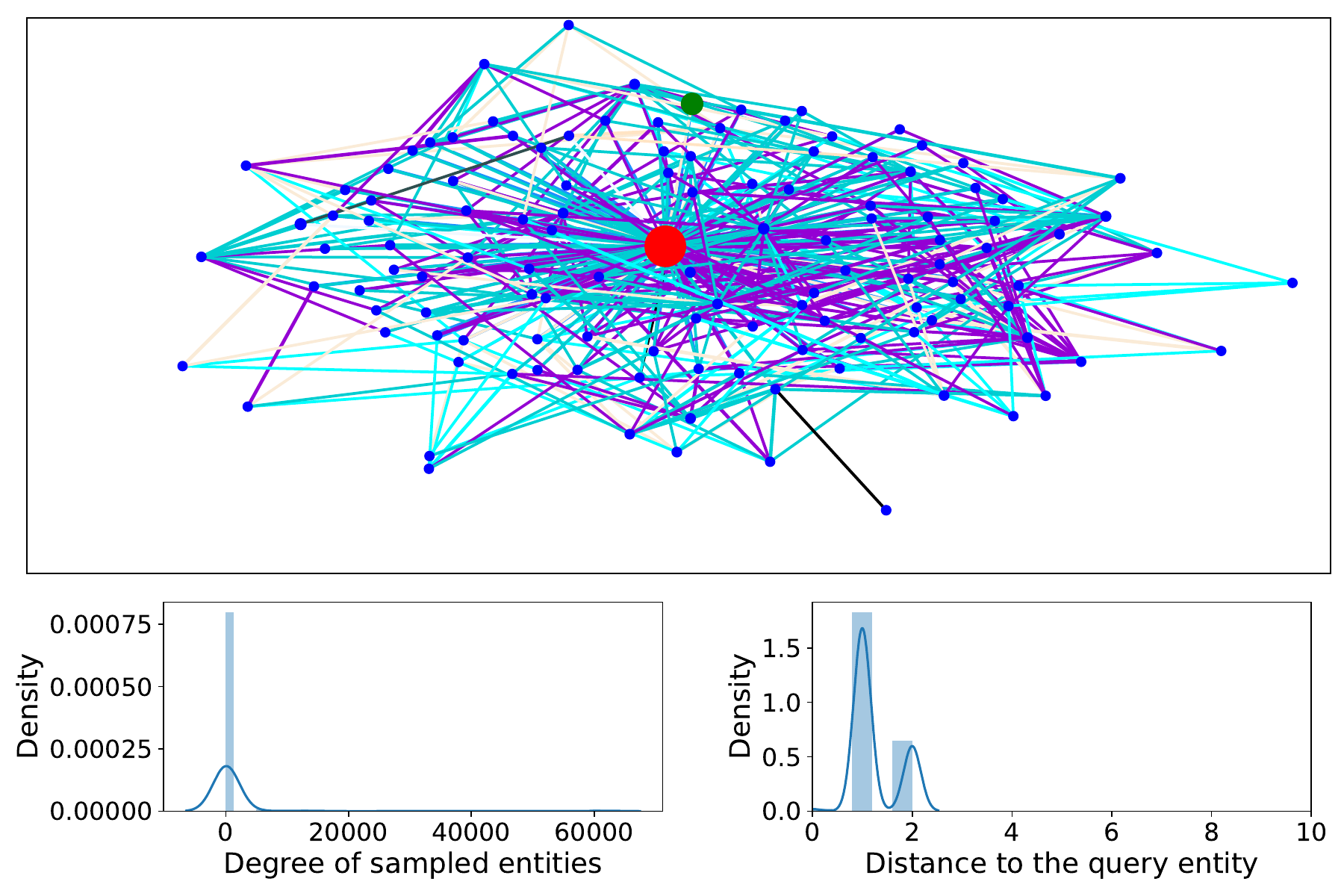}
		\hfill
		\includegraphics[width=6.8cm]{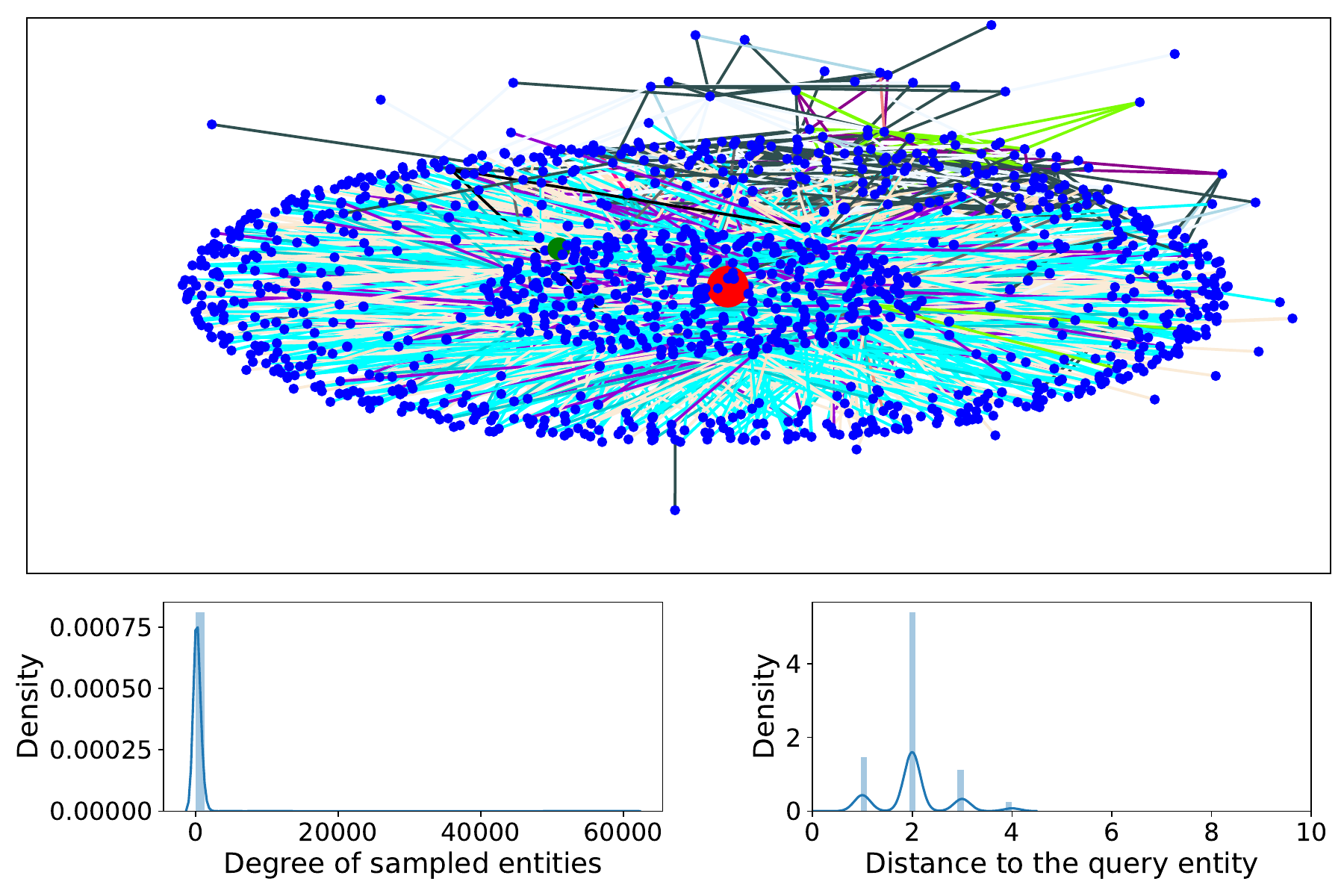}
		\hfill
		\vspace{-8px}
		\caption{
			Subgraphs (0.1\% and 1\%) from YAGO3-10:
			$u \! = \! 27, q \! = \! 39, v \! = \! \{ 3801 \}$.  
		}
		\vspace{-4px}
	\end{figure*}
	
	\begin{figure*}[ht]
		\centering
		\hfill
		\includegraphics[width=6.8cm]{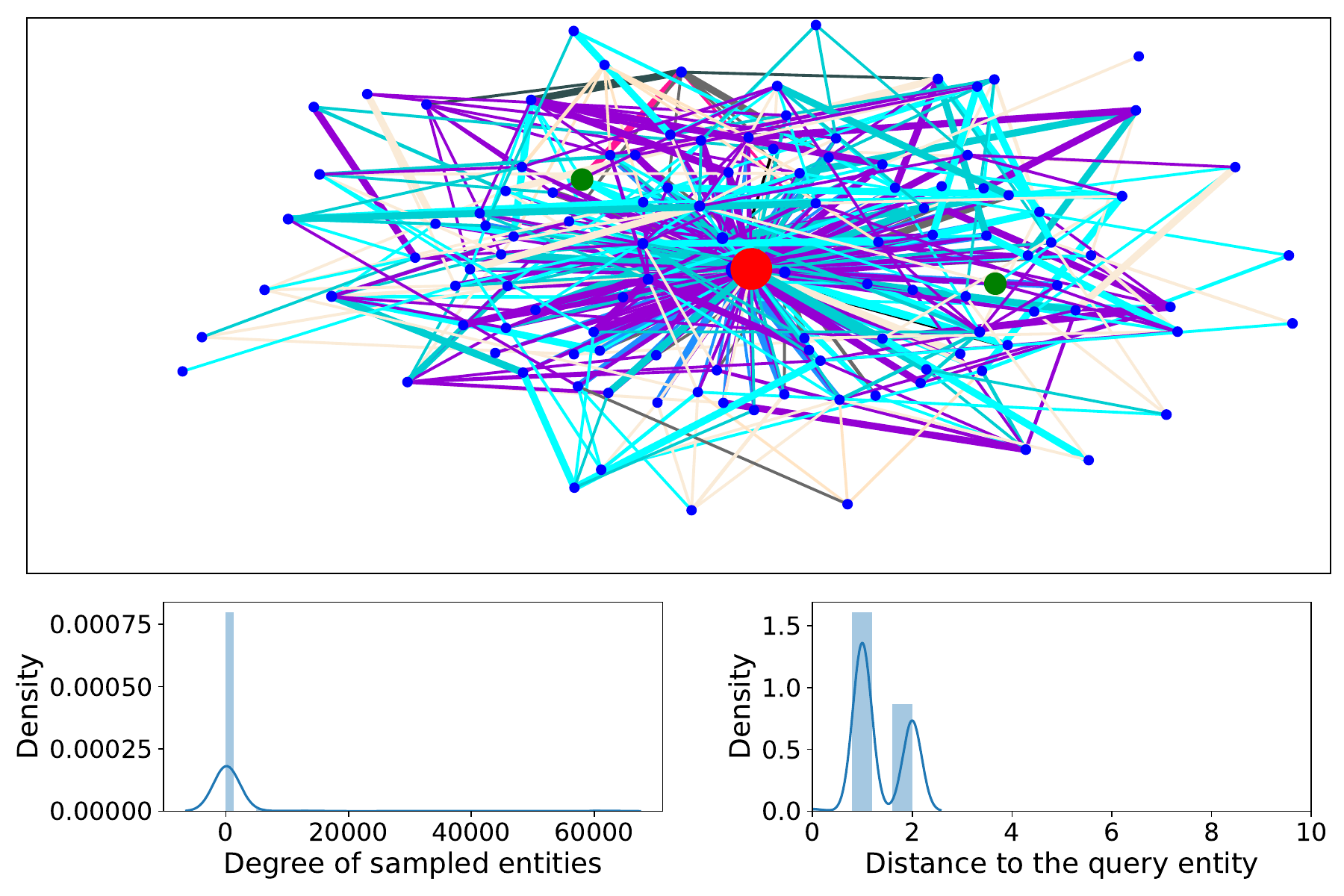}
		\hfill
		\includegraphics[width=6.8cm]{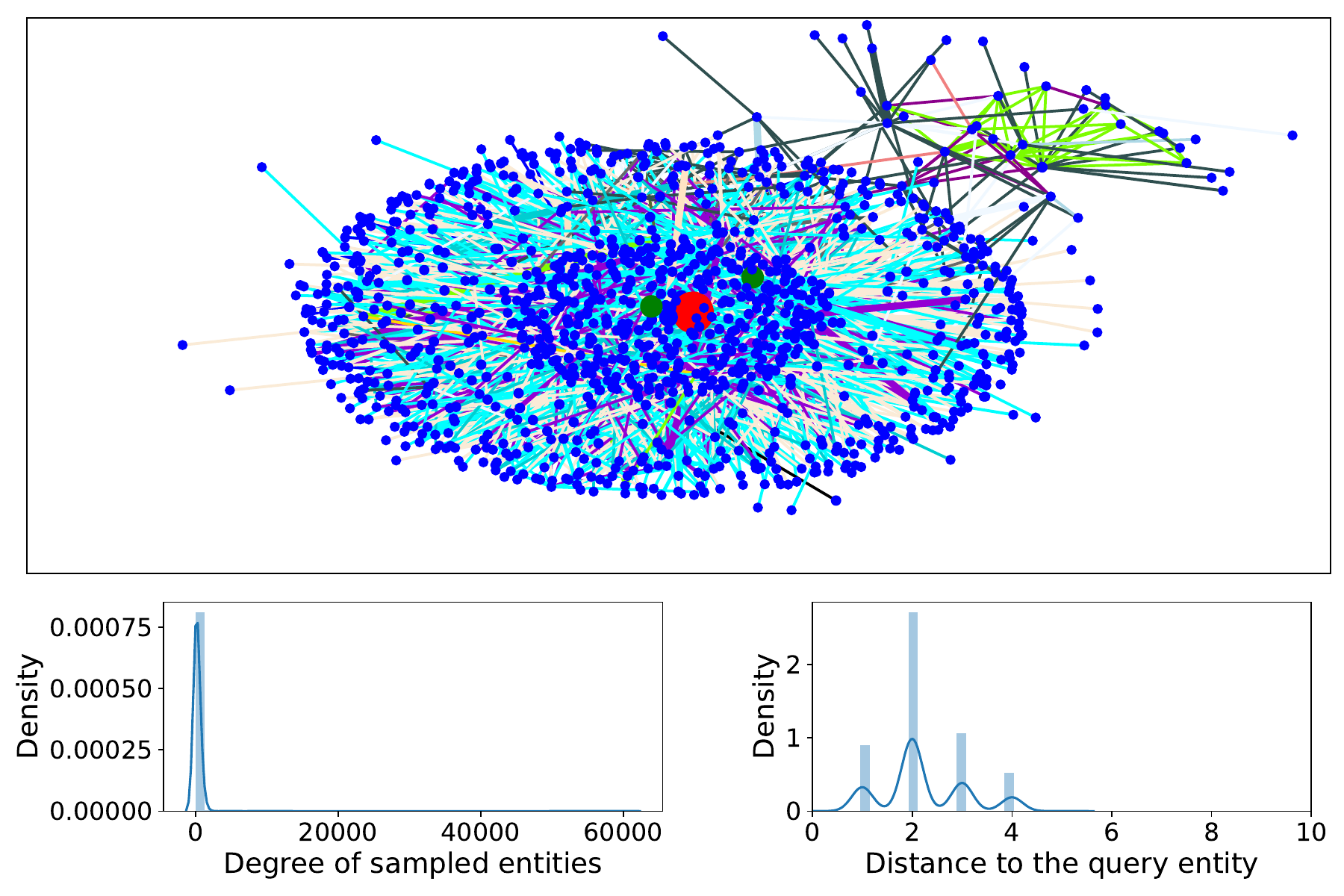}
		\hfill
		\vspace{-8px}
		\caption{
			Subgraphs (0.1\% and 1\%) from YAGO3-10:
			$u \! = \! 29, q \! = \! 38, v \! = \! \{ 33723, 82573 \}$. 
		}
		\vspace{-4px}
	\end{figure*}
	
	\begin{figure*}[ht]
		\centering
		\hfill
		\includegraphics[width=6.8cm]{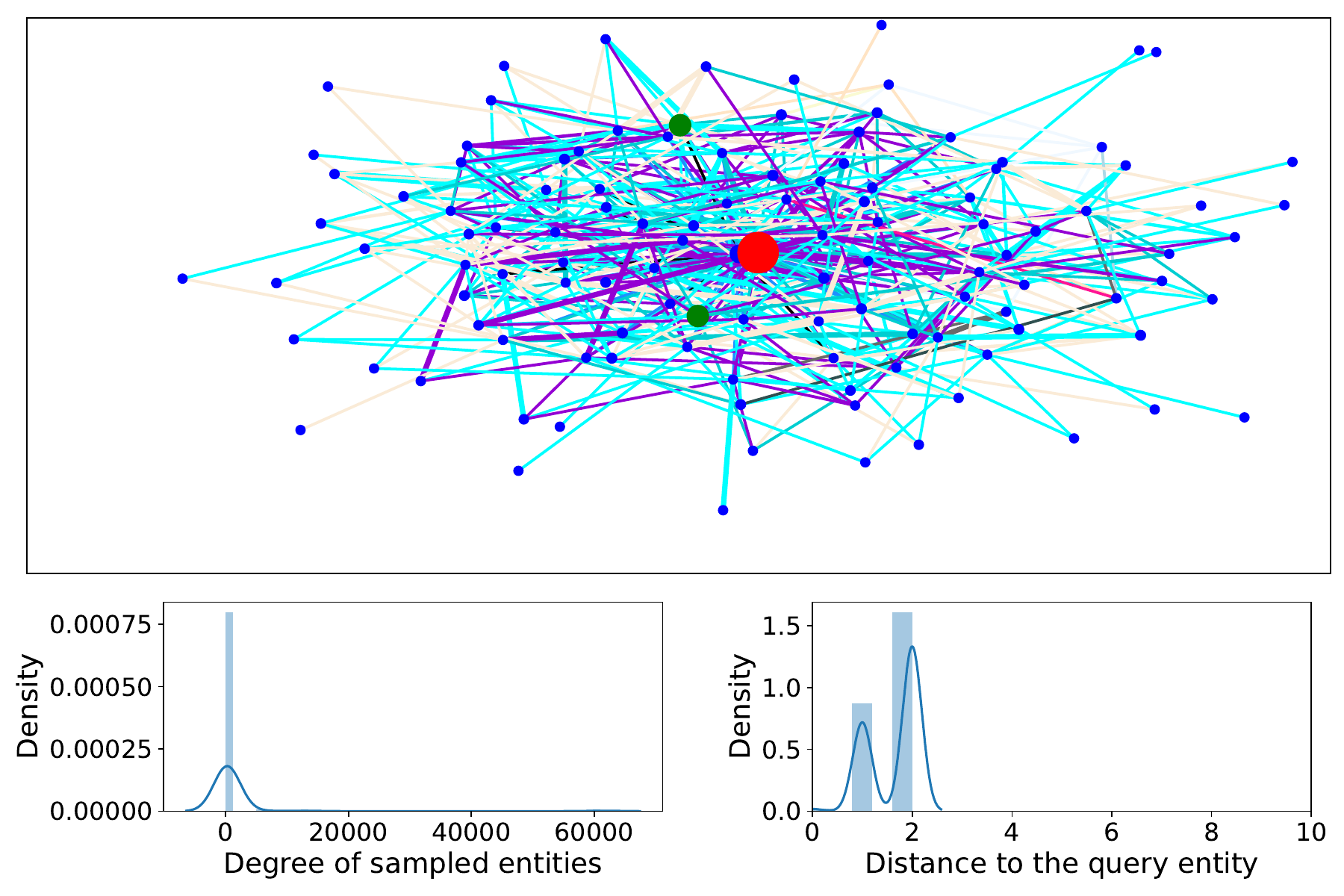}
		\hfill
		\includegraphics[width=6.8cm]{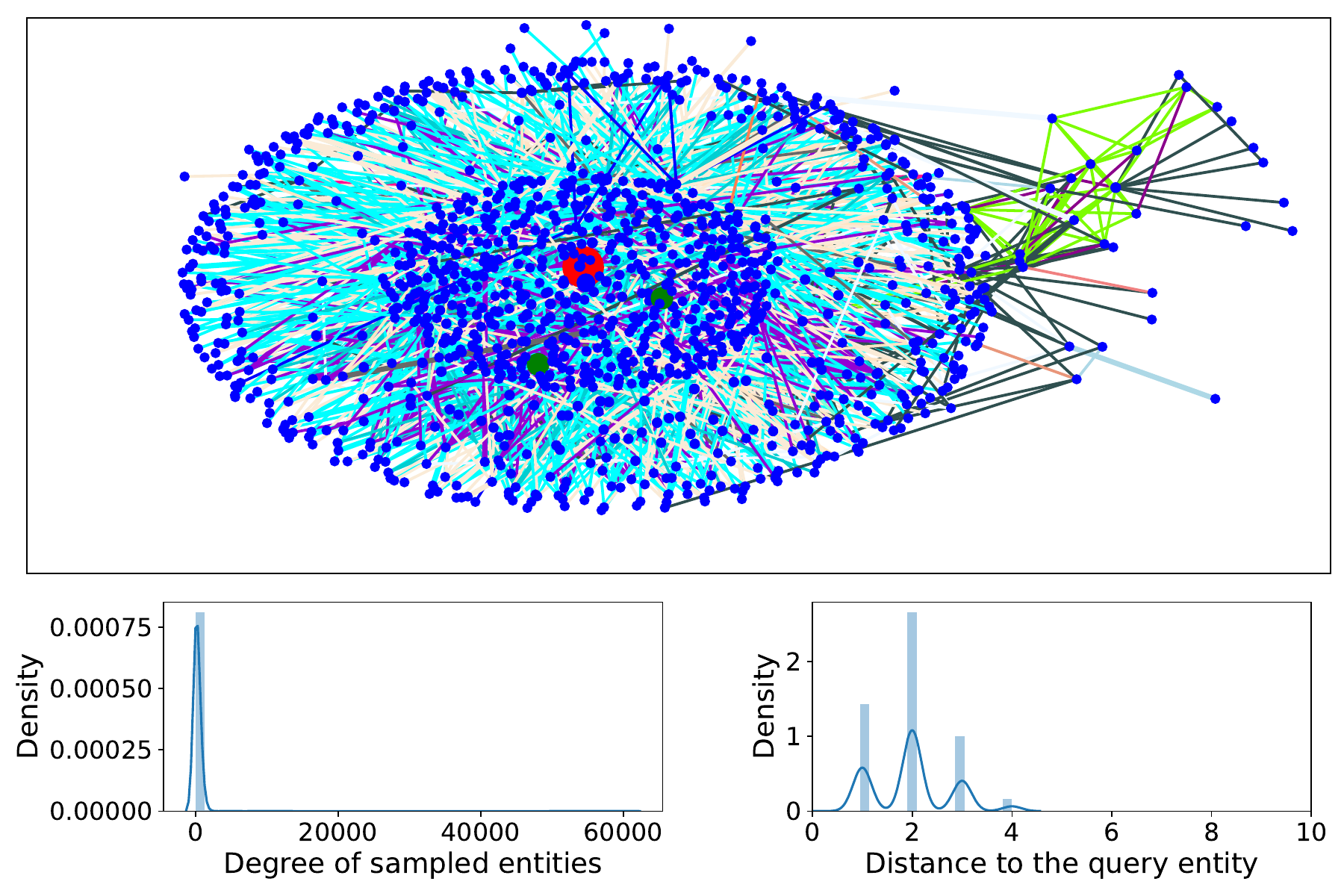}
		\hfill
		\vspace{-8px}
		\caption{
			Subgraphs (0.1\% and 1\%) from YAGO3-10:
			$u \! = \! 55, q \! = \! 38, v \! = \! \{ 14834, 67740 \}$.
		}
		\vspace{-4px}
	\end{figure*}
	
	\begin{figure*}[ht]
		\centering
		\hfill
		\includegraphics[width=6.8cm]{figures/subgraph_data/YAGO_queryidx_39_ratio_0.001}
		\hfill
		\includegraphics[width=6.8cm]{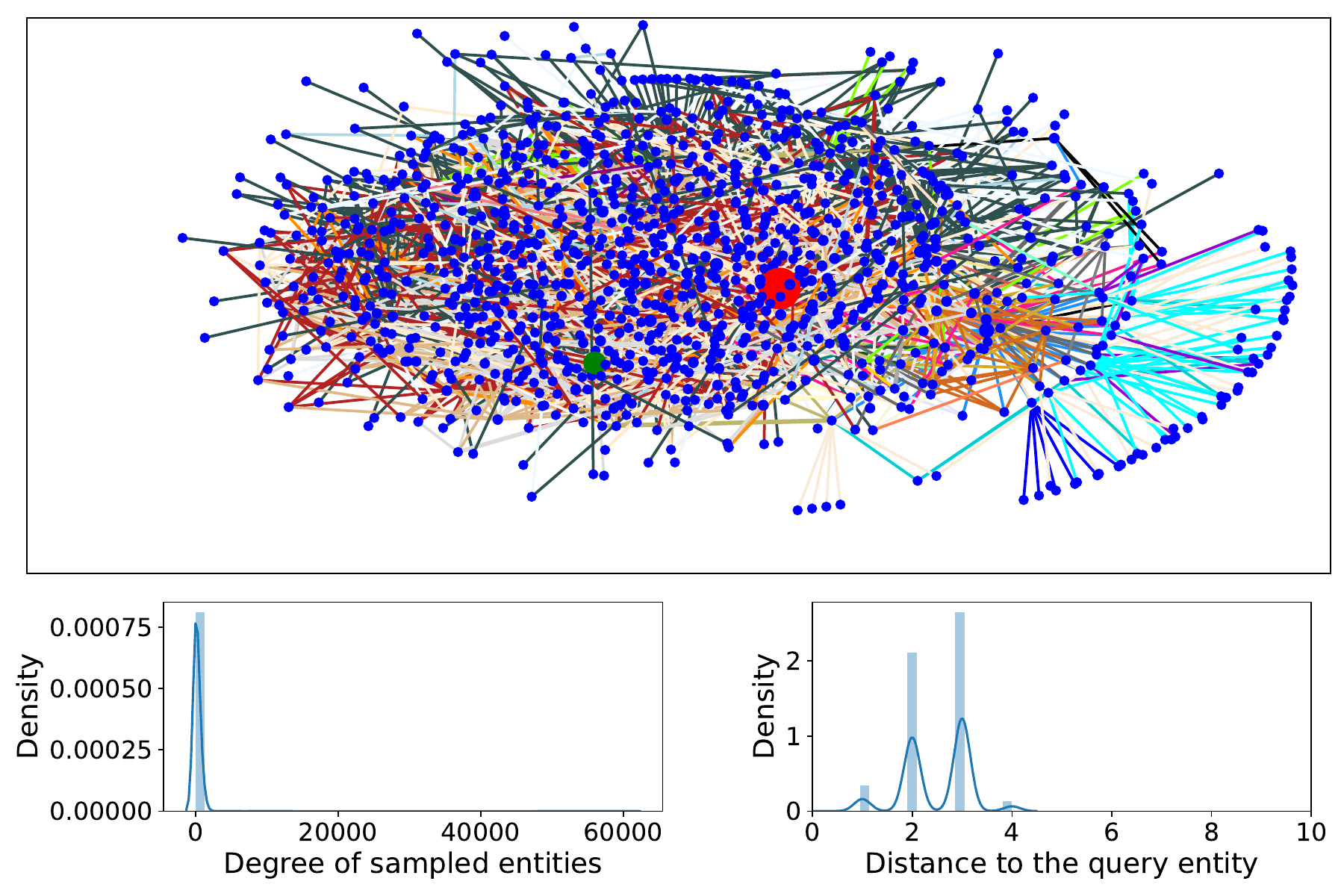}
		\hfill
		\vspace{-8px}
		\caption{
			Subgraphs (0.1\% and 1\%) from YAGO3-10:
			$u \! = \! 102, q \! = \! 12, v \! = \! \{ 15823 \}$. 
		}
		\vspace{-4px}
	\end{figure*}
	
	\begin{figure*}[ht]
		\centering
		\hfill
		\includegraphics[width=6.8cm]{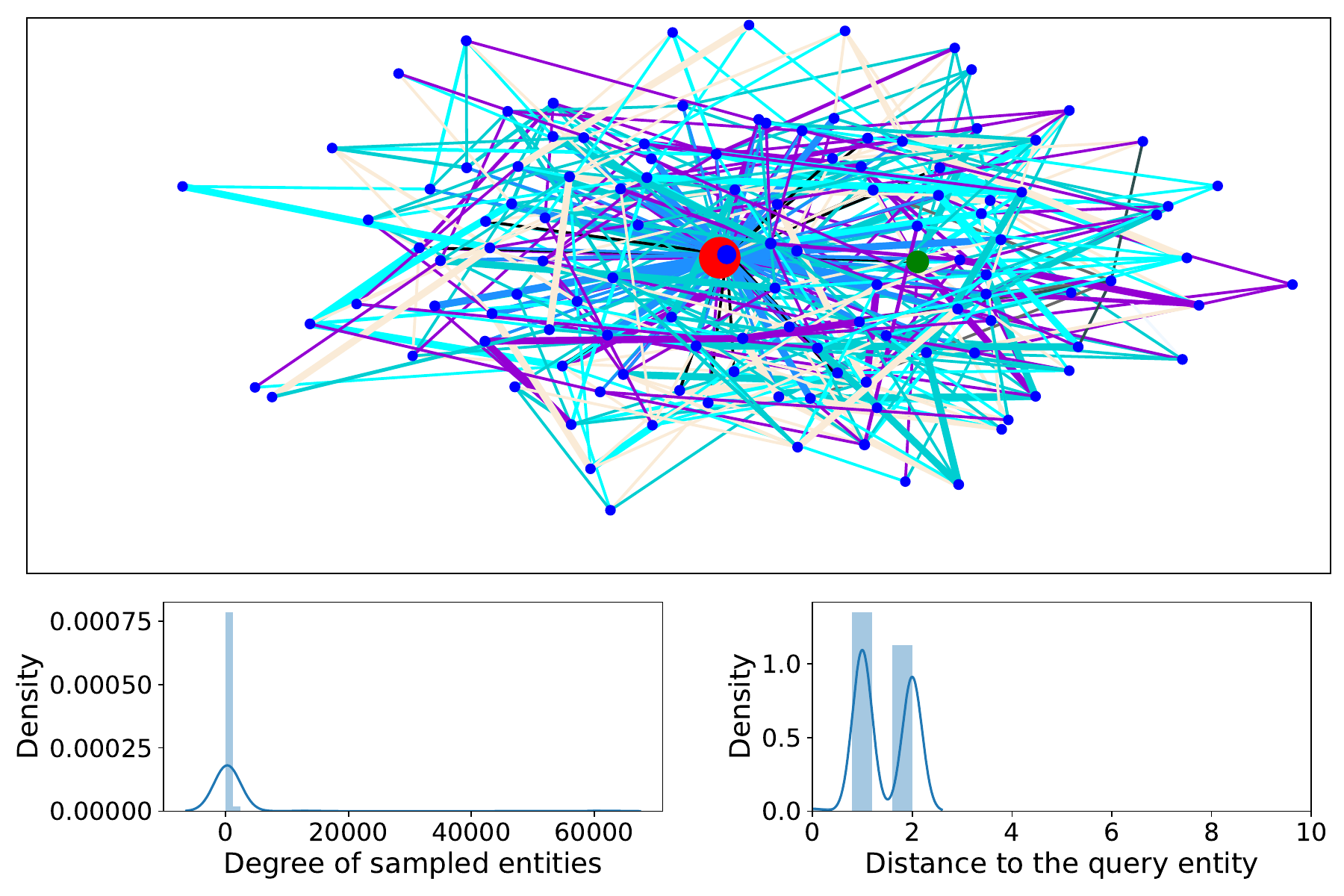}
		\hfill
		\includegraphics[width=6.8cm]{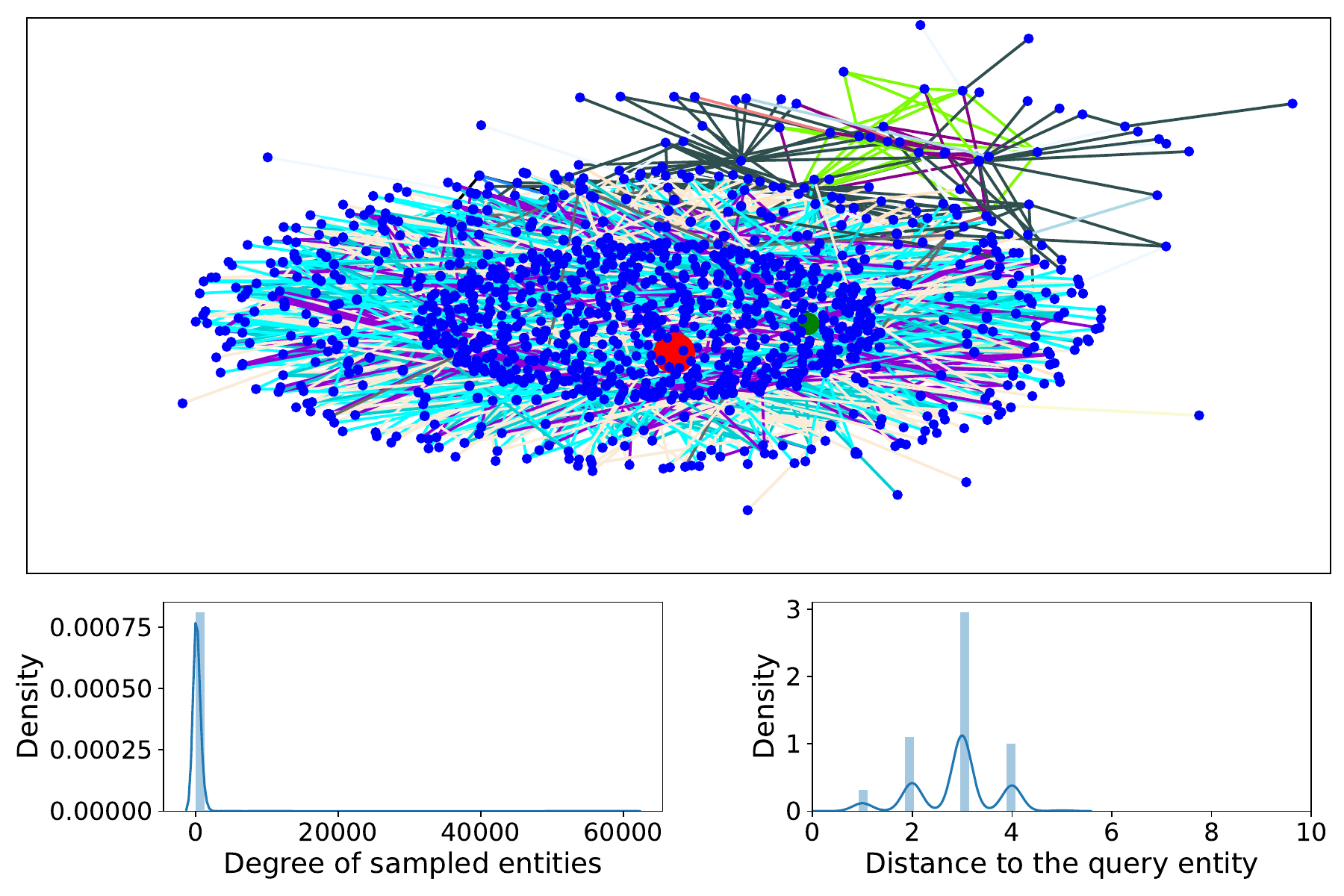}
		\hfill
		\vspace{-8px}
		\caption{
			Subgraphs (0.1\% and 1\%) from YAGO3-10:
			$u \! = \! 108, q \! = \! 39, v \! = \! \{ 7271 \}$. 
		}
		\vspace{-4px}
	\end{figure*}
	
	\begin{figure*}[ht]
		\centering
		\hfill
		\includegraphics[width=6.8cm]{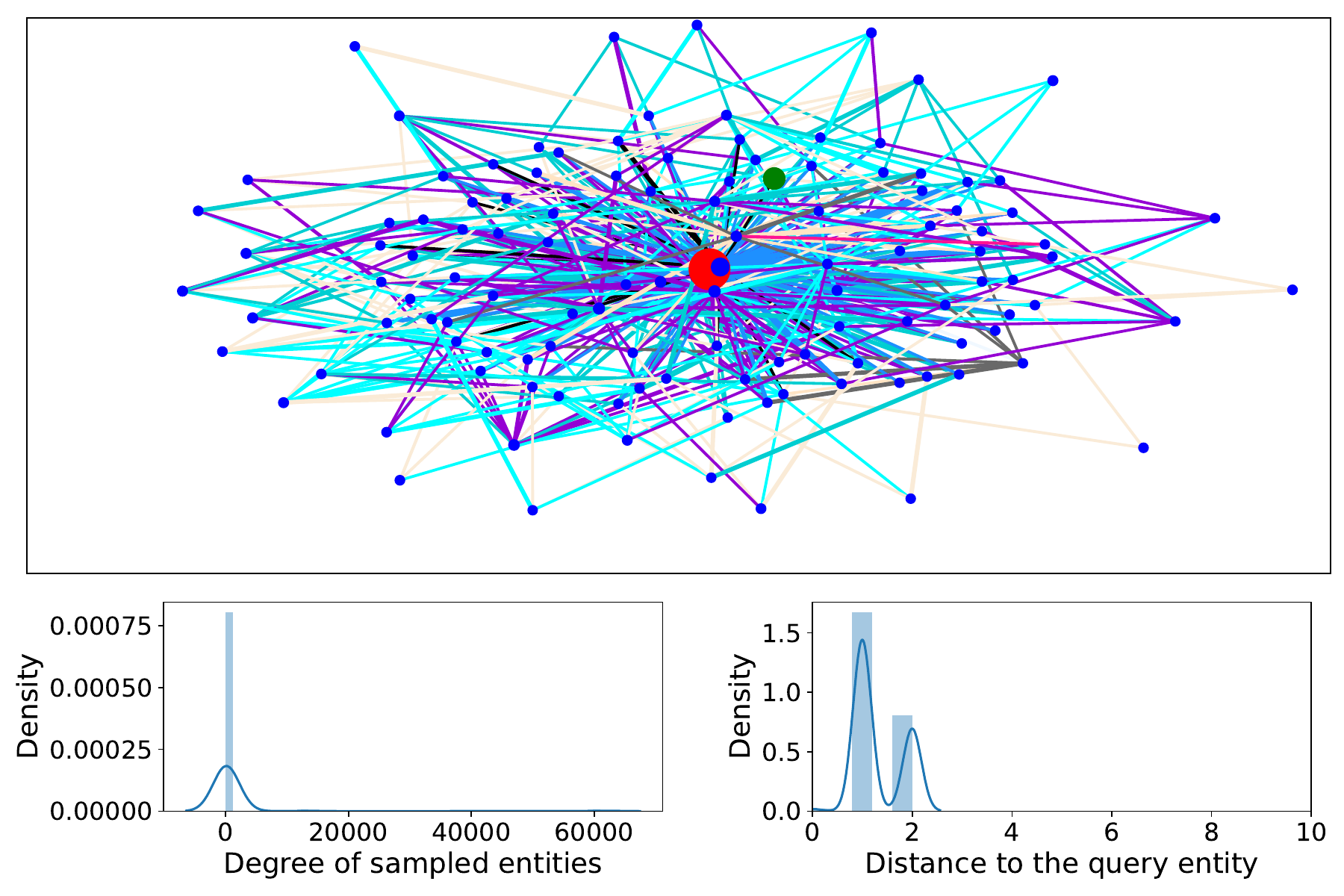}
		\hfill
		\includegraphics[width=6.8cm]{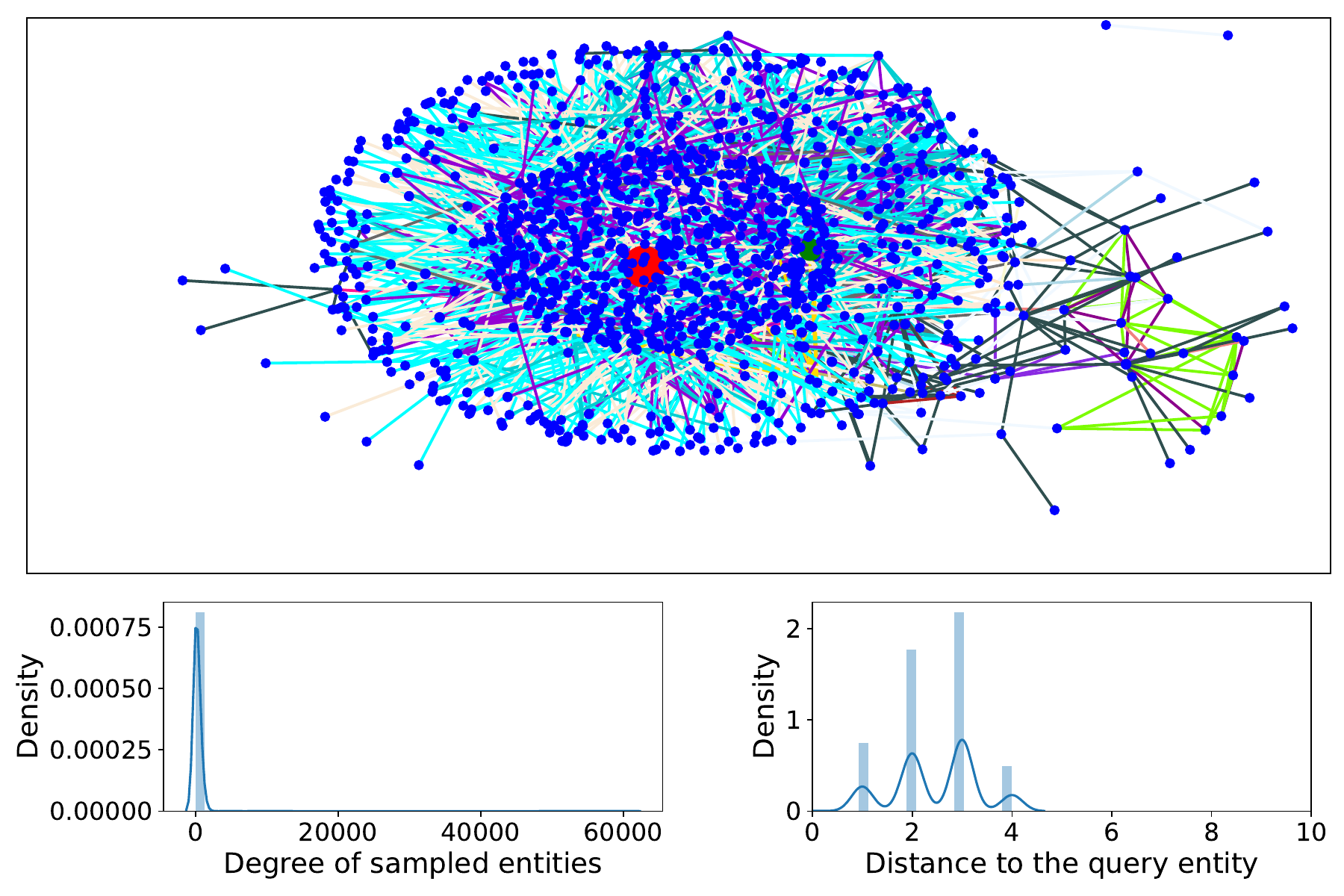}
		\hfill
		\vspace{-8px}
		\caption{
			Subgraphs (0.1\% and 1\%) from YAGO3-10:
			$u \! = \! 135, q \! = \! 38, v \! = \! \{ 51096 \}$. 
		}
		\vspace{-4px}
	\end{figure*}

	\begin{figure*}[ht]
		\centering
		\hfill
		\includegraphics[width=6.8cm]{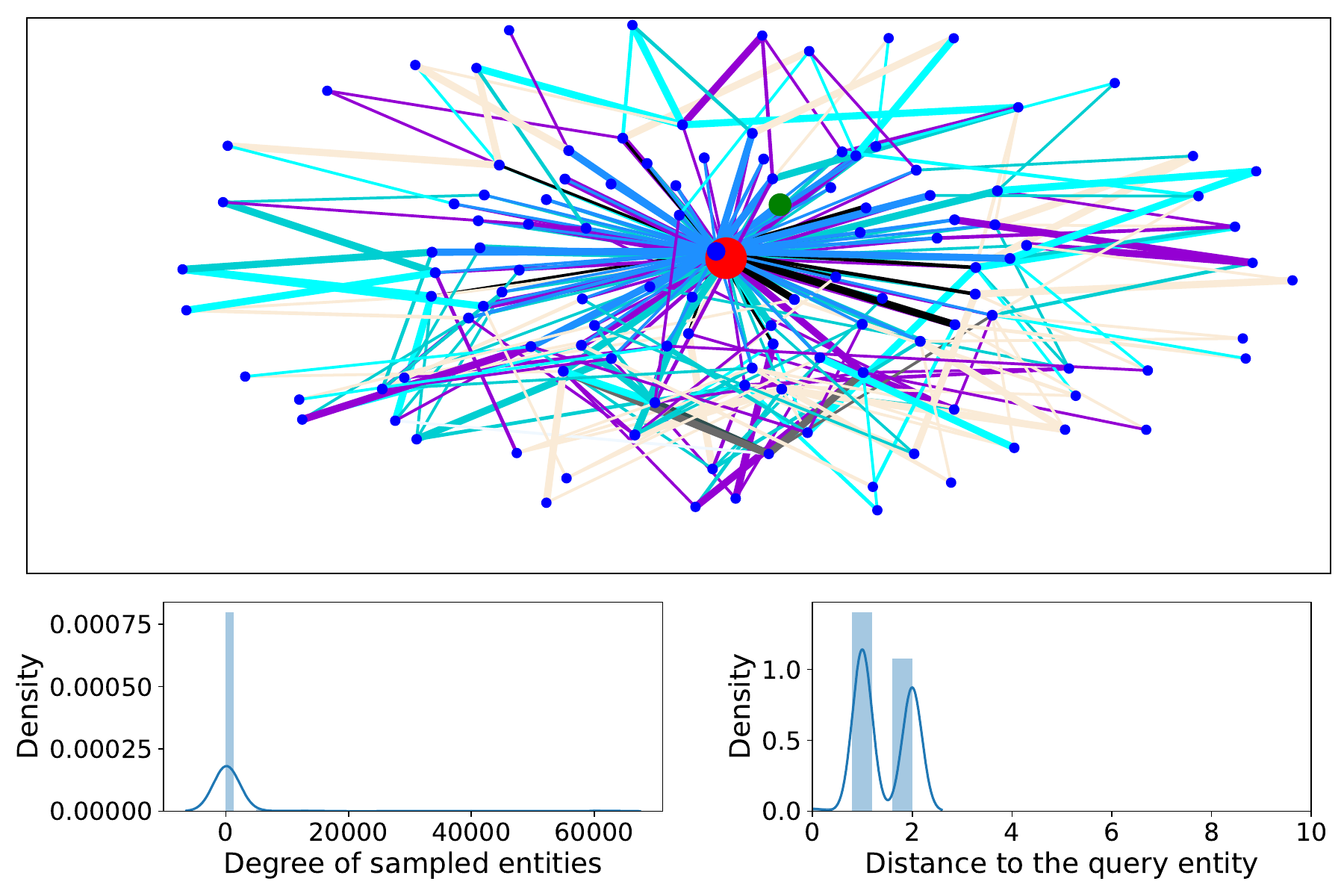}
		\hfill
		\includegraphics[width=6.8cm]{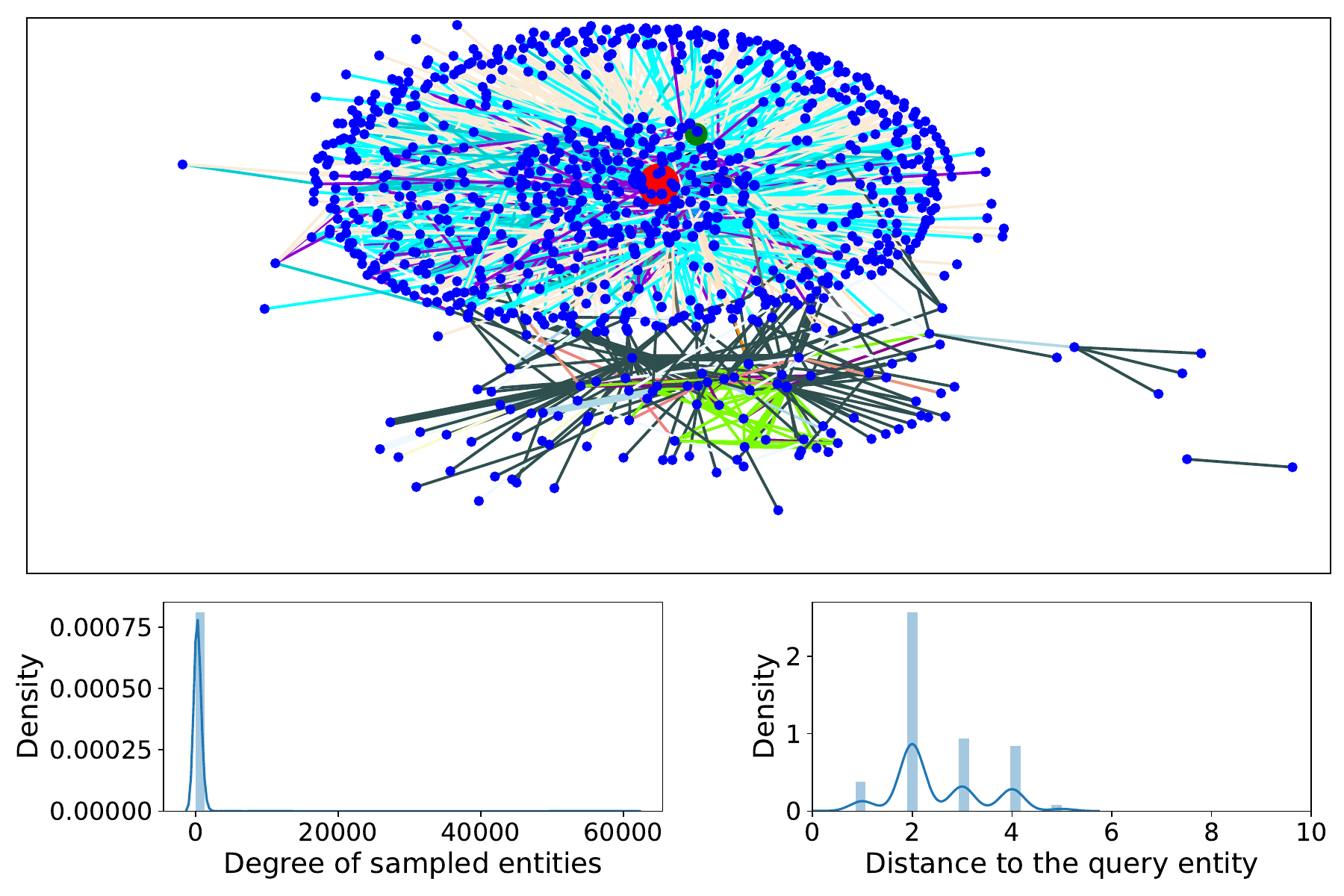}
		\hfill
		\vspace{-8px}
		\caption{
			Subgraphs (0.1\% and 1\%) from YAGO3-10:
			$u \! = \! 137, q \! = \! 39, v \! = \! \{ 26722 \}$. 
		}
		\vspace{-4px}
	\end{figure*}
	
	\begin{figure*}[ht]
		\centering
		\hfill
		\includegraphics[width=6.8cm]{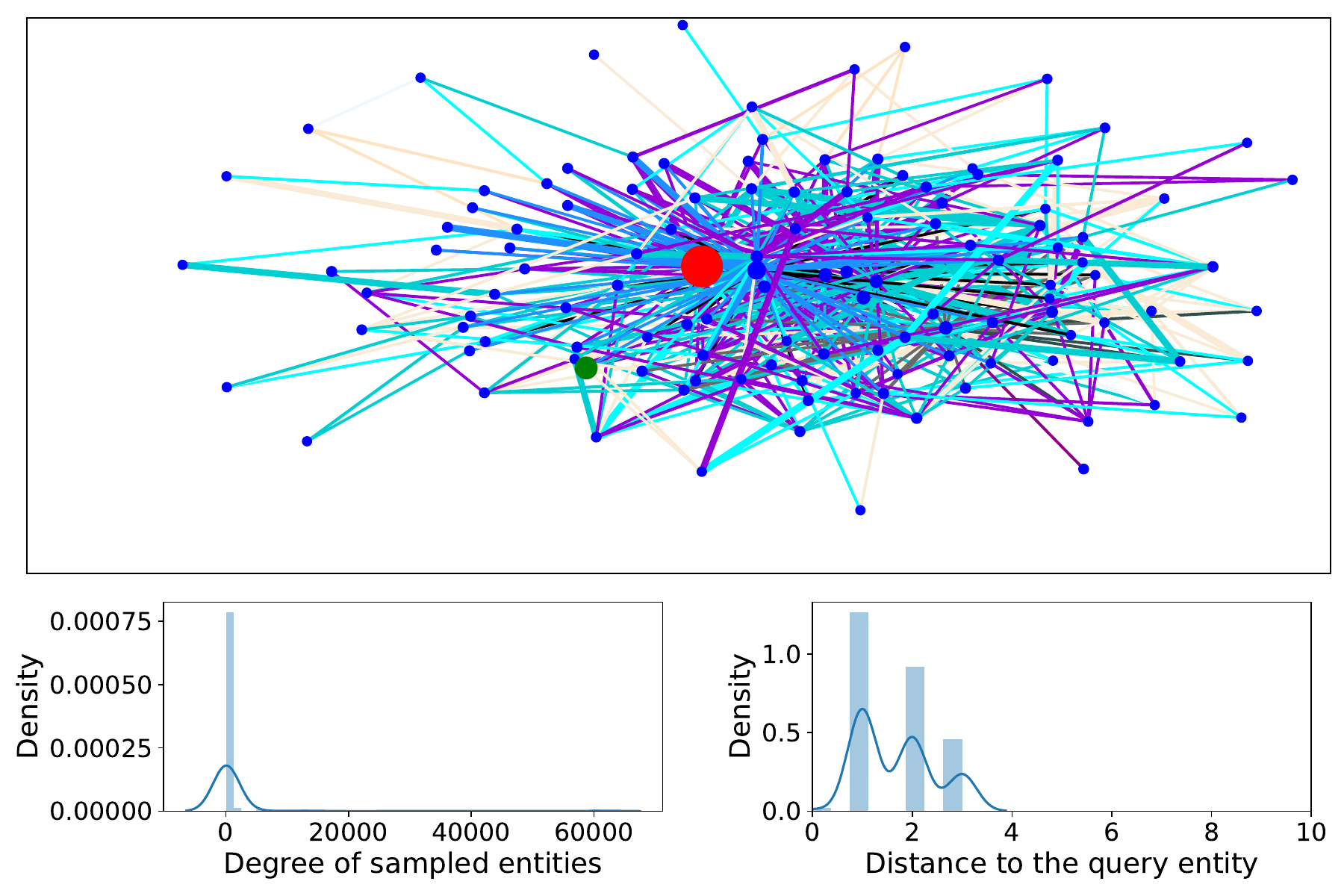}
		\hfill
		\includegraphics[width=6.8cm]{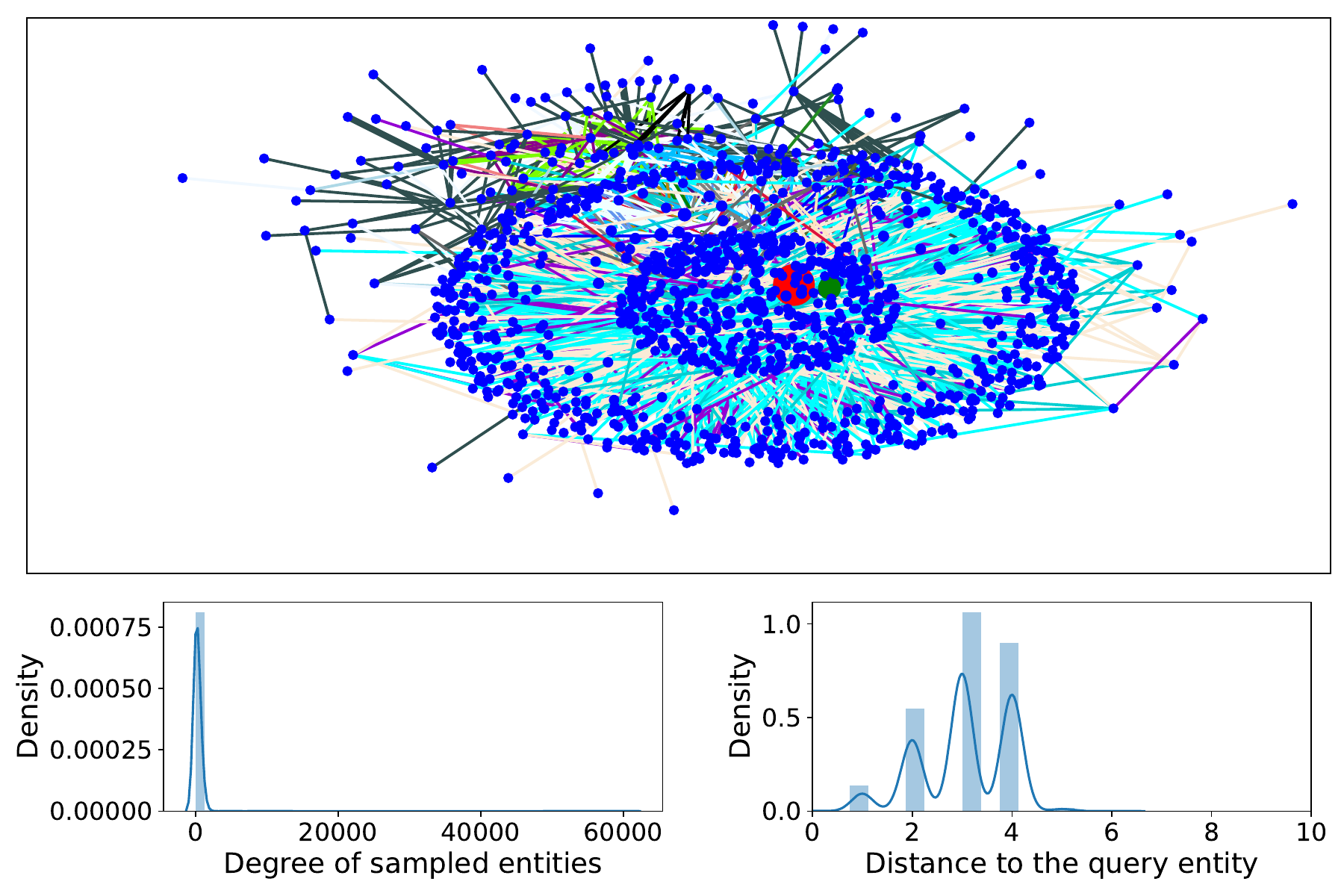}
		\hfill
		\vspace{-8px}
		\caption{
			Subgraphs (0.1\% and 1\%) from YAGO3-10:
			$u \! = \! 446, q \! = \! 38, v \! = \! \{ 104297 \}$.
		}
		\vspace{-4px}
	\end{figure*}
	
	\begin{figure*}[ht]
		\centering
		\hfill
		\includegraphics[width=6.8cm]{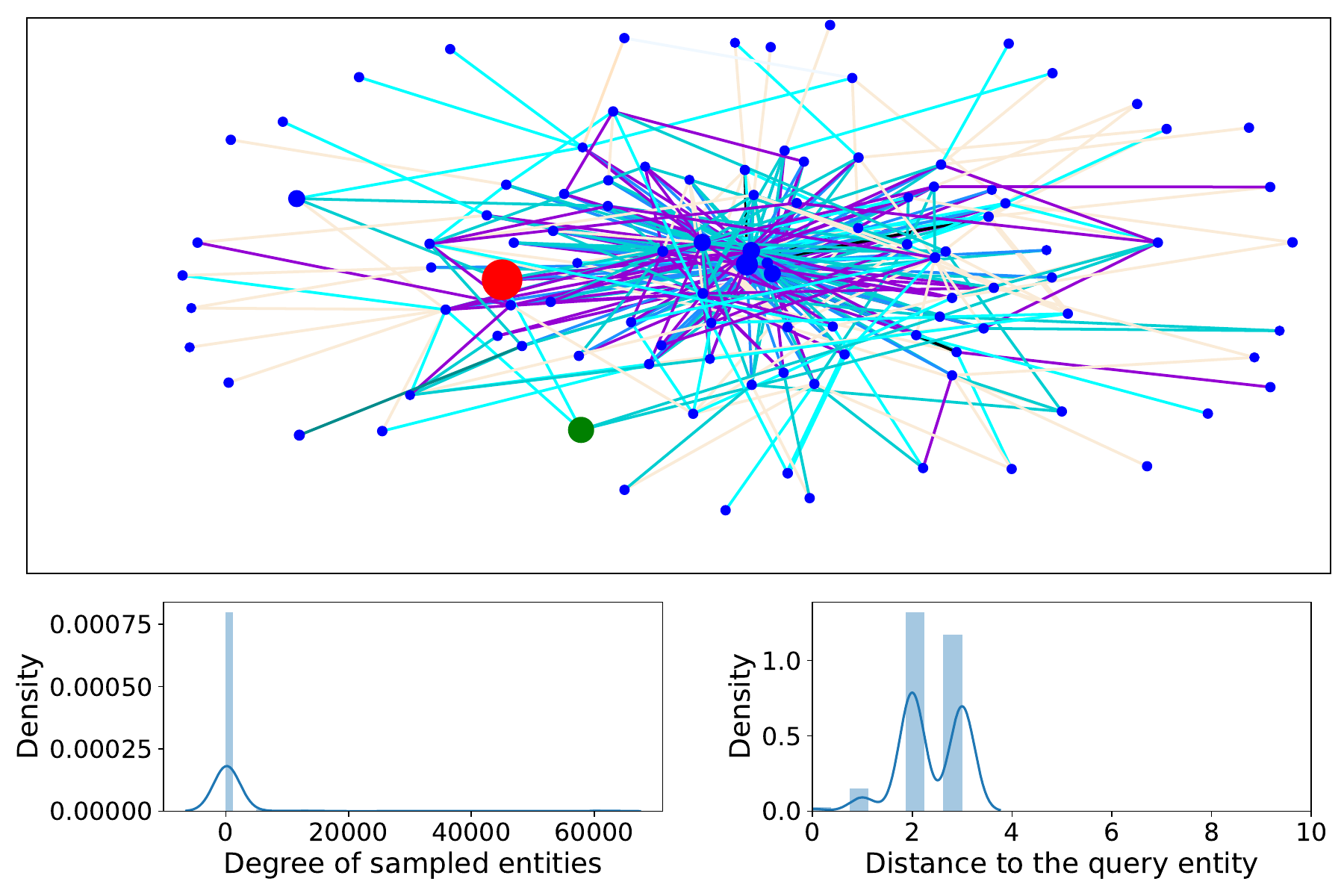}
		\hfill
		\includegraphics[width=6.8cm]{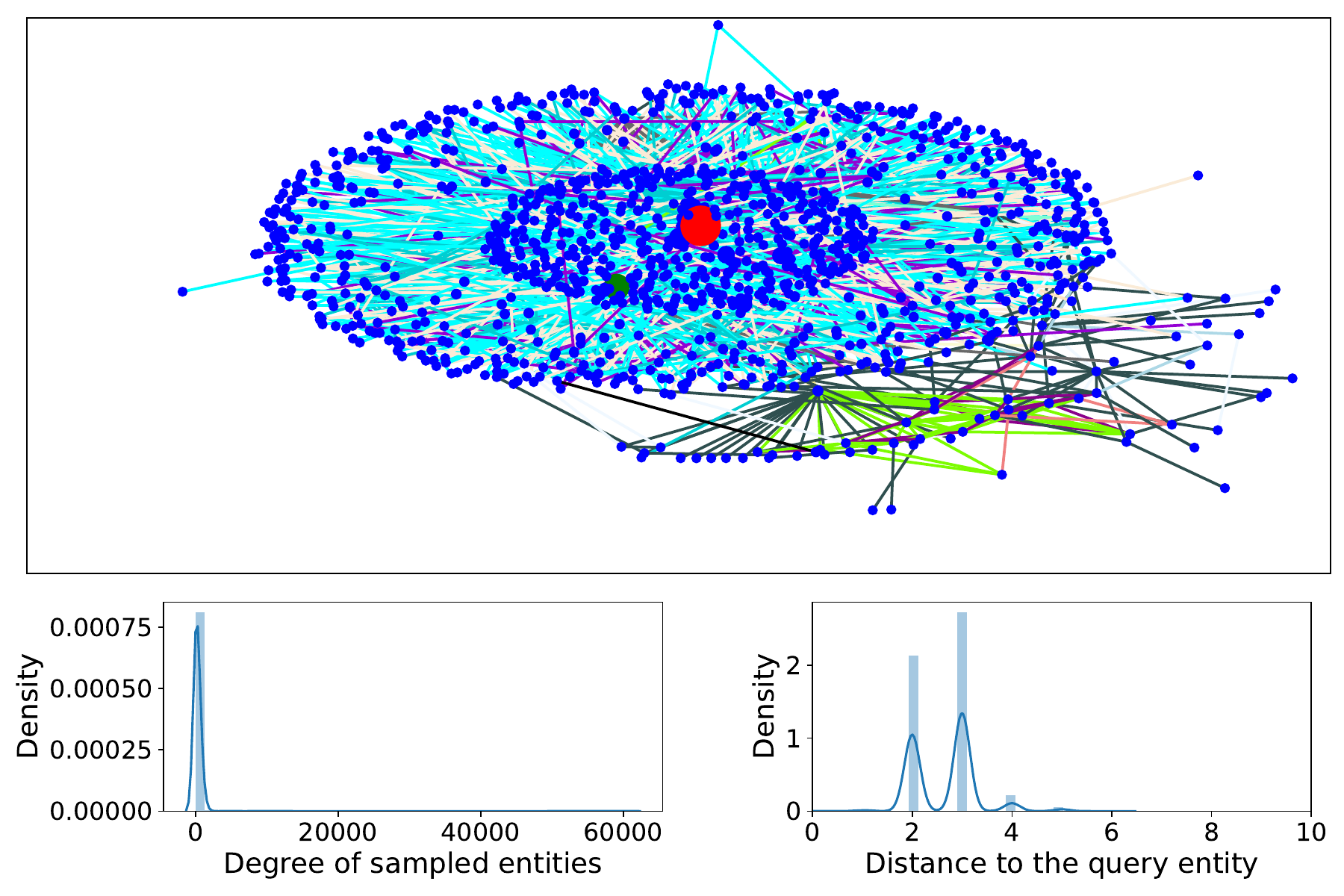}
		\hfill
		\vspace{-8px}
		\caption{
			Subgraphs (0.1\% and 1\%) from YAGO3-10:
			$u \! = \! 1072, q \! = \! 1, v \! = \! \{ 23394 \}$.
		}
		\vspace{-4px}
	\end{figure*}
	
	\begin{figure*}[ht]
		\centering
		\hfill
		\includegraphics[width=6.8cm]{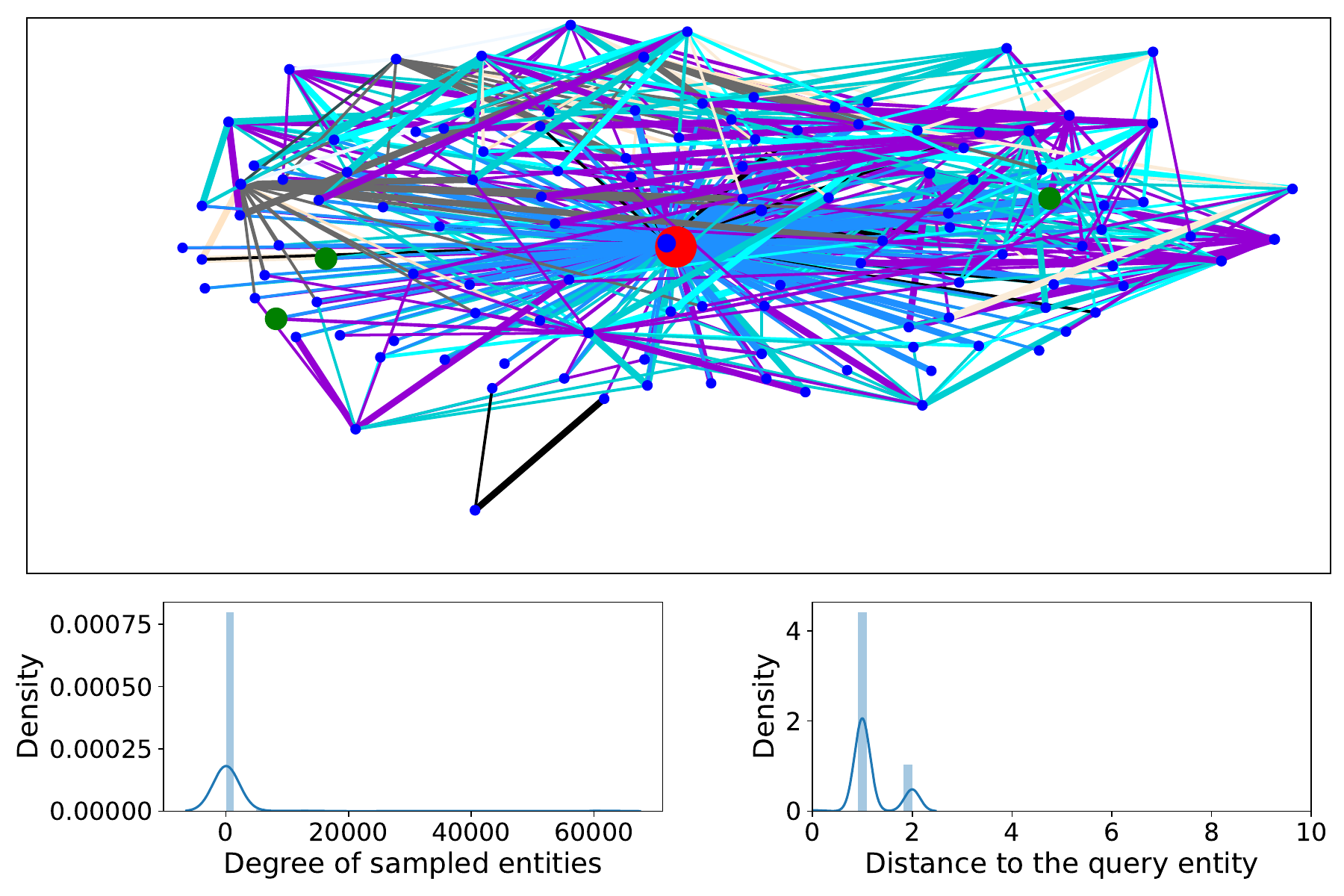}
		\hfill
		\includegraphics[width=6.8cm]{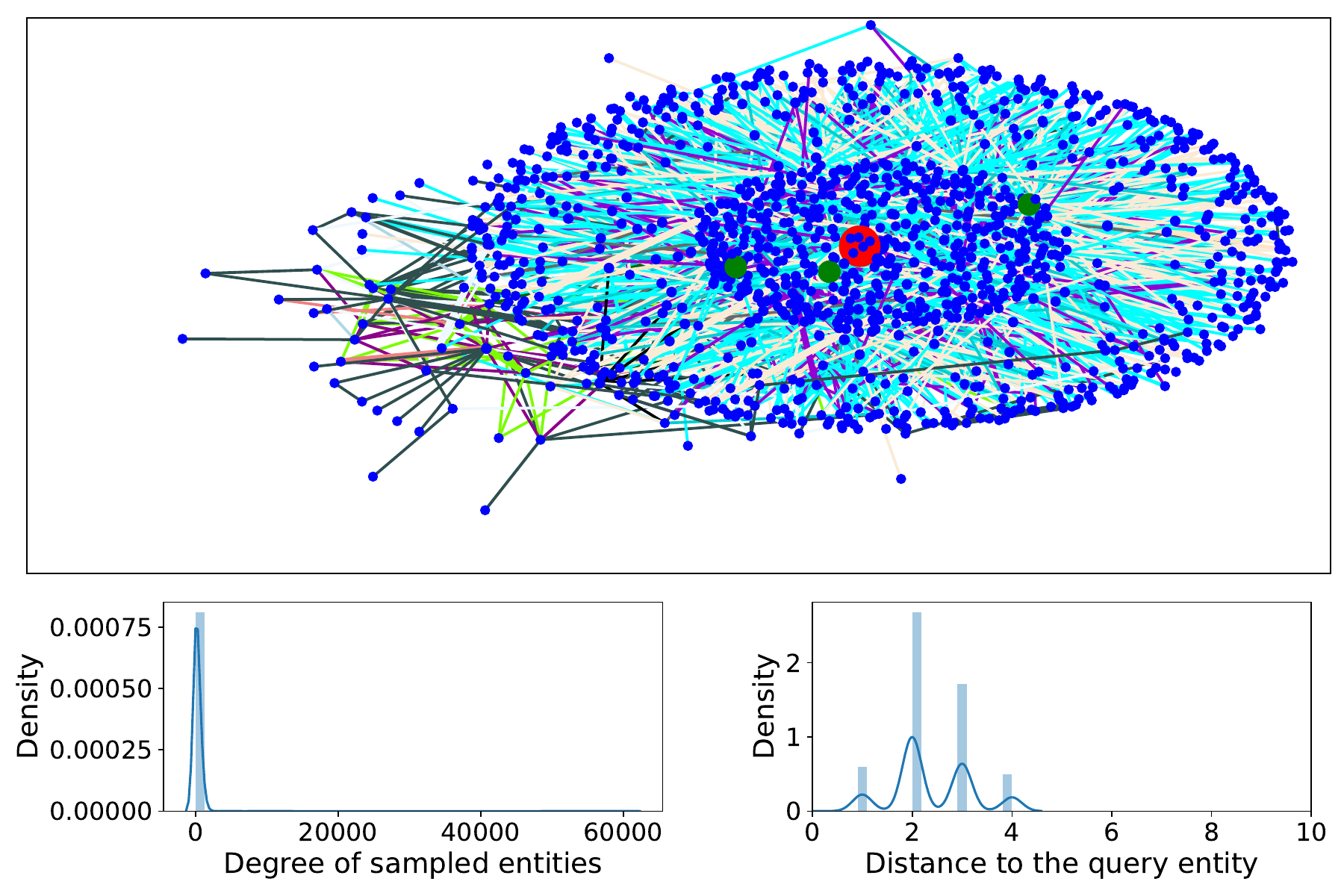}
		\hfill
		\vspace{-8px}
		\caption{
			Subgraphs (0.1\% and 1\%) from YAGO3-10:
			$u \! \! = \! \! 1255, q \! \! = \! \! 38, v \! \! = \! \! \{ 12418, 28138, 71366 \}$. 
		}
		\vspace{-4px}
	\end{figure*}

	\begin{figure*}[ht]
		\centering
		\hfill
		\includegraphics[width=6.8cm]{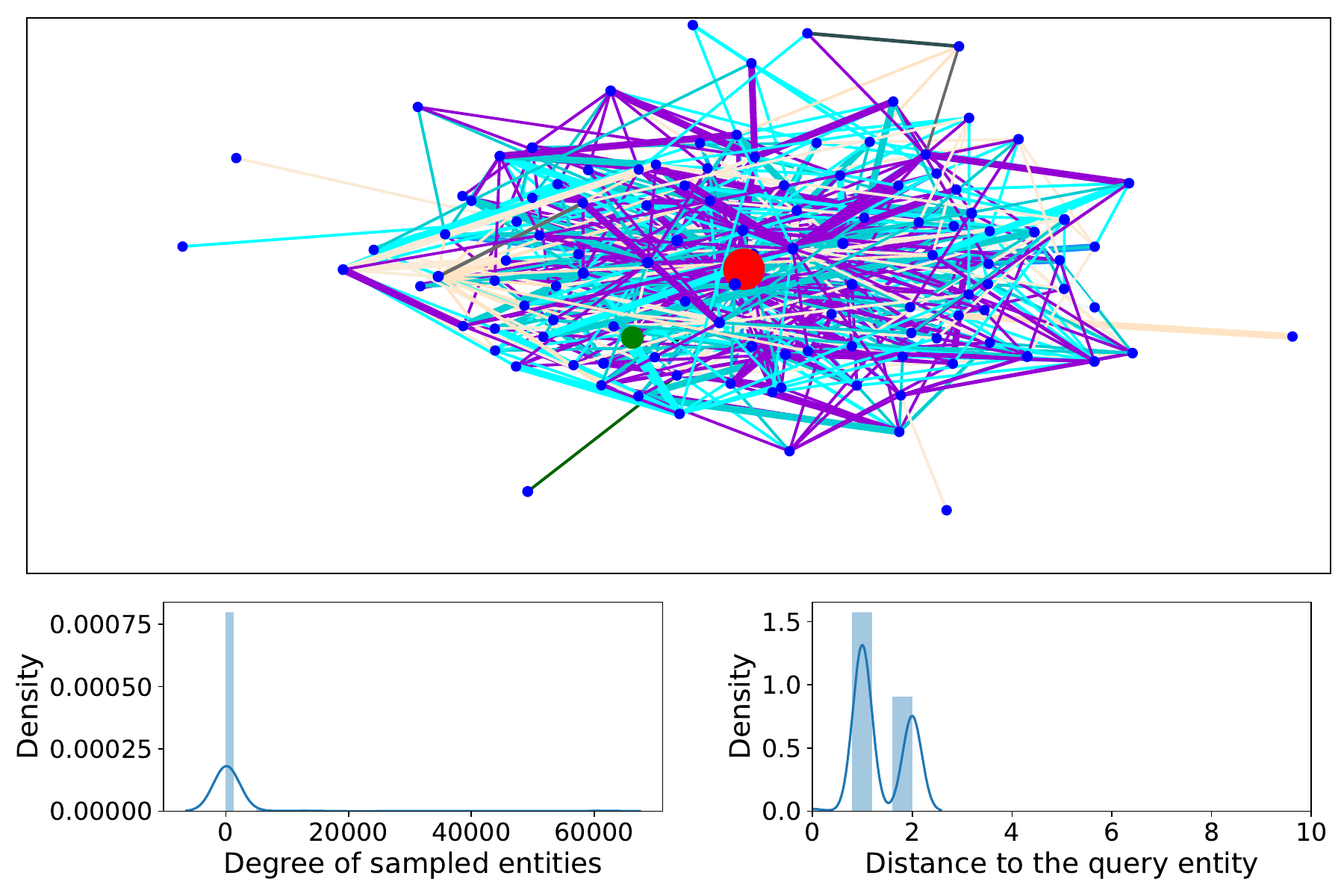}
		\hfill
		\includegraphics[width=6.8cm]{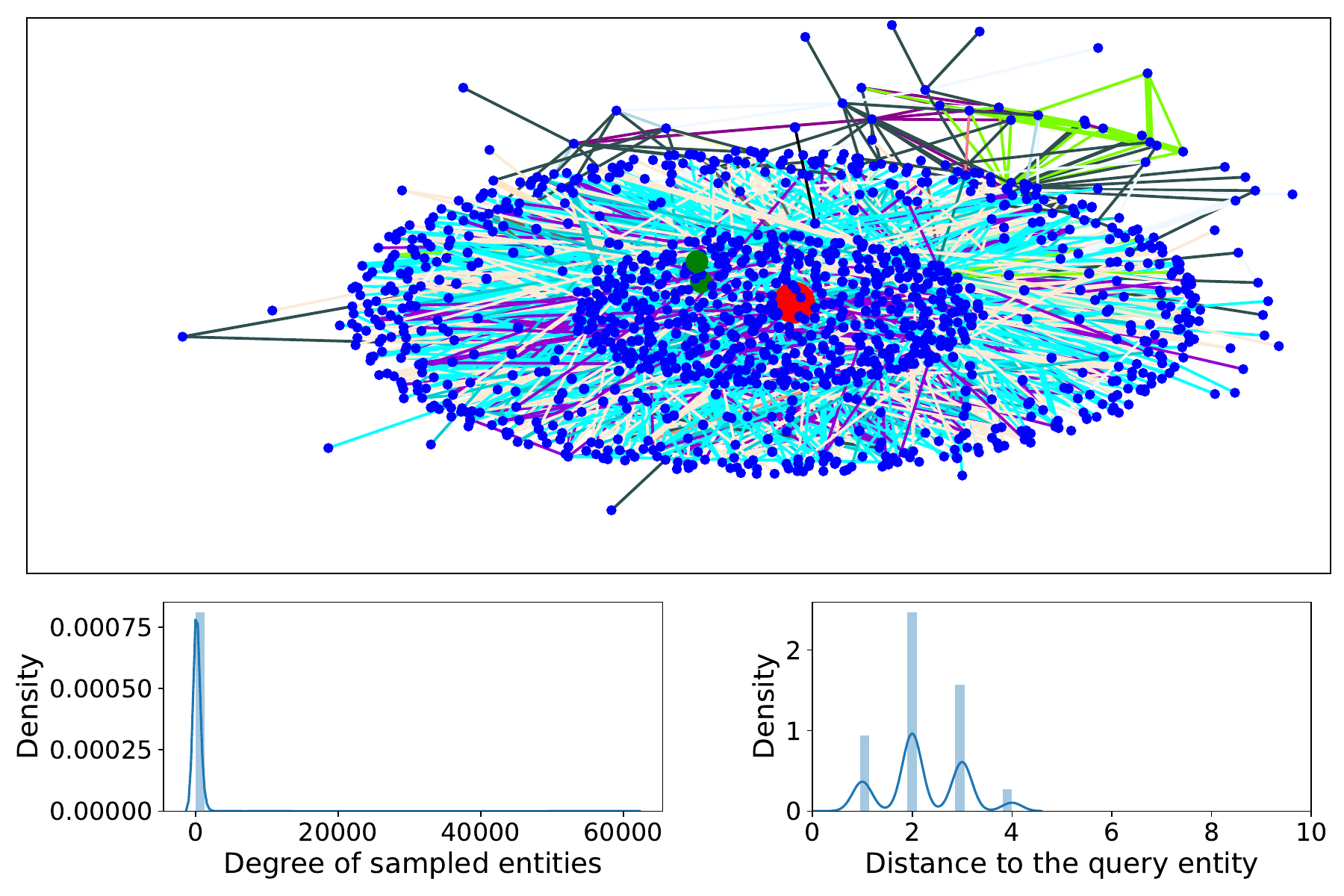}
		\hfill
		\vspace{-8px}
		\caption{
			Subgraphs (0.1\% and 1\%) from YAGO3-10:
			$u \! = \! 2252, q \! = \! 39, v \! = \! \{ 9476, 77502 \}$. 
		}
		\vspace{-4px}
	\end{figure*}


	\end{document}

%% file: math_commands.tex
\usepackage{amsmath,amsfonts,bm}

\def\eqref#1{equation~\ref{#1}}

\def\1{\bm{1}}

\DeclareMathAlphabet{\mathsfit}{\encodingdefault}{\sfdefault}{m}{sl}
\SetMathAlphabet{\mathsfit}{bold}{\encodingdefault}{\sfdefault}{bx}{n}

